\RequirePackage{fix-cm}
\documentclass[11pt,a4paper,twocolumn]{article}
\usepackage[T1]{fontenc}
\usepackage{acrc_preprint}
\usepackage{graphicx,booktabs,tabularx,longtable,enumitem,amsmath,amssymb,amsthm,pgf,needspace,multirow}
\usepackage{balance}
\usepackage{hyperref}
\hypersetup{colorlinks=true,citecolor=blue,linkcolor=blue,urlcolor=blue,
  pdftitle={Efficient Conformal Calibration for Paired Recommendation Decisions},
  pdfauthor={Xinyu Qiao, Yichen Lin, Zhewen Chen, Xue Wang, Tao Yao},
  pdfsubject={Research preprint},
  pdfkeywords={Conformal risk control, Recommendation, Finite-population inference}}
\setlist{nosep}
\colorlet{bestred}{red}
\colorlet{secondblue}{blue}
\newcommand{\best}[1]{\textcolor{bestred}{#1}}
\newcommand{\second}[1]{\textcolor{secondblue}{#1}}
\newcommand{\acrcours}{ACRC\,{\scriptsize(ours)}}
\newcommand{\E}{\mathbb E}
\providecommand{\mathdefault}[1]{#1}
\newtheorem{corollary}{Corollary}
\newtheorem{theorem}{Theorem}[section]
\newtheorem{proposition}[theorem]{Proposition}
\providecommand{\theHalgorithm}{\arabic{algorithm}}
\begin{document}
\twocolumn[
\begin{center}
{\fontsize{16}{19}\selectfont\bfseries
Efficient Conformal Calibration\par
for Paired Recommendation Decisions\par}
\vspace{1.2em}
{\large\bfseries
Xinyu Qiao\textsuperscript{1}\quad
Yichen Lin\textsuperscript{1}\quad
Zhewen Chen\textsuperscript{1}\quad
Xue Wang\textsuperscript{2}\quad
Tao Yao\textsuperscript{1,*}\par}
\vspace{0.8em}
{\normalsize
\textsuperscript{1}Antai College of Economics and Management\par
Shanghai Jiao Tong University, Shanghai, China\par
\textsuperscript{2}Meta\par}
\vspace{0.6em}
{\small\ttfamily
\{qxinyu25,linyichen,chenzhewen,taoyao\}@sjtu.edu.cn\par
xuewang0725@meta.com\par}

\vspace{1.4em}
\end{center}
]
\begingroup
\renewcommand{\thefootnote}{*}
\footnotetext{Corresponding author: Tao Yao, \href{mailto:taoyao@sjtu.edu.cn}{taoyao@sjtu.edu.cn}.}
\endgroup
\setcounter{footnote}{0}

\begin{abstract}
Risk control can suppress useful changes to a frozen recommendation rule. We study how to recover valid activation from the same observed fractional losses. Two specified internal calibrators can lose efficiency by compressing losses into their sum or clipping auxiliary inverse counts before averaging. We propose Averaged Conformal Risk Calibration, or ACRC, which averages before clipping. A product identity proves expected-risk validity, while pointwise bounds characterize the additional activation. For fixed populations sampled without replacement, we connect activation to signed utility under heterogeneous users and switching constraints. Near the budgets, positive population mean gain and the stated nondegeneracy conditions give positive leading-order utility increments despite sampling dependence. Across six datasets, changing only calibration improves discounted net utility by $0.010226$ over raw and $0.008337$ over RB. A separately frozen Jester confirmation gives $0.011025\,{\scriptstyle\pm\,0.001082}$ over raw with the rule and filter fixed, using a five-seed approximate 95\% paired interval. Against information-matched SCoRE selecting risk or reward scores, the mean utility difference is $+0.015934\,{\scriptstyle\pm\,0.001440}$. These gains use more mean realized risk within the declared budgets. Upstream failures delimit the benefit: efficient calibration releases useful modifications when the proposals themselves have value.

\end{abstract}

\section{Introduction}
A postprocessing rule can change a frozen recommender's items, order, and list length. Such modifications may improve average utility while harming some users. Conformal risk control limits this downside using held-out losses \cite{crc,rcrecsys,conformalk}. For a fixed proposal, we ask whether the same observed losses can permit more useful modifications under the same expected-risk guarantee.

Our starting point is information already present in fractional losses. Their magnitudes distinguish a small deterioration from a large one. Compressing them into a sum discards structure, while auxiliary Bernoulli randomization followed by clipping each inverse count introduces a different loss. ACRC averages the inverse count before clipping. This simple change needs two separate arguments. Jensen's inequality establishes no smaller activation, and an augmented product identity proves that the larger probability remains valid. Randomization and conditional averaging are existing tools \cite{schlag,fuzzy,score}. The new result is a specific valid operator with a characterized efficiency improvement.

Whether additional activation improves decisions is a second question. Some users have negative gains, and calibration and evaluation use complementary subsets of a fixed graph population. We retain the resulting sampling covariance and account for switching risk constraints. Near the budgets, positive population mean gain and nondegenerate auxiliary variance yield positive leading-order utility increments. This connects operator efficiency to a decision benefit under explicit conditions, rather than treating higher activation as a utility guarantee.

We separate calibration from candidate construction and selection. Same-rule raw and RB controls and a method-frozen Jester confirmation isolate calibration. Information-matched SCoRE and complete-method comparisons assess wider performance, whose gains also depend on the proposals.

Our contributions are threefold.
\begin{enumerate}[leftmargin=*,nosep]
\item \textbf{More efficient valid calibration.} For a frozen rule, we construct a fractional-loss operator, prove exchangeable risk validity, and characterize its pointwise activation improvement over two specified internal calibrators, including strictness and saturation conditions.
\item \textbf{A utility connection for fixed populations.} We derive the exact sampling-covariance correction and positive local utility increments of order $n^{-1}$ over raw and $n^{-1/2}$ over RB under positive population mean gain and the stated nondegeneracy conditions, allowing heterogeneous gains and switching constraints.
\item \textbf{Isolated recommendation gains.} Same-rule controls and a frozen confirmation establish a calibration increment distinct from fitted heads and optional filtering. Information-matched and complete-method comparisons test its applicability beyond the scalar operator class.
\end{enumerate}

\section{Related Work}
\paragraph{Risk control and decision policies.}
CRC controls expected bounded monotone losses \cite{crc}, while Conformal Risk Training optimizes through calibration \cite{crt}. Reference decisions and fallback policies already appear in risk-controlled postprocessing and policy control \cite{postproc,arbitrage,cpc}. Conformal Robustness Control optimizes a probability-constrained robustness certificate \cite{robustcontrol2026}. Joint certification covers an iid selective-risk ratio, acceptance floor, and utility lower bound \cite{jointcert2026}. We instead analyze the calibration efficiency of a frozen rule for mean downside magnitude on fixed populations.

\paragraph{Evidence and calibration operators.}
SCoRE uses risk-adjusted $e$-values satisfying $\E[EL]\le1$ \cite{score}. Our $E=w_+/\alpha$ meets this condition, so validity belongs to that general framework. P2E, evidence averaging, and adaptive coverage provide related transformations \cite{p2e2026,escope2025,adaptivecoverage2026}. SCoRE MDR makes user-specific decisions from pretrained risk scores and equals CRC on its threshold family. Our theoretical comparisons concern a specified scalar operator class. The same-information experiment tests, rather than assumes, its decision advantage over user selection.

\paragraph{Recommendation calibration.}
PerK combines userwise probability calibration with personalized prefixes \cite{perk}. UMC learns a contextual monotone calibrator whose item-dependent context can change ranking \cite{umc}. Conformal@K controls prefix miss-risk and also considers fractional $1-\mathrm{Recall}$ loss \cite{conformalk}. CARE treats interactive threshold adaptation \cite{care}, and ReCIRC learns conditional risk curves before CRC \cite{recirc2026}. MIR learns reranking through interactions between history and candidates \cite{mir}. Appendix~\ref{app:mir-comparison} reports its supplementary external comparison. Dynamic length, frozen-model wrapping, and fractional recommendation losses are therefore existing capabilities. Our claim concerns the final calibration operator and its finite-population utility consequences, not those capabilities themselves.

\section{Paired Recommendation Decisions}\label{sec:problem}
Let $\pi_0(u)$ be a frozen reference list and $a$ a modified list of length at most $K$. For observed binary relevance $Y_{ui}$, define discounted net utility with an irrelevant-item cost $\lambda\ge0$.

\begin{align}
 U_\lambda(a,Y_u)&=\frac{1}{W_K}\sum_{r=1}^{|a|}d_r\big[(1+\lambda)Y_{u,i_r}-\lambda\big],\label{eq:utility}\\
 d_r&=\frac{1}{\log_2(r+1)},\qquad W_K=\sum_{r=1}^K d_r.\notag
\end{align}

Each relevant item earns one unit and each irrelevant item costs $\lambda$ units before position discounting. This makes ranking and length jointly consequential: deleting an irrelevant item can help, while deleting a relevant item loses reward. The common denominator prevents shorter lists from gaining utility solely by changing normalization. We use $K=20$ and the symmetric valuation $\lambda=1$. This is a declared evaluation preference, not an estimated platform cost or measured profit. Rating-derived labels represent relevance rather than clicks. NDCG, Recall, and length are reported separately, and Table~\ref{tab:cost-support} tests alternative costs.

A candidate rule $j$ specifies a frozen distribution $\nu_j(a\mid u)$ over complete lists. Write the action's paired gain as $G_{ua}=U_\lambda(a,Y_u)-U_\lambda(\pi_0(u),Y_u)$ and its Recall change as $R_{ua}$. These gains lie in $[-B,B]$ for $B=1+\lambda$. Define

\begin{align}
 g_{ju}&=\sum_a\nu_j(a\mid u)G_{ua},\notag\\
 x_{ju}&=\sum_a\nu_j(a\mid u)(-G_{ua})_+,\label{eq:actual}\\
 x^R_{ju}&=\sum_a\nu_j(a\mid u)(-R_{ua})_+.\notag
\end{align}

The positive part must precede the action average. In general $x_{ju}\ne(-g_{ju})_+$. If the calibrated activation probability is $p_C$, the target risks are $\E[p_Cx_{ju}]\le\epsilon_U$ and $\E[p_Cx^R_{ju}]\le\epsilon_R$. They control mean downside magnitude, not the probability of any loss or the loss of every user.

We condition on the historical graph, the trained backbone, and the observed feedback array. Users are assigned without replacement to a fitting set $F$, a selection set $S$, a final calibration set $C$, and evaluation set $T$. The heads use $F$, candidate selection uses $S$, and final probabilities use $C$. The test labels do not enter these decisions. This protocol permits graph dependence because its randomness is the user-role assignment. It does not identify unobserved feedback or responses to changed online positions.

\paragraph{What is randomized in graph recommendation.}
Graph users share embeddings and neighborhoods, so disjoint labels do not imply independence. After fitting and selection, each remaining user has fixed losses and gain. Randomly assigning $C$ and evaluating its complement gives the design in Theorem~\ref{thm:dual-risk}. Algorithms read only their assigned labels. Repeated assignments vary allocation, while training seeds vary the backbone.

\section{ACRC with Fractional Losses}\label{sec:calibrator}
Consider a frozen rule with losses $l=(l_1,\ldots,l_n)\in[0,1]^n$. Let $n\ge1$, $0<\alpha<1$, $N=n+1$, and $A=\alpha N$. Introduce conditionally independent auxiliary variables $B_i\sim\mathrm{Bernoulli}(l_i)$. They are a calculation device and do not assume independent user feedback. With $K_B=\sum_iB_i$, compare

\begin{align}
 w_0(l)&=\min\left\{1,\frac{A}{1+\sum_i l_i}\right\},\label{eq:raw}\\
 w_{\rm RB}(l)&=\E_{B\mid l}\min\left\{1,\frac{A}{1+K_B}\right\},\label{eq:rb}\\
 a(l)&=A\E_{B\mid l}\frac{1}{1+K_B}
       =A\int_0^1\prod_{i=1}^n(1-tl_i)\,dt,\label{eq:unclipped}\\
 w_+(l)&=\min\{1,a(l)\}.\label{eq:plus}
\end{align}

The proposed $w_+$ is the ACRC calibrator. We call $w_0$ the raw internal calibrator and $w_{\rm RB}$ the Rao-Blackwell internal calibrator. They are specified operator comparisons, not reproductions of complete external methods. The difference is the placement of the final clipping operation.

\paragraph{What the loss magnitudes retain.}
Binary indicators count small and large deteriorations alike, whereas our budget charges their magnitudes. Equal loss sums can also conceal different opportunities. Consider $n=2$ and $\alpha=0.4$. The loss vectors $(1,0)$ and $(0.5,0.5)$ both have sum one. The binary vector makes every operator equal to $0.6$. For the fractional vector, the auxiliary count has probabilities $0.25$, $0.5$, and $0.25$ at zero, one, and two, giving
\begin{equation*}
 w_0=0.6,\qquad w_{\rm RB}=0.65,\qquad w_+=0.7.
\end{equation*}
Raw discards the fractional structure. RB retains it but clips the zero-count outcome before averaging. Clipping only the final average recovers another $0.05$ from the same observed losses.

\subsection{Validity and pointwise efficiency}
The next identity controls all leave-one-out risks at once. Its comparison statements concern activation probability, not signed utility.
\begin{theorem}\label{thm:valid}
For any exchangeable $(L_1,\ldots,L_N)\in[0,1]^N$,

\begin{equation}
 \E[L_Nw_+(L_{1:n})]\le\alpha.
\end{equation}

For every observed vector $l$, $w_+(l)\ge\max\{w_0(l),w_{\rm RB}(l)\}$. Moreover,

\begin{equation}
 \sum_{i=1}^N l_i a(l_{-i})
   =A\left[1-\prod_{i=1}^N(1-l_i)\right]\le A.\label{eq:identity}
\end{equation}
\end{theorem}

\begin{proof}
Integrate the negative derivative of $\prod_i(1-tl_i)$ on $[0,1]$. Since $w_+\le a$, divide the leave-one-out identity by $N$ and use exchangeability. Convexity of the reciprocal gives the raw comparison. Concavity of $z\mapsto\min\{1,z\}$ gives the RB comparison.
\end{proof}

\paragraph{Why the larger probability remains valid.}
Jensen's inequality compares probabilities but does not control their product with future loss. Equation~\eqref{eq:identity} supplies that control uniformly over augmented loss vectors.

\paragraph{A finite-sample boundary for the improvement.}
The unquantized scalar operators admit a directly computable bound. Write $s=\sum_i l_i$ and $V=\sum_i l_i(1-l_i)$. The exact reciprocal identity yields
\begin{equation}
 a-\frac{A}{1+s}
 =\frac{A}{(1+s)^2}\mathbb E_{B\mid l}
   \frac{(K_B-s)^2}{1+K_B}.
 \label{eq:main-fractional-excess}
\end{equation}
Consequently,
\begin{equation}
 0\le w_+-w_0\le
 \min\left\{1-w_0,\frac{AV}{(1+s)^2}\right\}.
 \label{eq:main-fractional-headroom}
\end{equation}
The difference is strictly positive exactly when $w_0<1$ and $V>0$. Here $V$ measures auxiliary variance, not dispersion across users. Equal interior losses can therefore support an improvement, whereas widely dispersed binary losses cannot. The two terms in the bound expose distinct limits: no fractional uncertainty remains when $V=0$, and no additional probability remains when raw already activates fully. This describes potential activation improvement without assuming favorable evaluation gains. Proposition~\ref{prop:fractional} supplies the proof.

The additional truncation condition for strict improvement over RB appears in Proposition~\ref{prop:rb-strict-regime}.

Within the class of scalar calibrators valid for every exchangeable loss law, $w_+$ is admissible and shares the exact minimax value with RB. Appendix Theorem~\ref{thm:minimax} characterizes this class. These comparisons do not establish dominance over input-adaptive methods.

\section{From Activation to Recommendation Utility}\label{sec:utility-theory}
\subsection{Actual losses and two constraints}
A smaller valid action range improves calibration without changing the physical budget. Before final calibration, fix label-independent bounds $x_{ju}\le H_j$ and $x^R_{ju}\le H^R_j$. For constraint $k$, use loss $L_{kj}=x^k_j/H^k_j$ and level $\alpha_{kj}=\epsilon_k/H^k_j$ when $H^k_j>\epsilon_k$. If $H^k_j\le\epsilon_k$, set that probability to one. For an active constraint let $w_{+,kj}$ denote~\eqref{eq:plus} with its corresponding level. The implemented probability is

\begin{align}
 p_{+,j}(C)&=Q_q\left(\min_k w_{+,kj}(L_{kj,C})\right),\label{eq:dual}\\
 Q_q(z)&=q\lfloor z/q\rfloor,\qquad q=2^{-53}.\notag
\end{align}

The minimum is one if there are no active constraints.
\begin{theorem}\label{thm:dual-risk}
Condition on all fitting information and on any candidate choice that reads only $S$. Uniformly sample $n<M$ calibration users from the remaining fixed pool $R$ of size $M$, and evaluate its complement. With independent activation and internal action coins,

\begin{align}
 \E_C[p_{+,j}(C)\overline x_{j,R\setminus C}]&\le\epsilon_U,\notag\\
 \E_C[p_{+,j}(C)\overline{x^R}_{j,R\setminus C}]&\le\epsilon_R.\label{eq:dual-risk}
\end{align}
\end{theorem}

The losses need not be independent. The guarantee averages over calibration assignments and serving randomization. It is not conditional on a single realized calibration set. The proof is a leave-one-out application of Theorem~\ref{thm:valid}, followed by range rescaling and pointwise domination of each gate by their minimum.

\subsection{From additional activation to signed utility}
Fix the same rule and ranges, and replace $w_+$ by $w_0$ in~\eqref{eq:dual} to define $p_0$. Write $d(C)=p_+(C)-p_0(C)\ge0$, $\tau=\E_Cd(C)$, $\mu=\overline g_R$, and $\Delta D=\E_C[d(C)\overline g_{R\setminus C}]$. Uniform sampling gives
\begin{equation}
 \Delta D=\mu\tau+\operatorname{Cov}_C(d(C),\overline g_{R\setminus C}).\label{eq:increment}
\end{equation}
Positive mean gain alone does not remove this dependence. A product $\mu\tau$ larger than the absolute covariance is sufficient for a positive increment. Appendix Theorem~\ref{thm:increment} bounds the correction for randomized actions and two constraints. The bound uses population quantities and is not an implemented $S$-only utility certificate. For each realized split, let $U_c(T)$ average both activation and internal action coins analytically. Then
\begin{equation}
 U_+(T)-U_0(T)=d(C)\overline g_T.\label{eq:realized}
\end{equation}
The next theorem gives positive expected increments with heterogeneous users and switching constraints.

\paragraph{Dependence and risk bottlenecks.}
The covariance in~\eqref{eq:increment} varies the calibration set and its evaluation complement. It can have either sign. A second obstacle is the minimum across constraints: increasing a nonbinding probability may leave activation unchanged. The next theorem handles both effects jointly.

\subsection{Heterogeneous gains under multiple risk constraints}
When both risks approach their budgets, the limiting constraint can change across
user splits. We retain this switching and the dependence between losses.

\begin{theorem}[Utility gains with heterogeneous users]\label{thm:heterogeneous-utility-main}
After freezing the proposal, consider fixed pools of size $M$ with loss vectors
$l_u\in[0,1]^2$, mean $r_M$, and covariance $\Sigma_M\to\Sigma$.
Gains satisfy $|g_u|\le B_g$ and population mean $\mu_M\to\mu>0$.
Sample $n$ calibration users uniformly without replacement and evaluate the complement,
with
\begin{gather*}
 n\to\infty,\qquad n/M\to\rho\in[0,1),\\
 \sqrt n(\alpha_k-r_{M,k})\to h_k\in\mathbb R.
\end{gather*}
Budgets $\alpha_k\in(0,1)$ are fixed and
$\nu_k=\alpha_k(1-\alpha_k)-\Sigma_{kk}>0$.
Let $X\sim N_2(0,(1-\rho)\Sigma)$, $Z\sim N(0,1)$, and define
\begin{align*}
 q_k(x)&=(x_k-h_k)_+/\alpha_k,\\
 b_k(x)&=\mathbb E_Z(x_k-h_k+\sqrt{\nu_k}Z)_+/\alpha_k.
\end{align*}
For unquantized $p_c=\min_kw_{c,k}$ and
$\Delta D_c=\mathbb E_C[(p_+-p_c)\overline g_T]$,
\begin{multline}
 \sqrt n\Delta D_{\rm RB}\longrightarrow\\
 \mu\mathbb E\left[\max_kb_k(X)-\max_kq_k(X)\right]>0.
 \label{eq:heterogeneous-utility-main-rb}
\end{multline}
If $\Sigma$ is positive definite, set
$I_k=\mathbf1\{q_k(X)>\max(0,q_{3-k}(X))\}$. Then
\begin{equation}
 n\Delta D_0\longrightarrow
 \mu\sum_{k=1}^2\frac{\nu_k}{\alpha_k^2}\mathbb E I_k>0.
 \label{eq:heterogeneous-utility-main-raw}
\end{equation}
\end{theorem}

The two rates reflect different losses of efficiency. Against raw, averaging the reciprocal retains a curvature correction of order $n^{-1}$. Against RB, averaging before clipping removes an auxiliary fluctuation from the clipping loss, giving an order $n^{-1/2}$ increment near the budget. The coefficients account for changing bottlenecks, and sampling covariance is of smaller order under the theorem's conditions. RB averages within each constraint before taking the maximum deficit, so these operations cannot be interchanged. Appendices~\ref{app:heterogeneous-multigate} and~\ref{app:heterogeneous-efficiency} give the proofs and one-constraint specialization. The limits are not exact effect predictions at $n=200$ or guarantees against arbitrary user selectors.

\section{Recommendation with ACRC}\label{sec:algorithm}
ACRC calibrates a rule frozen before $C$ with valid loss bounds. Our recommendation instance fits heads on $F$ and uses common templates for Top20 or observed Top80 reorderings, replacements, prefixes, and mixtures. Fitted parameters and selected rules vary across datasets, while templates and the selection criterion remain fixed. Appendix~\ref{app:implementation} specifies the construction.

\paragraph{Label-independent action ranges.}
Their action structure also yields tighter valid ranges. For $\lambda=1$, let $c_a$ place each normalized position discount on its item and let $c_0$ represent the reference list. Then
\begin{align*}
 G_{ua}&=(c_a-c_0)^\top(2Y_u-1),\\
 |G_{ua}|&\le\|c_a-c_0\|_1.
\end{align*}
The maximum of this bound across the frozen user pool does not read evaluation labels. For a mixture, the corresponding weighted sum bounds its expected downside. A reordered full reference list preserves Recall, so mixing it with its prefix with probability $\eta$ gives Recall downside at most $\eta$. These bounds normalize each candidate's losses without changing the physical risk budgets. Final calibration still uses observed action-level downside, not the structural bound as a substitute for it.

The default selector shares $b=24$ size-$n$ pseudocalibration subsets $P_t$ of $S$ and chooses
\begin{equation}
 \widehat j\in\arg\max_j\frac1b\sum_{t=1}^b
 p_{0,j}(P_t)\overline g_{j,S\setminus P_t}.\label{eq:selection}
\end{equation}
This criterion estimates gain after raw calibration. Keeping the same selected rule in the raw, RB, and ACRC controls isolates the final operator. It does not jointly train the proposal through ACRC. Ties follow frozen candidate order, and selection precedes final calibration on $C$, as required by Theorem~\ref{thm:dual-risk}.

\begin{algorithm}[t]
\caption{Candidate selection followed by ACRC calibration}\label{alg:main}
\begin{algorithmic}[1]
\REQUIRE Frozen backbone, disjoint $F,S,C,T$, budgets $\epsilon_U,\epsilon_R$
\STATE Fit heads using $F$ and freeze candidate distributions and ranges
\FOR{each shared subset $P_t\subset S$}
\STATE Compute actual action downsides and two raw gates
\STATE Evaluate signed gain on $S\setminus P_t$
\ENDFOR
\STATE Choose by the raw objective~\eqref{eq:selection} and freeze $\widehat j$
\STATE Use only $C$ to compute $p_{+,\widehat j}$ by~\eqref{eq:dual}
\STATE Serve the original list with probability $1-p_{+,\widehat j}$
\STATE Otherwise sample a complete list from $\nu_{\widehat j}$
\STATE Evaluate all metrics on $T$ by averaging both coins analytically
\end{algorithmic}
\end{algorithm}

For numerical execution, a Poisson--binomial recurrence uses count probabilities $q_k$ to compute $a=A\sum_kq_k/(k+1)$ exactly for the stored binary64 inputs. Downward rounding to 53 binary digits preserves validity. The recurrence uses $O(n^2)$ arithmetic operations and $O(n)$ coefficients, with integer bit length affecting runtime. An optional $S$-only user mask can be frozen into a rule and calibrated on all of $C$. It is a separate extension, not part of the operator contribution. Appendix~\ref{app:decision-diagnosis} isolates its effects.

\section{Experiments}\label{sec:experiments}
We use KuaiRec, KuaiRand-random, Amazon-Music, Epinions, and MovieLens-25M for development, followed by a method-frozen Jester confirmation \cite{kuairec,kuairand,sigformer,movielens,jester}. Full LightGCN, XSimGCL, and SIGformer backbones each use five training seeds \cite{lightgcn,xsimgcl,sigformer}. Training, validation, and evaluation pairs are disjoint. We evaluate observed feedback only. Dataset-specific labels, splits, eligibility, and short-pool handling appear in Appendix~\ref{app:implementation}.

Heads use $F200$. Forty shared assignments use $S400$, $C200$, and the remaining $T$. Metrics are penalized utility, NDCG, Recall, list length, and both downside risks at budgets $0.02/0.01$. Assignments are averaged within each training seed before constructing approximate paired $t_4$ intervals from five seeds. Entries report means $\pm$ 95\% interval half-widths unless marked as standard deviations. Same-information intervals are unadjusted and condition on six previously inspected domains. The frozen confirmation is a separate comparison.

\begin{table}[t]
\centering\small\setlength{\tabcolsep}{2.3pt}
\caption{Calibration for the same selected proposal. Five-seed means average forty splits per model, then equally weight six domains and three backbones. Risk limits are $0.02/0.01$. Decimal places are consistent within each column. A uses default budgets, B selects configurations on $S$. Red and blue mark the best and second-best means within each setting, not significance.}\label{tab:matched-calibration-main}\label{tab:revision-main}
\begin{tabular}{@{}llrrr@{}}\toprule
Set. & Calibrator & $U\uparrow$ & $D_U\downarrow$ & $D_R\downarrow$\\
\midrule
\multirow{3}{*}{A} & Raw & $0.1724$ & \best{$0.01113$} & \best{$0.00448$} \\
 & RB & \second{$0.1743$} & \second{$0.01129$} & \second{$0.00462$} \\
 & \acrcours{} & \best{$0.1826$} & $0.01249$ & $0.00498$ \\
\midrule
\multirow{4}{*}{B} & \acrcours{} & \best{$0.1826$} & $0.01249$ & $0.00498$ \\
 & SCoRE, risk prediction & $0.1657$ & \second{$0.00941$} & \best{$0.00320$} \\
 & SCoRE, reward ratio & \second{$0.1672$} & $0.00942$ & $0.00328$ \\
 & SCoRE, risk or reward & $0.1667$ & \best{$0.00940$} & \second{$0.00323$} \\
\bottomrule\end{tabular}
\end{table}

\begin{table*}[t]
\centering\footnotesize\setlength{\tabcolsep}{2.0pt}
\caption{Complete-procedure utility, higher is better. All six columns are ranked jointly within each row, with best means in red and second-best in blue. Displayed ties share a color. Values are five-seed means $\pm$ 95\% $t_4$ interval half-widths. LG, X, SIG denote LightGCN\cite{lightgcn}, XSimGCL\cite{xsimgcl}, SIGformer\cite{sigformer}. Native objectives are unchanged. SCoRE paths and ACRC share risk budgets, not realized risks. Jester uses common-support users without filtering. Risks and MIR results appear in Appendix~\ref{app:external-complete}.}\label{tab:utility}\label{tab:score-main}
\begin{tabular}{@{}llrrrrrr@{}}\toprule
Dataset & Model & Top20 & PerK\cite{perk} & UMC\cite{umc} & SCoRE/PerK\cite{score} & SCoRE/UMC\cite{score} & \acrcours{}\\\midrule
\multirow{3}{*}{KuaiRec} & LG & $0.2233$ {\scriptsize$\pm$ 0.0067} & $0.2242$ {\scriptsize$\pm$ 0.0066} & \second{$0.2763$ {\scriptsize$\pm$ 0.0471}} & $0.2242$ {\scriptsize$\pm$ 0.0066} & $0.2278$ {\scriptsize$\pm$ 0.0074} & \best{$0.5391$ {\scriptsize$\pm$ 0.0085}}\\
 & X & \second{$0.2532$ {\scriptsize$\pm$ 0.0086}} & $0.2344$ {\scriptsize$\pm$ 0.0114} & $0.1766$ {\scriptsize$\pm$ 0.1438} & $0.2464$ {\scriptsize$\pm$ 0.0088} & $0.2499$ {\scriptsize$\pm$ 0.0118} & \best{$0.5221$ {\scriptsize$\pm$ 0.0122}}\\
 & SIG & \second{$0.6812$ {\scriptsize$\pm$ 0.0162}} & \second{$0.6812$ {\scriptsize$\pm$ 0.0162}} & $0.6626$ {\scriptsize$\pm$ 0.0220} & \second{$0.6812$ {\scriptsize$\pm$ 0.0162}} & $0.6791$ {\scriptsize$\pm$ 0.0148} & \best{$0.7543$ {\scriptsize$\pm$ 0.0104}}\\
\midrule
\multirow{3}{*}{KuaiRand} & LG & $-0.4960$ {\scriptsize$\pm$ 0.0006} & \best{$-0.2069$ {\scriptsize$\pm$ 0.0008}} & $-0.5933$ {\scriptsize$\pm$ 0.0260} & $-0.4883$ {\scriptsize$\pm$ 0.0013} & $-0.4977$ {\scriptsize$\pm$ 0.0007} & \second{$-0.4846$ {\scriptsize$\pm$ 0.0007}}\\
 & X & $-0.5096$ {\scriptsize$\pm$ 0.0014} & \best{$-0.1584$ {\scriptsize$\pm$ 0.0022}} & $-0.6102$ {\scriptsize$\pm$ 0.0264} & $-0.5008$ {\scriptsize$\pm$ 0.0022} & $-0.5113$ {\scriptsize$\pm$ 0.0015} & \second{$-0.4977$ {\scriptsize$\pm$ 0.0015}}\\
 & SIG & $-0.4675$ {\scriptsize$\pm$ 0.0003} & \best{$-0.2127$ {\scriptsize$\pm$ 0.0005}} & $-0.5644$ {\scriptsize$\pm$ 0.0380} & $-0.4608$ {\scriptsize$\pm$ 0.0005} & $-0.4695$ {\scriptsize$\pm$ 0.0003} & \second{$-0.4555$ {\scriptsize$\pm$ 0.0005}}\\
\midrule
\multirow{3}{*}{Music} & LG & \second{$0.2603$ {\scriptsize$\pm$ 0.0009}} & $0.2239$ {\scriptsize$\pm$ 0.0027} & $0.2553$ {\scriptsize$\pm$ 0.0009} & $0.2589$ {\scriptsize$\pm$ 0.0008} & $0.2576$ {\scriptsize$\pm$ 0.0014} & \best{$0.2663$ {\scriptsize$\pm$ 0.0008}}\\
 & X & \second{$0.2595$ {\scriptsize$\pm$ 0.0011}} & $0.2119$ {\scriptsize$\pm$ 0.0010} & $0.2586$ {\scriptsize$\pm$ 0.0013} & $0.2582$ {\scriptsize$\pm$ 0.0012} & $0.2586$ {\scriptsize$\pm$ 0.0011} & \best{$0.2663$ {\scriptsize$\pm$ 0.0007}}\\
 & SIG & $0.2524$ {\scriptsize$\pm$ 0.0004} & $0.2202$ {\scriptsize$\pm$ 0.0013} & \second{$0.2570$ {\scriptsize$\pm$ 0.0027}} & $0.2511$ {\scriptsize$\pm$ 0.0005} & $0.2560$ {\scriptsize$\pm$ 0.0022} & \best{$0.2660$ {\scriptsize$\pm$ 0.0004}}\\
\midrule
\multirow{3}{*}{Epinions} & LG & \second{$0.2313$ {\scriptsize$\pm$ 0.0005}} & $0.1836$ {\scriptsize$\pm$ 0.0009} & $0.2247$ {\scriptsize$\pm$ 0.0049} & $0.2304$ {\scriptsize$\pm$ 0.0005} & $0.2302$ {\scriptsize$\pm$ 0.0008} & \best{$0.2596$ {\scriptsize$\pm$ 0.0020}}\\
 & X & $0.2328$ {\scriptsize$\pm$ 0.0006} & $0.1757$ {\scriptsize$\pm$ 0.0014} & \second{$0.2340$ {\scriptsize$\pm$ 0.0023}} & $0.2319$ {\scriptsize$\pm$ 0.0005} & $0.2336$ {\scriptsize$\pm$ 0.0011} & \best{$0.2613$ {\scriptsize$\pm$ 0.0013}}\\
 & SIG & $0.2140$ {\scriptsize$\pm$ 0.0005} & $0.1723$ {\scriptsize$\pm$ 0.0009} & \second{$0.2270$ {\scriptsize$\pm$ 0.0037}} & $0.2128$ {\scriptsize$\pm$ 0.0004} & $0.2185$ {\scriptsize$\pm$ 0.0013} & \best{$0.2603$ {\scriptsize$\pm$ 0.0028}}\\
\midrule
\multirow{3}{*}{ML-25M} & LG & $0.1137$ {\scriptsize$\pm$ 0.0014} & \second{$0.1164$ {\scriptsize$\pm$ 0.0013}} & $-0.0604$ {\scriptsize$\pm$ 0.0500} & $0.1163$ {\scriptsize$\pm$ 0.0013} & $0.1110$ {\scriptsize$\pm$ 0.0013} & \best{$0.1559$ {\scriptsize$\pm$ 0.0025}}\\
 & X & $0.1077$ {\scriptsize$\pm$ 0.0009} & \second{$0.1223$ {\scriptsize$\pm$ 0.0013}} & $-0.1258$ {\scriptsize$\pm$ 0.0878} & $0.1117$ {\scriptsize$\pm$ 0.0007} & $0.1051$ {\scriptsize$\pm$ 0.0008} & \best{$0.1616$ {\scriptsize$\pm$ 0.0115}}\\
 & SIG & $-0.0778$ {\scriptsize$\pm$ 0.0338} & \second{$-0.0761$ {\scriptsize$\pm$ 0.0339}} & $-0.1212$ {\scriptsize$\pm$ 0.0828} & $-0.0762$ {\scriptsize$\pm$ 0.0339} & $-0.0785$ {\scriptsize$\pm$ 0.0319} & \best{$0.0407$ {\scriptsize$\pm$ 0.0151}}\\
\midrule
\multirow{3}{*}{Jester} & LG & \second{$0.2608$ {\scriptsize$\pm$ 0.0012}} & $0.1897$ {\scriptsize$\pm$ 0.0021} & $0.2000$ {\scriptsize$\pm$ 0.0122} & $0.2598$ {\scriptsize$\pm$ 0.0012} & $0.2583$ {\scriptsize$\pm$ 0.0010} & \best{$0.3244$ {\scriptsize$\pm$ 0.0015}}\\
 & X & $0.2986$ {\scriptsize$\pm$ 0.0012} & $0.2310$ {\scriptsize$\pm$ 0.0012} & \second{$0.3253$ {\scriptsize$\pm$ 0.0026}} & $0.2978$ {\scriptsize$\pm$ 0.0012} & $0.3015$ {\scriptsize$\pm$ 0.0008} & \best{$0.3537$ {\scriptsize$\pm$ 0.0020}}\\
 & SIG & $0.1845$ {\scriptsize$\pm$ 0.0235} & $0.1058$ {\scriptsize$\pm$ 0.0189} & \second{$0.2169$ {\scriptsize$\pm$ 0.0684}} & $0.1832$ {\scriptsize$\pm$ 0.0236} & $0.1904$ {\scriptsize$\pm$ 0.0201} & \best{$0.2938$ {\scriptsize$\pm$ 0.0104}}\\
\bottomrule\end{tabular}
\end{table*}

\subsection{The increment from calibration alone}\label{sec:calibration-results}
Table~\ref{tab:revision-main} fixes the selected proposal, fitted heads, action losses, valid ranges, and $C/T$ users. In panel A, only the final scalar operator changes. The mean net-utility increments are $0.010226\pm0.000401$ over raw and $0.008337\pm0.000201$ over RB. These paired contrasts isolate calibration, rather than the full procedure's gain over Top20. They directly test the operator comparison in Sections~\ref{sec:calibrator} and~\ref{sec:utility-theory}. Raw and RB are the specified internal controls, not standalone external recommendation methods.

The mean risks increase from $0.011135/0.004481$ for raw to $0.012495/0.004983$ for $w_+$ under unchanged budgets. The result is more useful activation with additional risk expenditure. Section~\ref{sec:jester} gives a separate frozen Jester confirmation of this calibration increment.

Against raw, relevant reward and irrelevant-cost saving contribute $0.005024$ and $0.005202$. MovieLens improves without shortening, while KuaiRand mainly reduces irrelevant exposure. The five-domain cost sensitivity remains positive from $\lambda=0.25$ to $2$. Appendix~\ref{app:utility-accounting} separates these valuation-dependent gains from click utility, and Table~\ref{tab:cost-support} retains all costs.

\subsection{Comparison with information-matched SCoRE}
Panel B fixes proposals and fitted information. SCoRE retains official user-specific MDR decisions and receives both frozen $F$ heads. Besides risk prediction, we add the original risk-to-reward score using action-level positive gain as nonnegative reward. Features, hyperparameters and the choice between score families are selected only on $S$. Appendix~\ref{app:reward-score-comparison} specifies the signed-gain decomposition and its optimality limits.

\begin{figure*}[!t]\centering\includegraphics[width=\textwidth]{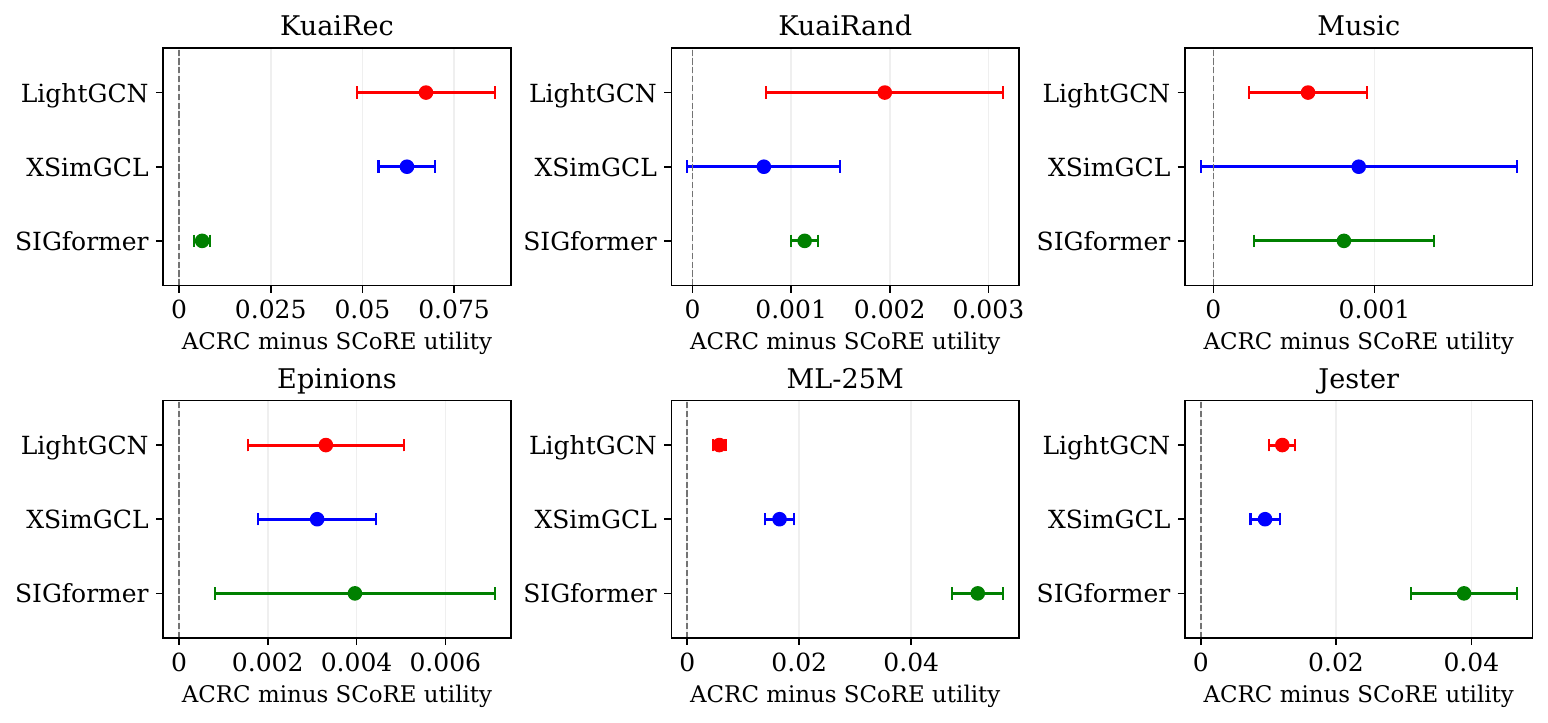}
\caption{ACRC minus SCoRE with feature and score families selected on $S$. Proposals and fitted information are shared. Bars are pointwise paired five-seed 95\% intervals. 18 of eighteen mean differences are positive and 16 intervals exclude zero in favor of ACRC. Panel scales differ.}\label{fig:restructured-paired}\end{figure*}

\begin{figure*}[!t]
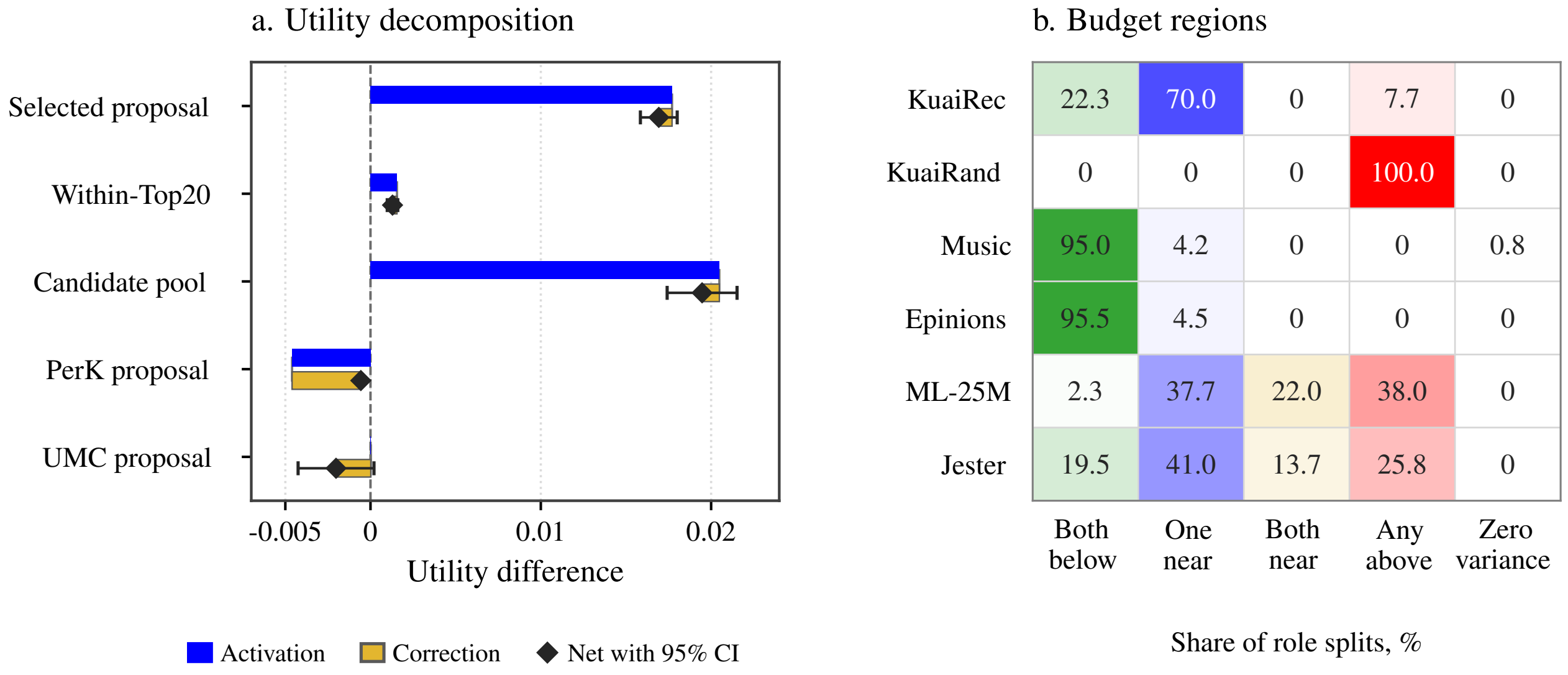

\centering
\begin{minipage}[t]{1.0\textwidth}
\centering
\begingroup\makeatletter
\resizebox{\linewidth}{!}{%
\begingroup%
\makeatletter%
%
\makeatother%
\endgroup%

}%
\makeatother\endgroup
\end{minipage}
\caption{Activation and its limits. Left, for risk-prediction SCoRE, blue gives $(p-\bar m)\overline g_T$, yellow adds $-\operatorname{Cov}_T(m,g)$, and diamonds show their sum with paired five-seed 95\% intervals. Right, percentages of splits by candidate-loss budget region, each row totaling 100\%. Near means within one sampling standard deviation. Regions use $C\cup T$ labels only for post-hoc description, not selection or verification of asymptotic assumptions. Details appear in Appendix~\ref{app:mechanisms-complete}.}
\label{fig:restructured-mechanism}
\end{figure*}

\begin{figure*}[!t]
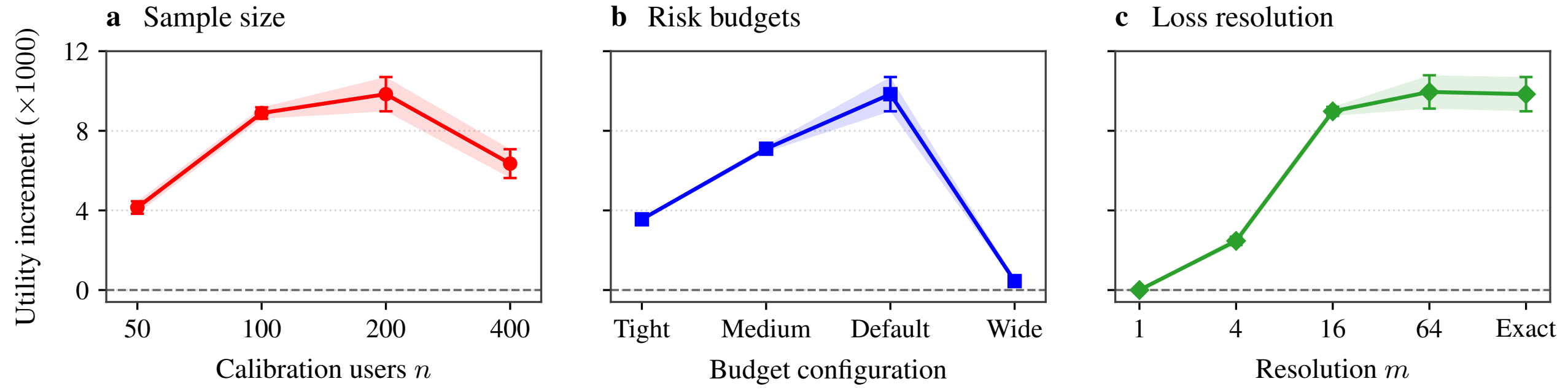

\centering
\begingroup\makeatletter
\resizebox{\textwidth}{!}{%
\begingroup%
\makeatletter%
%
\makeatother%
\endgroup%

}%
\makeatother\endgroup
\caption{Calibration sensitivity on fifteen development settings. Points show macro utility increments over raw, with pointwise five-seed 95\% intervals. All configurations are retained. Resolution $m$ uses $\lceil mL\rceil/m$, changing both fractionality and mean loss. Protocols and budgets appear in Appendix~\ref{app:mechanisms-complete}.}
\label{fig:mechanism}
\end{figure*}

With both score families available, ACRC's mean advantage is $0.015934\,{\scriptstyle\pm\,0.001440}$. All eighteen means are positive and sixteen intervals exclude zero in Figure~\ref{fig:restructured-paired}. The reward-ratio-only variant is stronger on $T$, with utility $0.167201$ and an ACRC advantage of $0.015445\,{\scriptstyle\pm\,0.001432}$, so it is also retained in Table~\ref{tab:revision-main}. Selection on $S$ need not outperform every family on $T$.

Mean risks are $0.012495/0.004983$ for ACRC and $0.009397/0.003230$ for the two-family SCoRE control. This compares common budgets, not matched realized risks. The separate risk-prediction comparison still favors SCoRE on native PerK proposals by $0.000566\,{\scriptstyle\pm\,0.000105}$, and the UMC contrast includes zero. Appendix~\ref{app:matched-information} retains these proposal-dependent failures.

\subsection{Complete recommendation performance}\label{sec:main-results}
Table~\ref{tab:utility} evaluates the full procedure against backbone Top20, PerK-PDCG, and UMC-Top20 on common-support users. SCoRE intersects its unchanged MDR selections for the two losses, with $\gamma=\alpha$ and risk scores fitted on $S$. PerK uses historical validation labels, whereas our heads and UMC use $F$. These complete comparisons therefore include proposal and information differences as well as calibration.

Our mean utility exceeds Top20, UMC, and both SCoRE applications in all eighteen settings. PerK remains better on all three KuaiRand settings, returning $5.7$--$9.0$ items with Recall downside $0.31$--$0.35$, versus below $0.01$ for our longer lists. Our default also loses in two of fifteen development settings against the best native or SCoRE-path point satisfying both empirical risk ceilings. Appendix~\ref{app:frontier} preserves this descriptive comparison of operating points with different objectives.

Appendix~\ref{app:mir-comparison} retains MIR as an external reranker. MIR leads on all three MovieLens backbones and our independent procedure on all three KuaiRand backbones, under different risk constraints. A separate frozen-MIR experiment isolates calibration by fixing proposals. MIR remains outside the main algorithm, and total modification gains are distinct from calibration increments.

\subsection{Why activation helps, and where it does not}\label{sec:mechanism}
The same-rule sampling identity~\eqref{eq:increment} has a direct finite-pool check in Appendix~\ref{app:sampling-decomposition}. Across 6,000 calibration draws with the selected rules fixed, the product term is $0.00992159$ and its covariance correction is $+0.00001291$. The correction is small on average, but its absolute ratio to the product reaches $22.06\%$ for one model. This supports retaining the dependence term, not setting it to zero.

Figure~\ref{fig:restructured-mechanism} addresses a different covariance. For the risk-prediction control on a fixed split, let $m_u$ be SCoRE's user decision and $\bar m$ its average, with $p$ our common probability. The gain is $(p-\bar m)\overline g_T-\operatorname{Cov}_T(m,g)$. Here covariance is across test users, unlike the variation across calibration sets in~\eqref{eq:increment}. The activation term $0.017707$ and selection correction $-0.000778$ yield $0.016928$ for that risk-prediction control after rounding. PerK proposals reverse the balance, showing the value of user selection in that setting.

Before final calibration, $S$ reveals whether candidate gains and recoverable activation coexist. Music has little gain and almost saturated raw activation. KuaiRand has headroom but candidate Recall downside about $5.47$ times its budget. The other domains offer more useful activation. Appendix~\ref{app:prospective-diagnostics} reports these descriptive signals, without treating selected-$S$ estimates as certificates. Removing KuaiRec still leaves a five-domain mean increment of $0.005143$.

The controlled diagnostic in Appendix~\ref{app:asymptotic-check} connects the rates to finite samples: at $n=200$, relative errors in the RB and raw leading terms are $2.62\%$ and $34.18\%$. These coefficients are not exact small-sample predictions. Figure~\ref{fig:mechanism} separately shows loss-coarsening and saturation effects. Appendix~\ref{app:split-risk-distribution} reports realized-risk variability.

\subsection{Frozen confirmation and computational cost}\label{sec:jester}
With the method and primary endpoint frozen before downloading Jester, the same-rule, same-filter increment over raw on all eligible users is $0.011025\,{\scriptstyle\pm\,0.001082}$, with positive means for all fifteen backbone/seed combinations. Without filtering it remains $0.011040$. Figure~\ref{fig:jester-module-evidence} separates these calibration gains from the smaller $0.001949$ filter increment at fixed $w_+$. The primary interval exceeds the prespecified research margin $0.005$, which is a screening threshold on normalized net utility, not a validated business threshold. Actual risks are $0.014512/0.007563$, compared with $0.012431/0.006466$ for filtered raw.

This confirmation uses all eligible users, distinct from the later shared-support module comparison. Appendix~\ref{app:jester} reports all external metrics. Appendix~\ref{app:fresh-confirmation} retains the frozen MovieLens temporal confirmation, including an empirical risk exceedance. Given frozen outputs, exact $C200$ calibration of all three operators averages 11.264 CPU milliseconds. Further costs appear in Appendix~\ref{app:mechanisms-complete}.

\section{Discussion and Conclusion}
ACRC provides more efficient valid scalar calibration for frozen proposals. Its risk guarantee requires the stated sampling conditions and loss bounds, not our candidate templates. Utility gains additionally depend on proposal quality and available activation. Same-rule controls and frozen confirmation isolate calibration, while external comparisons evaluate the full recommendation instance. Evidence remains offline on observed pools under a declared net-utility valuation. It does not establish gains across arbitrary generators, full catalogs, or future traffic. The guarantee concerns expected risk, without protection for every user or realized calibration set.

\balance
\begingroup\small\raggedright\endgroup

\clearpage\onecolumn\appendix\raggedbottom
\renewcommand{\theequation}{EC.\arabic{equation}}
\renewcommand{\theHequation}{EC.\arabic{equation}}
\setcounter{equation}{0}
\section{Complete Theory}\label{app:theory-complete}
The following arguments establish validity, operator efficiency and utility transfer under the stated sampling protocols.
\subsection{Complete Proofs}\label{app:proofs}
All losses in the scalar results lie in $[0,1]$, $n\ge1$, and $0<\alpha<1$. The validity class requires every exchangeable joint law. The iid condition is used only where efficiency under a distribution is explicitly evaluated.

\subsubsection{Fractional-loss improvement}
The improvement has an exact boundary. Define $s=\sum_i l_i$ and $V=\sum_i l_i(1-l_i)$. Here $V$ is the auxiliary Bernoulli variance conditional on the losses, not the empirical variance of user losses.
\begin{proposition}\label{prop:fractional}
The inequality $w_+(l)>w_0(l)$ holds exactly when $1+s>A$ and $V>0$. In that case

\begin{equation}
 0<w_+(l)-w_0(l)
 \le\min\left\{1-w_0(l),\frac{AV}{(1+s)^2}\right\}.\label{eq:cap}
\end{equation}
\end{proposition}

\begin{proof}[Proof of Proposition~\ref{prop:fractional}]
Conditionally, $K_B$ has mean $s$ and variance $V$. For $x\ge0$, the exact reciprocal identity at $s$ gives

\begin{equation*}
\frac{1}{1+x}
=\frac{1}{1+s}-\frac{x-s}{(1+s)^2}
 +\frac{(x-s)^2}{(1+s)^2(1+x)}.
\end{equation*}
\begin{equation*}
\E\frac{1}{1+K_B}-\frac{1}{1+s}
 =\frac{1}{(1+s)^2}\E\frac{(K_B-s)^2}{1+K_B}
 \le\frac{V}{(1+s)^2}.
\end{equation*}

If $V>0$, strict convexity makes the difference strictly positive. If $V=0$, the auxiliary count is constant and all operators agree. When raw is saturated, Jensen's inequality implies that the new operator is also one. Otherwise multiplying the display by $A$ and clipping yields~\eqref{eq:cap}.
\end{proof}

\subsubsection{Admissibility}
Let $\mathcal W_{n,\alpha}$ contain all measurable scalar probabilities that are valid for every exchangeable joint loss distribution on $[0,1]^N$. This class does not include arbitrary input-adaptive recommendation systems. Efficiency below is evaluated under iid losses with any marginal law $F$ on $[0,1]$, while validity is required for all exchangeable distributions.
\begin{theorem}\label{thm:minimax}
No member of $\mathcal W_{n,\alpha}$ dominates $w_+$ pointwise with a strict improvement somewhere. Let $r_F=\E_FL$, $w^\star(r)=\min\{1,\alpha/r\}$, and $w^\star(0)=1$. Then

\begin{align}
 \inf_{w\in\mathcal W_{n,\alpha}}\sup_F
 \big[w^\star(r_F)-\E_{F^n}w\big]
 &=\frac{\E|Y/N-\alpha|}{2\alpha},\label{eq:minimax}\\
 Y&\sim\mathrm{Binomial}(N,\alpha).\notag
\end{align}
\end{theorem}

Both $w_+$ and $w_{\rm RB}$ attain this value, with Bernoulli losses as a worst case. This shared result characterizes the class rather than giving $w_+$ a unique minimax advantage.

\begin{proof}[Pointwise part of Theorem~\ref{thm:minimax}]
First consider symmetric $w$. A value $w_+(l)=1$ cannot increase. If $a(l)<1$, replacing a coordinate by one can only decrease $a$. Thus all leave-one-out terms for the augmented vector $(l,1)$ are untruncated. Equation~\eqref{eq:identity} becomes

\begin{equation*}
w_+(l)+\sum_{i=1}^n l_iw_+(l_{-i},1)=A.
\end{equation*}

The uniform permutation law of this vector is exchangeable. Validity of $w$ requires the same expression with $w$ in place of $w_+$ to be at most $A$. Pointwise dominance and a strict increase at $l$ contradict this constraint. For nonsymmetric $w$, average over all calibration-coordinate permutations. Exchangeable validity, dominance, and a strict increase somewhere are preserved, reducing the claim to the symmetric case.
\end{proof}

\subsubsection{The exact minimax value}
Let $c_k=\min\{1,A/(k+1)\}$ and $g_n(r)=\E c_K$ for $K\sim\mathrm{Binomial}(n,r)$. Integrating auxiliary Bernoulli variables under iid losses with mean $r$ gives $\E_{F^n}w_{\rm RB}=g_n(r)$. The new operator therefore has expectation at least $g_n(r)$, with equality on Bernoulli loss distributions.

\begin{proof}[Minimax part of Theorem~\ref{thm:minimax}]
Symmetrize any valid $w$ without changing its iid performance. Let $b_k$ be its value on a binary calibration vector with $k$ ones. On an augmented uniform permutation orbit with $k+1$ ones, exchangeable validity forces

\begin{equation*}
\frac{k+1}{N}b_k\le\alpha,\qquad b_k\le1,\qquad b_k\le c_k.
\end{equation*}

Consequently no valid operator has mean activation above $g_n(\alpha)$ on iid Bernoulli losses with mean $\alpha$. This proves the minimax lower bound.

For the upper bound, binomial size bias gives

\begin{equation*}
 r g_n(r)=\E\min\{Y/N,\alpha\},\qquad Y\sim\mathrm{Binomial}(N,r).
\end{equation*}
\begin{equation*}
w^\star(r)-g_n(r)=
\begin{cases}
r^{-1}\E(Y/N-\alpha)_+,&0<r\le\alpha,\\
r^{-1}\E(\alpha-Y/N)_+,&r\ge\alpha.
\end{cases}
\end{equation*}

For $r\le\alpha$, $1-g_n(r)$ increases with $r$. For $r\ge\alpha$, couple binomial counts through common uniforms. The numerator in the second branch decreases with $r$, and the denominator increases. Both branches are therefore maximized at $\alpha$. At $r=0$, directly use $g_n(0)=\min\{1,A\}$. Since $\E w_+\ge g_n(r)$ and iid validity implies $\E w_+\le w^\star(r)$ for $r>0$, the maximum deficit is $1-g_n(\alpha)$. At $r=\alpha$, the centered binomial has mean zero, so

\begin{equation*}
1-g_n(\alpha)
=\frac{\E(Y/N-\alpha)_+}{\alpha}
=\frac{\E|Y/N-\alpha|}{2\alpha}.
\end{equation*}
\end{proof}

The same calculation proves that RB attains the value. The argument requires validity for every exchangeable law, not merely every iid law. It also uses the full normalized loss range. A smaller known range changes the comparison class.

\subsubsection{Physical risk under sampling without replacement}
\begin{proof}[Proof of Theorem~\ref{thm:dual-risk}]
Choose a uniform evaluation user from the complement of a uniform $n$-user calibration subset. Equivalently, draw $n+1$ users uniformly and hold one out uniformly. Theorem~\ref{thm:valid} applies to each normalized loss on that finite array. The common probability is no larger than its corresponding gate, and downward quantization can only reduce it. Multiply by the legal action range to recover the physical budget. If the range is already below the budget, activation one is valid pointwise. Averaging the internal action coin first yields the actual downside in~\eqref{eq:actual}. Conditioning on $S$ preserves the remaining-pool sampling because selection never reads the final labels.
\end{proof}

\subsubsection{Incremental utility with randomized actions}
For the same selected rule and the same ranges, let $p_0$ replace each $w_+$ in~\eqref{eq:dual} by $w_0$. Define $d(C)=p_+(C)-p_0(C)\ge0$, $\tau=\E_Cd(C)$, $\mu=\overline g_R$, and $\Delta D=\E_C[d(C)\overline g_{R\setminus C}]$. Let $v_g=\operatorname{Var}_R(g)$ and $v_d=\operatorname{Var}_C(d(C))$. For $u\ne v$, take $A_{uv}$ uniformly among size $n-1$ subsets excluding $u,v$, and define

\begin{align}
 \omega_{uv}&=\E_{A_{uv}}|d(A_{uv}\cup\{v\})-d(A_{uv}\cup\{u\})|,\notag\\
 B_\Delta&=\frac{1}{2M^2}\sum_{u\ne v}|g_u-g_v|\omega_{uv}.\label{eq:swap}
\end{align}

\begin{theorem}\label{thm:increment}
Under the fixed-pool protocol, including randomized candidate actions and two constraints,

\begin{align}
 \Delta D&=\mu\tau+\operatorname{Cov}_C(d(C),\overline g_{R\setminus C}),\notag\\
 |\Delta D-\mu\tau|&\le
 \min\left\{\frac{nB_\Delta}{M-1},
 \sqrt{\frac{n v_gv_d}{(M-n)(M-1)}}\right\}.\label{eq:increment-bound}
\end{align}
\end{theorem}

A product $\mu\tau$ larger than either correction bound is sufficient for a positive expected increment. If all $g_u\ge0$, the increment is nonnegative for every split. These are fixed-population targets, not automatically observable from $S$. They support mechanism analysis and do not constitute an implemented positive-utility certificate.

\begin{proof}[Proof of Theorem~\ref{thm:increment}]
Uniform sampling gives $\E_C\overline g_{R\setminus C}=\mu$, proving~\eqref{eq:increment}. Let $t_u^d=\E_C[d(C)\mid u\notin C]$. Then

\begin{equation*}
\Delta D-\mu\tau
=\frac{1}{2M^2}\sum_{u,v}(g_u-g_v)(t_u^d-t_v^d).
\end{equation*}

Couple the uniform subsets excluding $u$ and $v$. They coincide except with probability $n/(M-1)$, when their common part $A_{uv}$ has size $n-1$ and the other user is exchanged. This gives

\begin{equation*}
t_u^d-t_v^d=\frac{n}{M-1}
\E_{A_{uv}}[d(A_{uv}\cup\{v\})-d(A_{uv}\cup\{u\})].
\end{equation*}

Taking absolute values proves the first bound. The finite-population variance of the complementary mean is

\begin{equation*}
\operatorname{Var}_C(\overline g_{R\setminus C})
=\frac{n}{(M-n)(M-1)}v_g.
\end{equation*}

Cauchy--Schwarz applied to the covariance proves the second bound. No independence of graph users, deterministic action assumption, or covariance sign was used.
\end{proof}

For a counterexample to a utility guarantee from activation alone, take two users with $n=1$, utility budget $0.02$, range one, and an inactive Recall constraint. For the first user, the candidate chooses gains $0.9$ and $-0.8$ equally. For the second, both gains are $-0.01$. Hence $(g_1,x_1)=(0.05,0.4)$ and $(g_2,x_2)=(-0.01,0.01)$, with positive population mean gain $0.02$. The extra probabilities are

\begin{equation*}
d_1=0.04\left(0.8-\frac1{1.4}\right),\qquad
d_2=0.04\left(0.995-\frac1{1.01}\right).
\end{equation*}
\begin{equation*}
\Delta D=\frac{-0.01d_1+0.05d_2}{2}<0.
\end{equation*}

The 53-bit quantized difference remains negative. This is a mathematical boundary example for the bounded randomized-action class, not a recommendation performance experiment.

\subsubsection{A sharp calibration increment}
\begin{proposition}\label{prop:sharp-gap}
Let $v_i=l_i(1-l_i)$ and $J=\E(1+K_B)^{-1}-(1+s)^{-1}$. Define
\begin{align}
 J_-&=\frac1{1+s}\sum_i\frac{v_i}{(1+s-l_i)(2+s-l_i)},\notag\\
 J_+&=\frac1{1+s}\sum_i v_i\min\{1/2,(s-l_i)^{-2}\}.\label{eq:sharp-j}
\end{align}
The inverse square is $+\infty$ at zero. Then $J_-\le J\le J_+$ and
\begin{equation}
 w_+-w_0=\min\{1-w_0,AJ\}.\label{eq:exact-gap}
\end{equation}
For $s>1$ and $V>0$,
\begin{equation}
 \frac{V}{(1+s)^2(s+2)}\le J\le\frac{V}{(1+s)(s-1)^2}.
\end{equation}
Consequently $J=(V/s^3)[1+O(1/s)]$ uniformly as $s\to\infty$. In finite samples one may also take the minimum of $J_+$ and the previous bound $V/(1+s)^2$.
\end{proposition}

\begin{proof}[Proof of Proposition~\ref{prop:sharp-gap}]
For any function $f$ on the nonnegative integers, independence of the auxiliary Bernoulli variables gives
\begin{equation*}
 \E[(K_B-s)f(K_B)]=\sum_i v_i\E[f(K_{-i}+1)-f(K_{-i})].
\end{equation*}
Set $f(k)=(1+k)^{-1}$. Since $\E[(1+K_B)f(K_B)]=1$,
\begin{equation}
 J=\frac1{1+s}\sum_i v_i\E\frac1{(1+K_{-i})(2+K_{-i})}.\label{eq:stein-gap}
\end{equation}
The function on the right is convex on $[0,\infty)$, giving $J_-$. Its expectation equals
\begin{equation*}
 \int_0^1 t\prod_{j\ne i}(1-tl_j)\,dt
 \le\int_0^1 t e^{-(s-l_i)t}\,dt
 \le\min\{1/2,(s-l_i)^{-2}\}.
\end{equation*}
This gives $J_+$. If raw is saturated both gates equal one. Otherwise subtract raw before clipping to obtain~\eqref{eq:exact-gap}. For $s>1$, use $s-1\le s-l_i\le s$ in the two bounds. Dividing by $V/s^3$ proves the uniform expansion. The identity uses standard Bernoulli covariance calculus. The contribution here is its application to an explicit, quantitative calibration increment.
\end{proof}

For each active constraint set $b_k=w_{0,k}$ and
\begin{equation*}
 \delta_{k,-}=\min\{1-b_k,A_kJ_{k,-}\},\qquad
 \delta_{k,+}=\min\{1-b_k,A_kJ_{k,+}\}.
\end{equation*}
For an inactive constraint use $b_k=1$ and $\delta_{k,-}=\delta_{k,+}=0$. Let $b=\min_k b_k$. Before quantization the exact increment is
\begin{equation*}
 d=\min_k\{b_k-b+\delta_k\},\qquad \delta_k=w_{+,k}-w_{0,k}.
\end{equation*}
Hence, with $d_\pm=\min_k\{b_k-b+\delta_{k,\pm}\}$,
\begin{equation}
 \max\{0,d_- -q\}\le d_q\le\min\{1,d_++q\}.\label{eq:dual-sharp}
\end{equation}
The same minimum identity holds for comparator $c\in\{0,{\rm RB}\}$ with $b_k=w_{c,k}$ and $\delta_k=w_{+,k}-b_k$. Strict improvement requires and is implied by $\delta_k>0$ for every previously binding constraint with $b_k=b$. Downward quantization preserves weak dominance, and an unquantized increment above $2^{-53}$ is sufficient for strict improvement after execution. Multiplying the bounds by $\overline g_T$ gives retrospective utility bounds, with endpoints reversed when $\overline g_T<0$. These bounds are not certificates derived only from $S$.

\paragraph{Strict improvement over RB.}
\begin{proposition}\label{prop:rb-strict-regime}
Let $X=A/(1+K_B)$, $P=\E(X-1)_+$, and $Q=\E(1-X)_+$. For every loss vector,
\begin{equation}
 w_+-w_{\rm RB}=\min\{P,Q\}.
\end{equation}
Let $b=\#\{i:l_i=1\}$ and $r=\#\{i:0<l_i<1\}$. Then
\begin{equation}
 w_+>w_{\rm RB}\quad\Longleftrightarrow\quad 1+b<A<1+b+r.
\end{equation}
For $r>0$, write $c=A-1$, $u=[A/(1+b)-1]_+$, and $v=[1-A/(1+b+r)]_+$. Moreover,
\begin{align}
 w_+-w_{\rm RB}\le\min\big\{&u\exp[-2(s-c)_+^2/r],\notag\\
 &v\exp[-2(c-s)_+^2/r]\big\}.
\end{align}
The difference is zero when $r=0$.
\end{proposition}
\begin{proof}
The exact identity follows from $\E X=1+P-Q$ and $\E\min\{1,X\}=1-Q$. The support of $K_B$ contains every integer from $b$ through $b+r$, each with positive probability. Thus $P$ and $Q$ are both positive exactly when this support straddles $A-1$. Finally, $P\le u\Pr(K_B<c)$ and $Q\le v\Pr(K_B>c)$. Apply Hoeffding's inequality to the sum of the $r$ independent bounded Bernoulli variables, using the probability bound one when the threshold lies on the other side of the mean.
\end{proof}
The result characterizes the RB advantage region for heterogeneous losses. It guarantees neither positive candidate utility nor exponential utility improvement for arbitrary loss distributions.

\paragraph{A tighter upper bound.}

The upper bound in Proposition~\ref{prop:sharp-gap} can also retain the exact exponential integral. Let $h(0)=1/2$ and $h(t)=[1-(1+t)e^{-t}]/t^2$ for $t>0$. Then
\begin{equation}
 J\le\frac{1}{1+s}\sum_i v_i h(s-l_i).
\end{equation}
The proof uses the same integral following Equation~\eqref{eq:stein-gap}. Since $h(t)=\int_0^1 u e^{-tu}\,du\le\min\{1/2,t^{-2}\}$, this is no weaker than the previous integral bound. It can still be intersected with $V/(1+s)^2$.

\subsubsection{Constant-loss formulas}
The zero-variance cases below use the exact formulas for $l_1=\cdots=l_n=r\in(0,1)$:

\begin{align*}
w_+&=\min\!\left\{1,\frac{\alpha}{r}\big[1-(1-r)^{n+1}\big]\right\},\\
w_0&=\min\!\left\{1,\frac{\alpha(n+1)}{1+nr}\right\},\\
w_{\rm RB}&=\E_{K\sim\operatorname{Binomial}(n,r)}
\min\!\left\{1,\frac{\alpha(n+1)}{1+K}\right\}.
\end{align*}
At $r=\alpha$, these formulas imply $n(w_+-w_0)\to(1-\alpha)/\alpha$.

\subsection{Calibration Efficiency and Utility for Heterogeneous Users}\label{app:heterogeneous-efficiency}

\subsubsection{Efficiency and utility with heterogeneous losses}
For heterogeneous losses, the efficiency comparison depends on actual user variation and auxiliary Bernoulli variation. For a loss with mean $r\in(0,1)$, write
\begin{equation*}
 \sigma^2=\operatorname{Var}(L),\qquad
 \nu=\mathbb E[L(1-L)]=r(1-r)-\sigma^2.
\end{equation*}
Here $\sigma^2$ is the variance of actual user losses, whereas $\nu$ is the
additional variance introduced by auxiliary Bernoulli randomization.
Moreover, $\nu>0$ exactly when the loss has positive probability in $(0,1)$.

\begin{theorem}[Efficiency under heterogeneous iid losses]\label{thm:heterogeneous-iid}
Let the calibration losses be iid from a fixed distribution with $\sigma>0$
and $r=\alpha\in(0,1)$. Define $D_c=1-\mathbb E w_c$. Then
\begin{align}
 \sqrt n D_+\longrightarrow
 \frac{\sigma}{\alpha\sqrt{2\pi}},\qquad
 \sqrt n D_0\longrightarrow
 \frac{\sigma}{\alpha\sqrt{2\pi}},\label{eq:heterogeneous-iid-new}\\
 \sqrt n D_{\rm RB}\longrightarrow
 \frac{\sqrt{\alpha(1-\alpha)}}{\alpha\sqrt{2\pi}},\label{eq:heterogeneous-iid-rb}\\
 n\mathbb E(w_+-w_0)\longrightarrow
 \frac{\nu}{2\alpha^2}.\label{eq:heterogeneous-iid-raw}
\end{align}
Consequently, when $\nu>0$,
\begin{equation*}
 \mathbb E(w_+-w_{\rm RB})
 =\frac{\sqrt{\alpha(1-\alpha)}-\sigma}
 {\alpha\sqrt{2\pi n}}+o(n^{-1/2}),
\end{equation*}
while the improvement over raw is $\nu/(2\alpha^2n)+o(n^{-1})$.
These statements concern the real-valued operators before quantization.
\end{theorem}

This result identifies a nondegenerate distribution class with strict efficiency
gains. User losses need not be identical, and their variance need not vanish.
The improvement over RB removes the effect of auxiliary randomization on the
clipping loss. The improvement over raw is a smaller-order finite-sample
correction. These comparisons have different orders and should not be combined
into a single rate claim. If $\sigma=0$ and $L\equiv\alpha$, the existing exact
formulas instead give
$n\mathbb E(w_+-w_0)\to(1-\alpha)/\alpha=\nu/\alpha^2$.
Substituting $\sigma=0$ into \eqref{eq:heterogeneous-iid-raw} is invalid.

\paragraph{Budget position and finite samples.}
At $r=\alpha$, without requiring $\sigma>0$,
\begin{equation}
 0\le1-\mathbb E w_+
 \le1-\mathbb E w_0
 \le\min\left\{1,
 \frac{\sigma\sqrt n/2+1-\alpha}{\alpha(n+1)}\right\}.
 \label{eq:heterogeneous-finite-bound}
\end{equation}
When the mean is strictly below the budget, all three operators approach full
activation exponentially. When the mean is strictly above the budget, the iid
operators $w_+$ and RB approach the common scalar benchmark
$w^\star=\alpha/r$ exponentially, whereas
\begin{equation}
 n\{w^\star-\mathbb E w_0\}\longrightarrow
 \frac{\alpha\nu}{r^3},\qquad
 n\mathbb E(w_+-w_0)\longrightarrow\frac{\alpha\nu}{r^3}.
 \label{eq:heterogeneous-above}
\end{equation}
The exponential constants may depend on the fixed mean, budget, and distribution.
The assertion is not uniform as the budget gap vanishes. Thus the main opportunity
to improve RB is near the budget boundary, whereas the correction to raw can
persist above that boundary.

More precisely, fix the loss distribution and its mean $r$, and set
$\alpha_n=r+h/\sqrt n$ for fixed $h\in\mathbb R$. For $Z\sim N(0,1)$ define
\begin{equation*}
 H_s(h)=\mathbb E(sZ-h)_+
 =s\phi(h/s)-h\overline\Phi(h/s),\qquad s>0,
\end{equation*}
and $H_0(h)=(-h)_+$. Then
\begin{align}
 \sqrt n\mathbb E(w_+-w_{\rm RB})
 &\longrightarrow
 \frac{H_{\sqrt{r(1-r)}}(h)-H_\sigma(h)}{r},\label{eq:heterogeneous-local-rb}\\
 n\mathbb E(w_+-w_0)
 &\longrightarrow\frac{\nu}{r^2}\overline\Phi(h/\sigma),
 \qquad\sigma>0.\label{eq:heterogeneous-local-raw}
\end{align}
For $\nu>0$, the coefficient in \eqref{eq:heterogeneous-local-rb} is positive,
symmetric in $h$, maximized at zero, and tends to zero as $|h|$ increases.
For a gap to the scalar benchmark $w_n^\star=\min\{1,\alpha_n/r\}$, the coefficient for $w_+$ is instead
$\{H_\sigma(h)-(-h)_+\}/r$. The second term cannot be omitted.

\begin{theorem}[Efficiency and utility in a fixed user pool]\label{thm:heterogeneous-finite-pool}
Consider a sequence of fixed pools of size $M$ with losses $l_u\in[0,1]$,
mean exactly $\alpha\in(0,1)$, and population variance
$\sigma_M^2=M^{-1}\sum_u(l_u-\alpha)^2\to\sigma^2>0$.
Draw $n$ users uniformly without replacement, with $n\to\infty$ and
$n/M\to\rho\in[0,1)$. Set $\nu=\alpha(1-\alpha)-\sigma^2$. Then
\begin{align}
 \sqrt n(1-\mathbb E_Cw_+),\ 
 \sqrt n(1-\mathbb E_Cw_0)
 &\longrightarrow c_+
 =\frac{\sqrt{1-\rho}\,\sigma}{\alpha\sqrt{2\pi}},\label{eq:heterogeneous-pool-new}\\
 \sqrt n(1-\mathbb E_Cw_{\rm RB})
 &\longrightarrow c_{\rm RB}
 =\frac{\sqrt{\alpha(1-\alpha)-\rho\sigma^2}}
 {\alpha\sqrt{2\pi}},\label{eq:heterogeneous-pool-rb}\\
 n\mathbb E_C(w_+-w_0)&\longrightarrow\frac{\nu}{2\alpha^2}.
 \label{eq:heterogeneous-pool-raw}
\end{align}
Suppose also that the fixed rule has bounded user-specific expected gains
$|g_u|\le B_g$ and $\mu_M=M^{-1}\sum_ug_u\to\mu>0$.
For $\Delta D_c=\mathbb E_C[(w_+-w_c)\overline g_{R\setminus C}]$,
\begin{align}
 \sqrt n\Delta D_{\rm RB}&\longrightarrow\mu(c_{\rm RB}-c_+),\label{eq:heterogeneous-pool-utility-rb}\\
 n\Delta D_0&\longrightarrow\frac{\mu\nu}{2\alpha^2}.
 \label{eq:heterogeneous-pool-utility-raw}
\end{align}
Gains need not be constant or independent of losses. If $\nu>0$, both expected
utility improvements are eventually strictly positive along this sequence.
\end{theorem}

The randomness here comes from actual sampling without replacement, not iid graph
users. Bounded losses, a nonzero limiting variance, and $\rho<1$ ensure the finite
population central limit theorem \cite{lidingsampling2017}. RB retains both
sampling variation and auxiliary randomization, so its iid binomial constant
cannot be reused. The utility statement controls the covariance with the
complementary test set at an order smaller than the activation improvement.
The result concerns one constraint at the boundary. It also applies to the
minimum activation probability when exactly one constraint is at its boundary
and every other constraint has a fixed strict margin below its budget.
When several constraints are simultaneously at their boundaries, their joint
fluctuations and bottleneck switches matter. Theorem~\ref{thm:heterogeneous-multigate}
provides this joint analysis. Applying the single-constraint constants separately
and then taking their minimum would not preserve these effects.

\paragraph{A local budget distance in a fixed pool.}
Exact equality to the budget is not essential. Let the population mean be $r_M$
with $\sqrt n(\alpha-r_M)\to h\in\mathbb R$, defining each population variance around its own mean and keeping the other assumptions.
Set $s_C=\sqrt{1-\rho}\,\sigma$ and
$s_K=\sqrt{\alpha(1-\alpha)-\rho\sigma^2}$. Then
\begin{align}
 \sqrt n\mathbb E_C(w_+-w_{\rm RB})
 &\longrightarrow\{H_{s_K}(h)-H_{s_C}(h)\}/\alpha,\label{eq:heterogeneous-pool-local-rb}\\
 n\mathbb E_C(w_+-w_0)
 &\longrightarrow\frac{\nu}{\alpha^2}\overline\Phi(h/s_C).\label{eq:heterogeneous-pool-local-raw}
\end{align}
The corresponding utility limits remain these coefficients multiplied by $\mu$.
This compares increments between calibrators. It does not call full activation
feasible when the population mean is slightly above the budget.

\paragraph{Above the budget in a fixed pool.}
Under the same sampling protocol, if the population mean is fixed at $r>\alpha$
and its variance converges to $\sigma^2$, redefine $\nu=r(1-r)-\sigma^2$
for this paragraph. Comparison to the fixed scalar probability $\alpha/r$ yields
\begin{align}
 n\{\alpha/r-\mathbb E_Cw_+\},\ 
 n\{\alpha/r-\mathbb E_Cw_{\rm RB}\}
 &\longrightarrow\frac{\alpha\rho\sigma^2}{r^3},\label{eq:heterogeneous-pool-above}\\
 n\{\alpha/r-\mathbb E_Cw_0\}
 &\longrightarrow\frac{\alpha\{\nu+\rho\sigma^2\}}{r^3}.
\end{align}
Thus the iid exponential benchmark gap does not extend to finite pools with
$\rho>0$ and $\sigma^2>0$. This benchmark is not asserted to be optimal among all
finite-pool decisions that depend on the calibration set.

\subsubsection{Proofs of the heterogeneous efficiency results}
\paragraph{Uniform increment control and second moments.}
Write $S_n=\sum_iL_i$, $V_n=\sum_iL_i(1-L_i)$, and
$A_n=\alpha_n(n+1)$. The existing deterministic bounds imply
\begin{equation*}
 J_n=\mathbb E_{B\mid L}(1+K_B)^{-1}-(1+S_n)^{-1}
 =\frac{V_n}{S_n^3}\{1+O(S_n^{-1})\},
\end{equation*}
with a relative remainder uniform over all loss vectors with $S_n>1$.
If $V_n=0$, then $J_n=0$ and the relative expression is unnecessary.
For a fixed positive iid or population mean $r$, the laws of large numbers give
$S_n/n\to r$ and $V_n/n\to\nu$. Hence, if $\alpha_n\to\alpha$,
\begin{equation}
 nA_nJ_n\longrightarrow\frac{\alpha\nu}{r^3}
 \quad\text{in probability}.\label{eq:heterogeneous-j-limit}
\end{equation}
On $S_n\ge nr/2$, the deterministic upper bound makes $nA_nJ_n$ uniformly bounded.
The complementary event has probability at most $\exp(-nr^2/2)$ by Hoeffding's
bound for iid or sampling-without-replacement averages. On that event
$A_nJ_n\le A_n=O(n)$. Thus the first and second moments of $nA_nJ_n$ are uniformly
integrable. In particular, $d_0=w_+-w_0\le A_nJ_n$ satisfies
\begin{equation}
 \mathbb E d_0^2=O(n^{-2}).\label{eq:heterogeneous-raw-l2}
\end{equation}

\paragraph{The iid boundary and local budgets.}
Let $\sigma>0$ and $\alpha_n=r+h/\sqrt n$. The central limit theorem gives
$(S_n-nr)/\sqrt n\Rightarrow\sigma Z$, while the exact raw deficit is
\begin{equation*}
 1-w_0=
 \frac{\{S_n-nr-h\sqrt n+1-\alpha_n\}_+}{1+S_n}.
\end{equation*}
On $S_n\ge nr/2$, the denominator is bounded below and the normalized centered
sum has bounded second moment. The complementary event is exponentially rare.
Uniform integrability therefore gives
\begin{equation*}
 \sqrt n\mathbb E(1-w_0)\to H_\sigma(h)/r.
\end{equation*}
Equation \eqref{eq:heterogeneous-raw-l2} gives the same limit for $w_+$.
After integrating the auxiliary variables, iid losses give exactly
$K_B\sim\operatorname{Binomial}(n,r)$, so the same argument gives the RB limit
$H_{\sqrt{r(1-r)}}(h)/r$. Subtracting proves
\eqref{eq:heterogeneous-local-rb}.

For the smaller raw increment, use
\begin{equation*}
 n(w_+-w_0)=\min\{n(1-w_0),nA_nJ_n\}.
\end{equation*}
When the limiting variable $\sigma Z-h$ is positive, the first argument diverges.
When it is negative, the first argument is eventually zero. The limiting variable
has no atom at zero. Equation \eqref{eq:heterogeneous-j-limit} and uniform
integrability imply
\begin{equation*}
 n\mathbb E(w_+-w_0)\to(\nu/r^2)\Pr(\sigma Z>h).
\end{equation*}
One can formalize this step by restricting to the two regions separated from zero
by $\varepsilon$ and then letting $\varepsilon\downarrow0$. This proves
\eqref{eq:heterogeneous-local-raw}. Taking $h=0$ proves
Theorem~\ref{thm:heterogeneous-iid}. Moreover,
$\partial H_s(h)/\partial s=\phi(h/s)>0$ for $s>0$, and
$H_s(-h)=H_s(h)+h$. Their difference is therefore positive and symmetric,
with negative derivative for $h>0$. Gaussian tails make it vanish as
$|h|\to\infty$.
For $\sigma=0$, the first-order local-budget formula follows directly from the
exact constant-loss expression. The second-order raw formula still requires
$\sigma>0$.

\paragraph{A finite-sample boundary bound.}
At $r=\alpha$, a positive raw deficit implies $1+S_n>A_n$. Since
$\mathbb E(S_n-n\alpha)=0$,
\begin{equation*}
 \mathbb E(S_n-n\alpha)_+
 =\tfrac12\mathbb E|S_n-n\alpha|
 \le\tfrac12\sqrt{\operatorname{Var}(S_n)}.
\end{equation*}
Consequently,
\begin{equation*}
 1-\mathbb E w_+\le1-\mathbb E w_0
 \le\frac{\sqrt{\operatorname{Var}(S_n)}/2+1-\alpha}{\alpha(n+1)}.
\end{equation*}
Use $\operatorname{Var}(S_n)=n\sigma^2$ in the iid case and
$n(M-n)\sigma_M^2/(M-1)$ in a fixed pool.

\paragraph{Finite-sample bounds away from the budget under iid sampling.}
If $r<\alpha$, let $t_n=\alpha-r-(1-\alpha)/n>0$. Raw can be unsaturated only
when $S_n/n>r+t_n$, so Bernstein's inequality yields
\begin{equation*}
 1-\mathbb E w_+\le1-\mathbb E w_0
 \le\exp\left\{-\frac{nt_n^2}{2\sigma^2+2t_n/3}\right\}.
\end{equation*}
For RB, replace $\sigma^2$ by $r(1-r)$. If $r>\alpha$, independence gives
the exact untruncated mean
\begin{equation*}
 \mathbb E a(L)=\frac\alpha r\{1-(1-r)^{n+1}\}.
\end{equation*}
The bound $a(L)\le A_n/S_n$ for $S_n>0$ implies that $a>1$ requires
$S_n<A_n$. Setting $u_n=r-\alpha-\alpha/n>0$ gives
\begin{equation*}
 0\le\frac\alpha r-\mathbb E w_+
 \le\frac\alpha r(1-r)^{n+1}
 +\alpha(n+1)\exp\left\{-\frac{nu_n^2}{2\sigma^2+2u_n/3}\right\}.
\end{equation*}
The first term is the integral-endpoint correction and the second controls final
clipping. The exact binomial formula gives an exponential RB bound. Alternatively,
replace $\sigma^2$ by $r(1-r)$ in the last bound and keep the endpoint term.
At a fixed strict distance from the budget, clipping is exponentially rare and
\eqref{eq:heterogeneous-j-limit} passes to expectations, proving
\eqref{eq:heterogeneous-above}. If a finite-sample upper bound exceeds one,
it can be intersected with the trivial bound one.

\paragraph{The two sources of variation in a fixed pool.}
For pools with mean $\alpha$, the bounded finite population central limit theorem
\cite{lidingsampling2017} gives
\begin{equation*}
 \frac{S_n-n\alpha}{\sqrt n}\Rightarrow
 N(0,(1-\rho)\sigma^2).
\end{equation*}
Its condition holds because the maximum squared centered loss is at most one,
the variance tends to a positive number, and $\min(n,M-n)\to\infty$.
The raw positive-part and uniform-integrability argument proves
\eqref{eq:heterogeneous-pool-new}. Equation \eqref{eq:heterogeneous-j-limit}
and the absence of an atom at zero prove \eqref{eq:heterogeneous-pool-raw}.

For RB, conditional on the calibration users, $K_B-S_n$ is a sum of independent
bounded centered variables. For fixed $t$, its conditional characteristic function
satisfies, uniformly over calibration vectors,
\begin{equation*}
 \mathbb E_B e^{it(K_B-S_n)/\sqrt n}
 =\exp\{-t^2V_n/(2n)+O(n^{-1/2})\}.
\end{equation*}
Since $V_n/n\to\nu$, multiplication by the characteristic factor for $S_n$
and averaging yield
\begin{equation*}
 \frac{K_B-n\alpha}{\sqrt n}\Rightarrow
 N(0,\nu+(1-\rho)\sigma^2).
\end{equation*}
When $\nu=0$, the auxiliary centered term tends to zero in $L^2$, giving the same
formula. The law of total variance gives
\begin{equation*}
 \operatorname{Var}(K_B)=n\nu_M+
 \frac{n(M-n)}{M-1}\sigma_M^2=O(n).
\end{equation*}
Positive-part uniform integrability proves \eqref{eq:heterogeneous-pool-rb}.
This is not an iid binomial argument.

\paragraph{Nonconstant user gains.}
The RB increment satisfies $0\le d_{\rm RB}\le1-w_{\rm RB}$.
At the budget boundary, conditional Jensen's inequality gives
\begin{equation*}
 \mathbb E d_{\rm RB}^2
 \le\frac{\mathbb E(K_B-n\alpha+1-\alpha)^2}{\alpha^2(n+1)^2}
 =O(n^{-1}).
\end{equation*}
Together with \eqref{eq:heterogeneous-raw-l2}, the two increments have standard
deviations $O(n^{-1/2})$ and $O(n^{-1})$. Bounded gains satisfy
\begin{equation*}
 \operatorname{Var}_C(\overline g_{R\setminus C})
 =\frac{n v_{g,M}}{(M-n)(M-1)}=O(n^{-1}).
\end{equation*}
Thus the covariance in the existing increment identity is $O(n^{-1})$ for RB
and $O(n^{-3/2})$ for raw. Both are smaller than the corresponding leading terms.
Multiplying by $\mu_M\to\mu$ proves
\eqref{eq:heterogeneous-pool-utility-rb} and
\eqref{eq:heterogeneous-pool-utility-raw}. For $\nu>0$, the difference of square
roots is strictly positive because
$\alpha(1-\alpha)-\rho\sigma^2-(1-\rho)\sigma^2=\nu$.

\paragraph{The local-budget extension in a fixed pool.}
When $\sqrt n(\alpha-r_M)\to h$, the centered sample sum satisfies the same
finite population central limit theorem, while the limit of
$(S_n-n\alpha)/\sqrt n$ becomes $s_CZ-h$.
The auxiliary count similarly has limit $s_KZ-h$.
The local-budget positive-part and clipping argument proves
\eqref{eq:heterogeneous-pool-local-rb} and
\eqref{eq:heterogeneous-pool-local-raw}.
The increment second-moment orders are unchanged because
$\mathbb E(K_B-n\alpha)^2=O(n)$.
The complementary-utility covariance remains smaller-order, preserving utility
transfer.

\paragraph{A unique boundary constraint.}
If every other constraint has a fixed strict margin below its budget,
bounded sampling-without-replacement concentration makes the probability of
an unsaturated raw or new gate exponentially small. The expected RB deficit is
also exponentially small by concentrating the sampled loss sum and then the
conditional auxiliary Bernoulli sum. For finitely many constraints, the expected
difference between their minimum and the boundary gate is exponentially small.
Bounded utility preserves this order, so the leading terms remain unchanged.

\paragraph{A fixed-pool mean above the budget.}
Let the mean be fixed at $r>\alpha$ and
$\nu_M=r(1-r)-\sigma_M^2$. The sample mean has variance
$\sigma_M^2(M-n)/\{n(M-1)\}$ and third absolute centered moment
$O(n^{-3/2})$, the latter following by integrating bounded sampling concentration.
On $S_n\ge nr/2$, a second-order Taylor expansion of $A_n/(1+S_n)$,
with exponentially small exceptional and clipping events, gives
\begin{equation*}
 \mathbb E w_0
 =\frac\alpha r-
 \frac\alpha{nr^3}\{r(1-r)-(1-\rho)\sigma^2\}+o(n^{-1}).
\end{equation*}
Equation \eqref{eq:heterogeneous-j-limit} gives
$n\mathbb E(w_+-w_0)\to\alpha\nu/r^3$, establishing the expansion for $w_+$.
For RB, the total-variance identity yields
$\operatorname{Var}(K_B)/n\to r(1-r)-\rho\sigma^2$.
Conditional Bernoulli concentration and sample-sum concentration provide the
needed third-moment and clipping bounds. The same Taylor argument gives the RB
coefficient $\alpha\rho\sigma^2/r^3$, proving
\eqref{eq:heterogeneous-pool-above}.

\paragraph{Degenerate cases and scope.}
For $r=0$, all losses are zero and all three operators equal
$\min\{1,\alpha(n+1)\}$. For $r=1$, all losses are one and all three equal
$\alpha$. Binary losses have $\nu=0$ and make all three operators equal
pointwise. The fixed-pool boundary theorem assumes positive limiting actual
variance. The condition $\sigma_M^2\to0$ alone does not determine the raw
$n^{-1}$ coefficient, because its rate can affect boundary clipping.
Exactly constant pools use the existing exact formulas.
Fixed 53-bit quantization changes each probability by less than $2^{-53}$,
and changes a difference of two probabilities by less than the same amount.
The asymptotic results concern real-valued operators and do not claim gains
beyond a fixed numerical resolution. They do not compare arbitrary user-adaptive
methods and do not provide a deployment-time positive-utility certificate.

\subsubsection{Several constraints near their budgets}\label{app:heterogeneous-multigate}
A single boundary constraint does not describe bottleneck switching when both
risks approach their budgets. We retain the loss-vector dependence and follow
the implemented order: conditional averaging within each constraint, then the
minimum activation probability.

\begin{theorem}[Heterogeneous calibration and utility with several constraints]\label{thm:heterogeneous-multigate}
Fix the number of constraints $m$ and budgets $\alpha_k\in(0,1)$.
In a sequence of fixed pools of size $M$, let loss vectors $l_u\in[0,1]^m$
have population mean $r_M$ and covariance
\begin{equation*}
 \Sigma_M=M^{-1}\sum_{u\in R_M}(l_u-r_M)(l_u-r_M)^\top\longrightarrow\Sigma.
\end{equation*}
Sample $C$ uniformly without replacement with size $n$ and let $T$ be its complement.
Assume
\begin{equation*}
 n\to\infty,\qquad n/M\to\rho\in[0,1),\qquad
 \sqrt n(\alpha_k-r_{M,k})\to h_k\in\mathbb R.
\end{equation*}
Set $\nu_k=\alpha_k(1-\alpha_k)-\Sigma_{kk}\ge0$.
Apply each operator to the corresponding loss column and budget, and define
$p_c=\min_{1\le k\le m}w_{c,k}$.
For $X\sim N_m(0,(1-\rho)\Sigma)$, define
\begin{align}
 q_k(x)&=(x_k-h_k)_+/\alpha_k,\notag\\
 b_k(x)&=\mathbb E_Z(x_k-h_k+\sqrt{\nu_k}Z)_+/\alpha_k,\qquad Z\sim N(0,1),\notag\\
 Q(x)&=\max_kq_k(x),\qquad B(x)=\max_kb_k(x).
 \label{eq:multigate-functions}
\end{align}
The expectation defining $b_k$ is taken separately for each constraint, not over
the maximum. Then
\begin{align}
 \sqrt n\mathbb E_C(1-p_+),\ \sqrt n\mathbb E_C(1-p_0)
 &\longrightarrow\mathbb E Q(X),\label{eq:multigate-deficit-new}\\
 \sqrt n\mathbb E_C(1-p_{\rm RB})
 &\longrightarrow\mathbb E B(X),\label{eq:multigate-deficit-rb}\\
 \sqrt n\mathbb E_C(p_+-p_{\rm RB})
 &\longrightarrow\mathbb E\{B(X)-Q(X)\}\ge0.
 \label{eq:multigate-rb-gain}
\end{align}
If every $\nu_k>0$, the last coefficient is strictly positive, even for singular
$\Sigma$.

If additionally $\Sigma$ is positive definite, define
\begin{equation*}
 I_k(x)=\mathbf1\{q_k(x)>\max(0,\max_{j\ne k}q_j(x))\}.
\end{equation*}
The empty inner maximum for $m=1$ is zero. Then
\begin{equation}
 n\mathbb E_C(p_+-p_0)\longrightarrow
 \sum_{k=1}^m\frac{\nu_k}{\alpha_k^2}\mathbb E I_k(X).
 \label{eq:multigate-raw-gain}
\end{equation}
This coefficient is also strictly positive if every $\nu_k>0$.
Positive definiteness can be weakened to $\Sigma_{kk}>0$ for every $k$ and
$\Pr\{Q(X)>0,\ \arg\max_kq_k(X)\text{ has multiple elements}\}=0$.

Suppose further that the fixed rule has $|g_u|\le B_g$ and population mean gain
$\mu_M\to\mu>0$. Write
$\Delta D_c=\mathbb E_C[(p_+-p_c)\overline g_T]$.
Replacing the left sides of \eqref{eq:multigate-rb-gain} and
\eqref{eq:multigate-raw-gain} by $\sqrt n\Delta D_{\rm RB}$ and
$n\Delta D_0$, respectively, multiplies each right side by $\mu$.
Gains need not be independent of any loss. All statements concern unquantized
probabilities.
\end{theorem}

For $m=1$, $\mathbb E Q=H_{s_C}(h)/\alpha$ and
$\mathbb E B=H_{s_K}(h)/\alpha$, recovering the single-constraint result.
The multiconstraint result includes bottleneck switches. The raw coefficient
weights each constraint by its probability of being the unique positive
bottleneck. The region where all constraints saturate contributes zero.
The RB result requires neither a unique positive bottleneck nor independent risks.
A weaker sufficient condition for strict RB improvement is
\begin{equation*}
 \Pr\{\exists k\in\arg\max_jq_j(X),\ \nu_k>0\}>0.
\end{equation*}
Fractionality confined to constraints that never affect the final probability
does not alone ensure strict improvement. Additional constraints whose population
means remain a fixed positive distance below their budgets have exponentially
small deficits and do not change these leading terms.

\begin{proof}
Write $S_{n,k}=\sum_{u\in C}l_{u,k}$,
$V_{n,k}=\sum_{u\in C}l_{u,k}(1-l_{u,k})$,
$X_{n,k}=(S_{n,k}-nr_{M,k})/\sqrt n$, and
$h_{n,k}=\sqrt n(\alpha_k-r_{M,k})$.
The finite population central limit theorem and Cram\'er--Wold argument give
\begin{equation*}
 X_n\Rightarrow X,\qquad V_{n,k}/n\to\nu_k.
\end{equation*}
For any linear projection with positive limiting variance, boundedness ensures
the maximum-deviation condition of the finite population central limit theorem.
A projection with zero limiting variance tends to zero in $L^2$.
Thus a singular $\Sigma$ does not prevent the vector limit.

\paragraph{First-order deficits of raw and the new operator.}
The single-constraint argument gives, for every $k$,
\begin{equation*}
 \sqrt n(1-w_{0,k})
 -\frac{(X_{n,k}-h_{n,k})_+}{\alpha_k}\longrightarrow0
 \quad\text{in }L^1.
\end{equation*}
The scaled deficits have uniformly bounded second moments.
The deterministic Jensen-increment bound also gives
\begin{equation*}
 \mathbb E(w_{+,k}-w_{0,k})^2=O(n^{-2}),
\end{equation*}
so the same first-order deficit applies to $w_{+,k}$.
The maximum is Lipschitz in the maximum norm and $m$ is fixed. Taking maxima
and expectations proves \eqref{eq:multigate-deficit-new}.

\paragraph{RB must retain the position of conditional averaging.}
Given $C$, let $K_{B,k}$ be the auxiliary Bernoulli count for constraint $k$ and
$Y_{n,k}=(K_{B,k}-S_{n,k})/\sqrt n$.
The coupling of auxiliary variables across constraints does not change any
individual conditional average. Uniformly over calibration vectors,
\begin{equation*}
 \mathbb E_B[e^{itY_{n,k}}\mid C]
 =\exp\{-t^2V_{n,k}/(2n)+O(n^{-1/2})\}.
\end{equation*}
The conditional law therefore converges in probability in first Wasserstein
distance to $N(0,\nu_k)$. To see this, every subsequence has a further subsequence
on which $V_{n,k}/n\to\nu_k$ almost surely. Along these conditional loss arrays,
the central limit theorem for independent bounded terms and the uniform second
moment bound imply weak convergence and convergence of first moments.
If $\nu_k=0$, use the conditional second moment directly.
Wasserstein distance is uniformly bounded by the sum of the two first absolute
moments, so convergence also holds in expectation.

The positive-part function is one-Lipschitz in its random argument, uniformly
over the conditional shift $X_{n,k}-h_{n,k}$.
Moreover, $\mathbb E(K_{B,k}-n\alpha_k)^2=O(n)$.
Replacing the exact deficit denominator $1+K_{B,k}$ by $n\alpha_k$ gives
\begin{equation*}
 \mathbb E\left|
 \sqrt n(1-w_{{\rm RB},k})
 -\frac1{\alpha_k}\mathbb E_B
       (X_{n,k}-h_{n,k}+Y_{n,k})_+
 \right|=O(n^{-1/2}).
\end{equation*}
Here a positive deficit implies $1+K_{B,k}\ge\alpha_k(n+1)$, and the count's
second moment controls the denominator replacement. Consequently,
\begin{equation*}
 \sqrt n(1-w_{{\rm RB},k})-b_k(X_n)\longrightarrow0
 \quad\text{in }L^1.
\end{equation*}
Conditional Jensen also gives uniformly bounded second moments for these scaled
deficits. Taking maxima and expectations proves \eqref{eq:multigate-deficit-rb}.
Subtracting the two deficits proves \eqref{eq:multigate-rb-gain}.
The calculation always uses $\max_k\mathbb E_B[\cdot\mid C]$, never
$\mathbb E_B[\max_k\cdot\mid C]$.

\paragraph{Strict improvement.}
Convexity gives $b_k(x)\ge q_k(x)$. If $\nu_k>0$, a Gaussian distribution has
positive probability on both sides of every finite threshold, so
$b_k(x)>q_k(x)$ for every finite $x$.
Whenever an index attaining $Q(x)$ has $\nu_k>0$, it follows that $B(x)>Q(x)$.
This proves the stated sufficient conditions for strictness.

\paragraph{The second-order raw increment and bottlenecks.}
Under the additional assumptions, each $X_k-h_k$ has no atom at zero.
The uniform single-constraint expansion jointly gives
\begin{equation*}
 n(w_{+,k}-w_{0,k})
 \Rightarrow\frac{\nu_k}{\alpha_k^2}\mathbf1\{X_k>h_k\}.
\end{equation*}
When $Q(X)>0$ and its maximizer is unique, away from ties its first-order deficit
differs from all others at order $n^{-1/2}$. The raw corrections are only
$n^{-1}$, so raw and the new operator eventually have the same bottleneck.
On $Q(X)=0$, all coordinates satisfy $X_k<h_k$ almost surely, and both raw and
the new operator eventually saturate, giving zero increment.
First restrict the positive maximum, tie margin, and distances of coordinates
from their thresholds away from zero by $\varepsilon$, then let
$\varepsilon\downarrow0$. This yields
\begin{equation*}
 n(p_+-p_0)\Rightarrow
 \sum_k\frac{\nu_k}{\alpha_k^2}I_k(X).
\end{equation*}
The bound $0\le p_+-p_0\le\max_k(w_{+,k}-w_{0,k})$ and uniform second moments
give uniform integrability, so expectations also converge.
Positive definite $\Sigma$ gives a full-dimensional Gaussian density.
Ties between positive deficits lie on nontrivial affine hyperplanes and have
zero probability. Each constraint's unique-positive-maximum region contains
a nonempty open set and thus has positive probability.
This proves \eqref{eq:multigate-raw-gain} and its strictness.

\paragraph{Signed utility on complementary users.}
The individual second-moment bounds and fixed $m$ imply
\begin{equation*}
 \mathbb E(p_+-p_{\rm RB})^2=O(n^{-1}),\qquad
 \mathbb E(p_+-p_0)^2=O(n^{-2}).
\end{equation*}
The complementary mean gain has standard deviation $O(n^{-1/2})$.
Its covariance with the increment is therefore $O(n^{-1})$ for RB and
$O(n^{-3/2})$ for raw, below the respective leading orders.
Substitution into the exact identity
\begin{equation*}
 \Delta D_c=\mu_M\mathbb E_C(p_+-p_c)
 +\operatorname{Cov}_C(p_+-p_c,\overline g_T)
\end{equation*}
proves the utility statements. Adding constraints with strict budget margins
uses the existing exponential tail bounds.
\end{proof}

\paragraph{Scope.}
The result allows both implemented risk constraints to approach their budgets
and change which constraint limits activation. It compares the specified
calibrators, not arbitrary rules that select users. The unique positive
bottleneck condition is used only for raw's second-order coefficient.
Finite samples still require exact increments or the established pointwise
bounds, rather than treating asymptotic coefficients as exact effects.
Quantization and the boundary $\nu_k=0$ retain the earlier qualifications.

\subsubsection{Finite-sample accuracy of the leading coefficients}\label{app:asymptotic-check}
\begin{figure}[htbp]
\centering
\includegraphics[width=\linewidth]{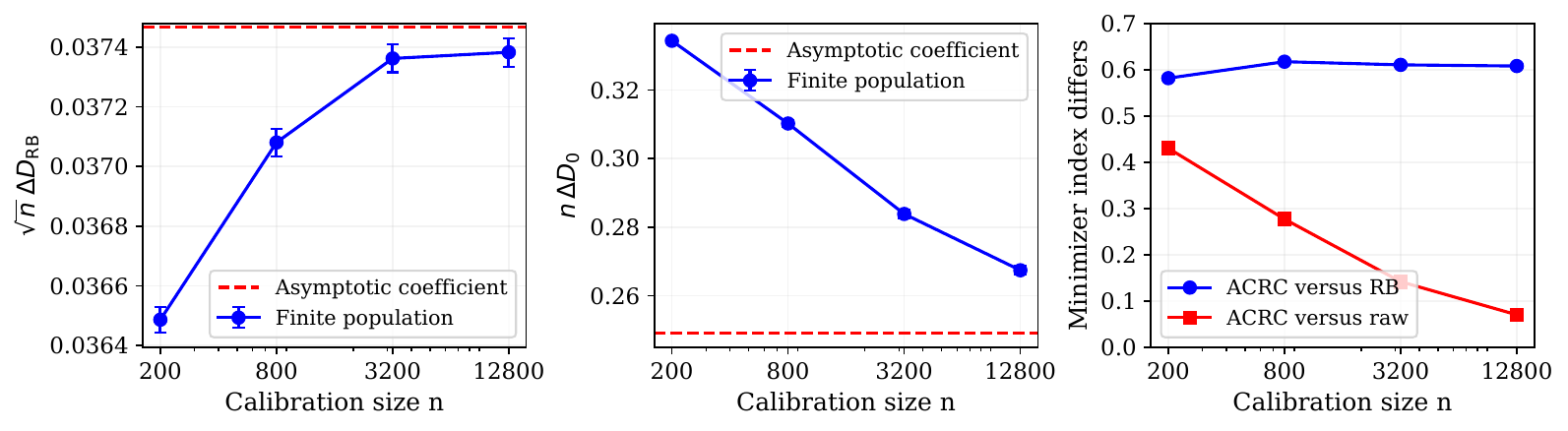}
\caption{Finite-sample increments and theoretical limits under two boundary constraints. Bars are approximate 95\% Monte Carlo intervals, often smaller than the markers. They quantify simulation error, not uncertainty across training seeds. The right panel compares minimizing constraint indices, breaking ties by the first index.}\label{fig:asymptotic-check}
\end{figure}

Figure~\ref{fig:asymptotic-check} consolidates the existing controlled finite-population diagnostic. Four user types have loss vectors $(0,0),(0,0.1),(0.2,0),(0.2,0.1)$ with proportions $(3,1,1,3)/8$, budgets $(0.1,0.05)$, and gains $(0.1,0.02,0.03,-0.04)$. Thus both population risks equal their budgets, the loss correlation is $0.5$, and mean gain is $0.02875$. We sample half the population without replacement. Each size uses $100000$ calibration draws. Conditional Bernoulli averages are computed analytically, rather than approximated by sampled activation coins.

The leading utility coefficients are $0.037468$ for $\sqrt n\Delta D_{\rm RB}$ and $0.249167$ for $n\Delta D_0$. At $n=200$, the measured values are $0.036486$ and $0.334335$, respectively. Their relative errors are $2.62\%$ and $34.18\%$. At $n=12800$, these errors fall to $0.23\%$ and $7.32\%$. The RB leading term is already informative at the smallest size in this example, whereas the raw approximation converges more slowly. The minimizing constraint indices for ACRC and RB differ in about $58$--$62\%$ of draws under a fixed first-index tie rule, including joint saturation. This is an index comparison, not a frequency of switches between unique active constraints. The measured utility uses complementary-user gains, retaining their sampling dependence on calibration.

At $n=200$, the scaled RB increment decomposes into $0.036869$ from mean gain times the activation increment and $-0.000383$ from their covariance. The corresponding raw terms are $0.329485$ and $+0.004850$. Thus heterogeneous gains affect both finite-sample results, but the larger raw approximation error mainly comes from the finite-sample activation term rather than from the covariance correction. 

This diagnostic supports the rates and coefficients under the stated conditions. It does not estimate an $n=200$ recommendation gain from an asymptotic coefficient. It uses one correlation and one heterogeneous gain pattern, so it does not establish uniform approximation accuracy over all populations.

\subsection{Risk validity after selection-set tuning}
\label{app:selection-tuning-validity}

The following proposition is a protocol consequence of existing calibration guarantees, rather than a new calibration contribution.
Let $\mathcal H$ contain the frozen model, fitting information, the selection set and its feedback, model and risk-score fitting seeds, and unlabeled information used according to the fixed protocol. $\mathcal H$ excludes the random seed and realization of the final without-replacement $C/T$ assignment. Conditional on $\mathcal H$, fix the selected proposal $j$, its two risk score functions, loss bounds $H^k_j$, and budget multiplier $\widehat m\in(0,1]$. Let $R$ be the remaining finite user pool. Draw $C$ uniformly without replacement and set $T=R\setminus C$.

Let $x^k_{ju}$ denote downside loss for risk $k$ when user $u$ receives proposal $j$, integrating its internal action randomization analytically. If the proposal takes actions $a_R$ and $a_Q$ with probabilities $1-\eta$ and $\eta$, define
\begin{align}
 x^k_{ju}
 &= (1-\eta)
 \big[M_k(\pi_0(u),u)-M_k(a_R,u)\big]_+ \notag\\
 &\quad+\eta
 \big[M_k(\pi_0(u),u)-M_k(a_Q,u)\big]_+ .
 \label{eq:module-action-loss}
\end{align}
Here $M_k$ is utility or Recall. Positive parts precede action averaging, so gains under one action cannot cancel losses under another.

\begin{proposition}\label{prop:selection-tuning-validity}
Suppose $0\le x^k_{ju}\le H^k_j$ and, conditional on the information above, the loss and score sequence obtained by randomly ordering the remaining pool is exchangeable. For each active constraint, set
\begin{equation*}
 L^k_{ju}=\frac{x^k_{ju}}{H^k_j},
 \qquad
 \alpha_k=\frac{\widehat m\epsilon_k}{H^k_j}.
\end{equation*}
If $H^k_j\le\widehat m\epsilon_k$, let the corresponding gate be identically one. Otherwise apply native SCoRE-MDR to $C$ at level $\alpha_k$ to obtain $\phi_k(C,u)$. Set $\phi(C,u)=\phi_U(C,u)\phi_R(C,u)$. Then each constraint satisfies
\begin{equation}
 \E_C\!\left[
 \frac{1}{|T|}\sum_{u\in T}\phi(C,u)x^k_{ju}
 \,\middle|\,\mathcal H
 \right]
 \le \widehat m\epsilon_k
 \le \epsilon_k .
 \label{eq:selection-tuning-risk}
\end{equation}
Raw, RB, or $w_+$ scalar gates selected using $S$ alone satisfy the same physical-budget guarantee when their two risk gates are combined by taking the minimum.
\end{proposition}

\begin{proof}
Conditional on $\mathcal H$, the upstream proposal, risk scores, multiplier, and bounds are fixed. Uniformly permute $R$, use the first $|C|$ entries for calibration, and draw one evaluation user uniformly from the remaining positions. This exchangeable sequence permits the native SCoRE-MDR marginal risk guarantee\cite{score}. For an active constraint, $\E[\phi_k L^k\mid\mathcal H]\le\alpha_k$, and multiplication by $H^k_j$ gives the physical budget. Inactive constraints meet their budgets directly through the userwise bound.
Nonnegativity of loss and $\phi\le\phi_k$ preserve each guarantee after intersection. Averaging over remaining users is equivalent to the uniform evaluation user, yielding Equation~\eqref{eq:selection-tuning-risk}. Equation~\eqref{eq:module-action-loss} is the conditional expected downside of the actual action, so the statement also covers internal action randomization.
For scalar gates, apply the existing single-gate risk guarantees and use that the common activation probability never exceeds either gate. Parameter selection depends only on $\mathcal H$, so it does not search across configurations using the final calibration set. Averaging over $\mathcal H$ yields the unconditional guarantee.
\end{proof}

The guarantee averages over calibration assignments, evaluation users, and deployment randomization. Realized risk for one split, user-level risks, and utility gains must still be reported separately and do not follow directly from Equation~\eqref{eq:selection-tuning-risk}.

\subsection{Finite-pool expected-risk evaluation}\label{app:expected-risk-audit}
Fix $F,S$ and the selected rule. For one constraint, let $\mathcal R$ be the remaining pool of $M>n$ users, with action-averaged downside $x_u\in[0,H]$, budget $\epsilon$, and $\alpha=\epsilon/H\in(0,1)$. Draw $C$ uniformly without replacement, set $N=n+1$, and define

\begin{equation*}
 \overline P_N=\binom{M}{N}^{-1}
 \sum_{\substack{A\subseteq\mathcal R\\|A|=N}}
 \prod_{u\in A}(1-x_u/H).
\end{equation*}
\begin{corollary}\label{cor:finite-risk-audit}
The implemented two-constraint, downward-quantized activation satisfies

\begin{equation}
 \E_C\left[p_C\overline x_{\mathcal R\setminus C}\right]
 \le \min\left\{\overline x_{\mathcal R},
                  \epsilon(1-\overline P_N)\right\}.
 \label{eq:finite-risk-audit}
\end{equation}
\end{corollary}
\begin{proof}
Uniformly drawing one user from the complement of $C$ is equivalent to drawing $N$ users and leaving one out uniformly. Average~\eqref{eq:identity} over these subsets and use that actual $p_C$ does not exceed either individual $w_+$ to obtain the second bound. The first follows from $p_C\le1$ and the unbiased complement mean. For inactive $H\le\epsilon$, use $\overline x_{\mathcal R}\le\epsilon$ directly.
\end{proof}
Subset enumeration is unnecessary. Let $D_{m,k}$ be the mean product of $k$ factors among the first $m$ users, with $z_m=1-x_m/H$. Then

\begin{equation*}
 D_{m,k}=\frac{m-k}{m}D_{m-1,k}
         +\frac{k}{m}z_mD_{m-1,k-1},\qquad D_{m,0}=1.
\end{equation*}
Unreachable states equal zero. The recursion gives $\overline P_N=D_{M,N}$ in $O(MN)$ time and $O(N)$ memory. Nonnegative recursion steps round downward and the final risk upper bound rounds upward. This retrospective evaluation uses complete remaining-pool feedback. It is not a deployment certificate available from calibration data alone.

Table~\ref{tab:expected-risk-audit} retains empirical risks alongside the conditional expected-risk bounds. Shared-cohort LightGCN has empirical utility downside $0.020595$, above budget by $0.000595$. We retain this result without retuning. Its conditional expected-risk bound is $0.019944$. A finite average over assignments differs from the randomization expectation and need not lie below its upper bound.

\begin{table}[!htb]\centering\small\setlength{\tabcolsep}{2.8pt}
\caption{Empirical risk and conditional expected-risk upper bounds, all lower is better. Budgets are $0.02$ and $0.01$. Bounds use the complete remaining pool for each fixed $S$, then average five seeds and forty assignments. The averaged bound is not a confidence bound for the reported finite empirical average.}\label{tab:expected-risk-audit}
\begin{tabular}{@{}llrrrr@{}}\toprule
Cohort & Model & Observed $R_U$ & Expected $R_U$ bound & Observed $R_R$ & Expected $R_R$ bound\\\midrule
\multirow{3}{*}{shared} & LightGCN & 0.02060 & 0.01994 & 0.007015 & 0.008328\\
 & XSimGCL & 0.01935 & 0.01948 & 0.006632 & 0.007360\\
 & SIGformer & 0.01762 & 0.01898 & 0.006378 & 0.007226\\
\midrule
\multirow{3}{*}{full} & LightGCN & 0.02024 & 0.01996 & 0.007067 & 0.008644\\
 & XSimGCL & 0.01964 & 0.01971 & 0.006855 & 0.007853\\
 & SIGformer & 0.01964 & 0.01955 & 0.006795 & 0.007463\\
\bottomrule\end{tabular}
\end{table}

\section{Experimental Protocols}\label{app:protocol-complete}
\subsection{Reproducibility Details}\label{app:implementation}

\subsubsection{Candidate construction and selection}
We retain the backbone and fit a local head on its observed Top20 and a contextual head on the available observed pool. The heads use an internal $160/40$ split of $F=200$ users for regularization selection and are then refit on all $F$. Features include the frozen score, historical user and item counts, smoothed positive fractions, and regularized item residuals. The objectives and bounds are specified below. Rule templates and parameter grids are common across datasets. Fitted heads, eligible item pools, and selected rules can differ, so sharing templates does not require identical realized lists.

Each head supplies ordering projections with strengths $0.5$ and $1$, both within the reference set and over the observed Top80. Additional rules replace the last one or four items when the fitted probability improves by more than $0.02$, or select a fitted-utility prefix. The reference list is included first. We also use fixed mixtures of a reordered reference set $R_h$ and its best prefix $Q_h$, with probabilities $1-\eta$ and $\eta$, for $\eta\in\{0.01,0.025,0.05,0.1,0.25,0.5\}$. Candidates are deduplicated only by their frozen action distributions. The loss in~\eqref{eq:actual} is computed before averaging these actions.

For selection, each of the $b=24$ shared subsets $P_t\subset S$ supplies candidate losses for the two raw gates. The complementary users $S\setminus P_t$ supply signed gains. Equation~\eqref{eq:selection} averages their product and chooses one rule, with ties resolved by the frozen order. The rule can return different rankings and lengths for different users, but its parameters and action distributions are fixed before $C$. Final calibration does not search over candidates. The default same-rule comparisons reuse this choice and change only the calibration operator.

\paragraph{Scope beyond these templates.}
Theorem~\ref{thm:dual-risk} applies to another proposal generator if the reference, selected rule, and valid loss bounds are frozen before $C$ under the stated sampling protocol. The proof does not require the heads or templates above. A learned reranker can therefore supply a proposal, but replacing its generator does not inherit the measured utility improvements. The cross-proposal and frozen-MIR analyses test particular alternatives. They do not establish performance for arbitrary learned generators or unobserved items. More activation is useful only when the proposed changes and their dependence on calibration support a positive utility increment.

\subsubsection{Observed feedback and user populations}
\paragraph{Data preparation.}
We use five development domains with the source versions, relevance labels, and interaction splits listed in Table~\ref{tab:data_protocol_verified}.
Backbone training and validation use disjoint interaction pairs.
The held-out observed feedback is subsequently divided by user into fitting, selection, calibration, and evaluation roles.
Evaluation candidates are restricted to users and items that occur in positive training interactions.
Unobserved pairs remain missing and are never converted to evaluation negatives.

\begin{table*}[htb]
\centering
\caption{Data protocol for the five development domains. The final column is the minimum number of held-out observed candidates required before fitting users are sampled.}
\label{tab:data_protocol_verified}
\small\setlength{\tabcolsep}{3pt}
\begin{tabular}{@{}p{0.12\textwidth}p{0.18\textwidth}p{0.13\textwidth}p{0.42\textwidth}r@{}}
\toprule
Domain & Source version & Positive label & Interaction split and duplicate handling & Minimum \\
\midrule
KuaiRec & KuaiRec 2.0 small matrix & Watch ratio $\geq 1$ & Keep the earliest observation per pair. A label-blind global permutation assigns 10\% to training, 10\% to validation, and 80\% to held-out feedback. & 20 \\
KuaiRand-random & KuaiRand-Pure & Official click label & Split the April 8--21 standard log globally in time into 80\% training and 20\% validation. Use the April 22--May 8 random log for held-out feedback. Keep the earliest pair in each log and remove all historical pairs from the random log. & 20 \\
Music & SIGformer-released Music benchmark & Rating $\geq 4$ & Retain the released training, validation, and test splits of 34,915, 4,986, and 9,974 pairs. No within-split duplicate or cross-split overlap is present. & 2 \\
Epinions & SIGformer-released Epinions benchmark & Rating $\geq 4$ & Retain the released training, validation, and test splits of 289,645, 41,376, and 82,753 pairs. No within-split duplicate or cross-split overlap is present. & 2 \\
MovieLens-25M & Complete official ML-25M release & Rating $\geq 4.0$ & Use all 25,000,095 ratings. Unique pairs are sorted within each user by timestamp, with item-ID tie breaking, and split chronologically into 80\% training, 10\% validation, and 10\% held-out feedback. & 40 \\
\bottomrule
\end{tabular}
\end{table*}

\paragraph{Observed candidate lists.}
For a user with $m_u$ eligible observed candidates, observedTop20 contains the first $\min(20,m_u)$ items in descending frozen backbone score, with item ID breaking score ties.
Top80 denotes the first $\min(80,m_u)$ items under the same ordering.
Short pools retain their actual size and are not filled with unobserved or repeated items.
For each nested pair, the ranking stage preserves the observedTop20 item set, and the truncation stage returns a nonempty prefix of that ranking.
The Top80 expansion rules are separate candidates and do not define these nested pairs.

\paragraph{User roles and repeated evaluation.}
For each domain, we fix 200 fitting users $F$ sampled without using held-out labels from the eligible population.
Their observed feedback fits the prediction heads.
After removing $F$, the shared comparison cohort additionally requires at least one positive historical validation interaction with an item in the training catalog.
This support condition does not use held-out labels.
Each of 40 random partitions draws 400 users for selection $S$ and 200 for calibration $C$ without replacement, with all remaining users assigned to evaluation $T$.
The four roles are disjoint within each partition, and the same role assignments are shared across methods, backbones, and five training seeds.
The shared-cohort sizes of $T$ are 611, 9,613, 545, 8,782, and 13,018 for KuaiRec, KuaiRand-random, Music, Epinions, and MovieLens-25M, respectively.
We first average over the 40 partitions within each training seed and then compare seed-level results.
These partitions reuse a fixed user pool and do not constitute independent datasets.
Table~\ref{tab:users} reports model catalog sizes and eligible evaluation populations. The common cohort excludes $F$ and requires historical PerK support. The full cohort includes all non-fitting eligible users. PerK is not extrapolated to users without its required historical positive feedback.

\begin{table}[h]\centering\small
\caption{Frozen populations. Eligible users include $F=200$. Common users and $T$ exclude $F$. $S=400$ and $C=200$ are removed from the common population to obtain $T$.}\label{tab:users}
\begin{tabular}{@{}lrrrrr@{}}\toprule
Data & Training users & Items & Eligible & Common & $|T|$\\\midrule

KuaiRec & 1411 & 3185 & 1411 & 1211 & 611\\
KuaiRand & 24231 & 7004 & 12626 & 10213 & 9613\\
Music & 3472 & 2498 & 2066 & 1145 & 545\\
Epinions & 17894 & 17660 & 13206 & 9382 & 8782\\
ML-25M & 162205 & 35015 & 14041 & 13618 & 13018\\

\bottomrule\end{tabular}\end{table}

\subsubsection{Probability heads and candidate ranges}
The local head is fitted on reference Top20 rows. It considers a score-only head, a contextual correction, and contextual item corrections with item penalties $0.1$ and $1$. The standardized score slope lies in $[0,10]$. Context coefficients and item residuals lie in $[-0.25,0.25]$. Feature regularization is $0.0001$. The $F$-internal validation compares joint order/prefix utility and favors a simpler head on ties.

The contextual head uses the frozen score, $\log(1+n_i)$, $(n_i^++1)/(n_i+2)$, $\log(1+n_u)$, and $(n_u^++1)/(n_u+2)$, with historical counts. Its user-normalized logistic objective is

\begin{equation*}
\mathcal L_F
=\frac1{|F|}\sum_{u\in F}\frac1{m_u}\sum_{i\in\mathcal I_u^F}
\big[\log(1+e^{h_{ui}})-Y_{ui}h_{ui}\big]
+\frac{0.0001}{2}\|\beta\|_2^2+\frac{\rho}{2I}\sum_i\delta_i^2,
\qquad h_{ui}=\beta^\top\widetilde x_{ui}+b+\delta_i.
\end{equation*}

The item penalty $\rho\in\{0.01,0.1,1\}$ is selected using only the internal $160/40$ split and the fixed Top20 utility objective. Standardization also uses only fitting rows. The standard-deviation floor is $10^{-8}$. L-BFGS-B uses function and gradient tolerances $10^{-12}$ and $10^{-7}$ and at most forty line-search steps. We accept a fit only when the optimizer reports success and its parameters are finite. We check the projected-gradient infinity norm for the bounded local head and the ordinary-gradient infinity norm for the global contextual head. Both must be at most $2\times10^{-5}$. The MovieLens iteration cap is ten thousand, with fitting restricted to $F$.

For $\lambda=1$, embed a list as a discount vector $c_a$ on the common observed item pool. Its gain is $(c_a-c_0)^\top(2Y-1)$, so $|G_{ua}|\le\|c_a-c_0\|_1$. The maximum across the frozen pool is a label-independent legal action bound. A mixture uses the corresponding weighted sum of bounds. Reordering the reference set preserves Recall. A mixture with its prefix therefore has a Recall downside range at most $\eta$. General replacement candidates use range one unless their action structure gives a smaller valid bound.

Ordering projection sorts $(1-\beta)c_0+\beta c_h$ and reassigns the standard discounts. Prefixes maximize predicted utility over lengths at least one. Two restricted-prefix variants retain at least $75\%$ of the reference length, rounded up. Tail replacement checks the last one or four positions and the fixed probability improvement threshold $0.02$. Candidate identity and range calculations do not use $S,C,T$ labels.

\subsubsection{External methods and training settings}
PerK uses the author userwise calibration dataset construction and Platt implementation. Its PDCG prefix objective follows the paper rather than the F1 objective of the public notebook. Unknown positives are not removed from its native negative sampling by consulting test feedback.

UMC retains the author contextual integral network with hidden widths $[50,50]$, fifty integration steps, positive contextual scaling, and Smooth Calibration Loss. It uses Adam with learning rate $0.001$, batch size $16384$, twenty fixed epochs, loss weight $0.01$, ten bins, and cache coefficient $0.95$. No test metric selects its cutoff or checkpoint.

Each backbone is trained with five random seeds. MovieLens retains $19888992$ historical feedback rows and $10077856$ positive rows. LightGCN uses batch size $65536$ and $154000$ optimizer updates per seed. XSimGCL uses batch size $16384$ and $616000$ updates per seed. Larger batches change update counts and XSimGCL's in-batch contrastive negatives, so these are explicit execution differences from author defaults.

SIGformer retains its complete original loss and one Adam update per epoch, using edge chunking, explicit sparse backward computation, and activation recomputation. Validation early stopping uses a patience of two hundred epochs. Training terminates between 240 and 620 epochs across the five seeds, with the best validation checkpoint used for evaluation. Training uses two RTX 5090 GPUs.

\subsubsection{Metrics and exact action averaging}
For an observed pool with $R_u$ positive items, let $D_u(a)=\sum_{r\le|a|}d_rY_{u,i_r}$ and $H_u(a)=\sum_{r\le|a|}Y_{u,i_r}$. Then

\begin{equation*}
\mathrm{NDCG@20}(a)=\frac{D_u(a)}{\sum_{r=1}^{\min(20,R_u)}d_r},\qquad
\mathrm{Recall@20}(a)=\frac{H_u(a)}{R_u},\qquad
\mathrm{F1}(a)=\frac{2H_u(a)}{|a|+R_u}.
\end{equation*}

NDCG and Recall are zero when $R_u=0$. All methods share the same observed-pool denominators. For any list metric $M$, deployed evaluation is

\begin{equation*}
M_u^{\rm deploy}=(1-p_C)M_u(\pi_0)+p_C\sum_a\nu_j(a\mid u)M_u(a).
\end{equation*}

Downside evaluation uses $p_Cx_{ju}$ and $p_Cx^R_{ju}$ from~\eqref{eq:actual}. The reference has zero relative downside by definition, not perfect absolute recommendation quality.

\subsubsection{Utility valuation and attribution}\label{app:utility-accounting}
Equation~\eqref{eq:utility} states an additive preference over observed relevance. Define the normalized relevant and irrelevant exposure weights
\begin{align}
 Q_u^+(a)&=\frac{1}{W_K}\sum_{r\le |a|}d_rY_{u,i_r}, &
 Q_u^-(a)&=\frac{1}{W_K}\sum_{r\le |a|}d_r(1-Y_{u,i_r}),\notag\\
 U_\lambda(a,Y_u)&=Q_u^+(a)-\lambda Q_u^-(a).\label{eq:utility-accounting}
\end{align}
Thus $\lambda$ is the relative cost of an irrelevant exposure when a relevant exposure earns one unit. At a fixed position, adding an item has positive expected contribution precisely when its relevance probability exceeds $\lambda/(1+\lambda)$. The default $\lambda=1$ treats the two outcomes symmetrically. It is a transparent benchmark preference, not a fitted economic parameter. The metric ranges from $-\lambda$ to $1$, so its absolute differences should not be interpreted as percentage changes in clicks or revenue. The position discount is an evaluation weight, not an estimated causal examination probability.

For a split with frozen proposal $\nu$ and scalar probabilities $p_+$ and $p_0$, define $h_u^\pm=\sum_a\nu(a\mid u)Q_u^\pm(a)-Q_u^\pm(\pi_0(u))$. Exact action averaging gives
\begin{equation}
 \Delta U_\lambda=(p_+-p_0)\overline h_T^+
                  -\lambda(p_+-p_0)\overline h_T^- .\label{eq:calibration-utility-accounting}
\end{equation}
The first term is the change in discounted relevant reward. The second is the saving in irrelevant exposure cost. They are accounting terms, not independent effects. If total discounted exposure is unchanged, $\Delta Q^-=-\Delta Q^+$ and $\Delta U_1=2\Delta Q^+$. Both terms retain signed changes, so positive net utility can coexist with lower relevant reward. A new valuation requires its own losses and valid bounds before final calibration. A certificate at one cost does not automatically cover a different cost chosen afterwards.

Table~\ref{tab:utility-accounting} decomposes the existing decisions without refitting, reselection, or recalibration. Per-action discounted mass and the saved utility recover $Q^+$ and $Q^-$, independently checked against the stored NDCG numerators. We average paired split differences within model before averaging backbones and training seeds. In particular, we do not substitute average list length into the nonlinear discount sum or difference utilities from distinct cost-selected policies. These are post-hoc descriptions on previously inspected data, not additional independent confirmations.

\begin{table}[!htb]\centering\small\setlength{\tabcolsep}{4pt}
\caption{Net-utility decomposition of ACRC minus raw with the proposal fixed, on the common-support population and default budgets of Table~\ref{tab:revision-main}. The first three columns give paired five-seed means and approximate 95\% interval half-widths, all higher is better. Length change is descriptive. Backbones are equally weighted within seed, followed by equal domain weighting for the last row. Jester here is supplementary, not the frozen confirmation. Columns are accounting components, not a ranking of methods.}\label{tab:utility-accounting}
\begin{tabular}{@{}lrrrr@{}}\toprule
Dataset & $\Delta Q^+\uparrow$ & $-\Delta Q^-\uparrow$ & $\Delta U_1\uparrow$ & $\Delta |a|$\\\midrule
KuaiRec & $+0.017711\pm0.000294$ & $+0.017931\pm0.000327$ & $+0.035643\pm0.000621$ & $-0.006606$\\
KuaiRand & $-0.000096\pm0.000012$ & $+0.000544\pm0.000049$ & $+0.000448\pm0.000051$ & $-0.016436$\\
Music & $+0.000027\pm0.000015$ & $+0.000081\pm0.000053$ & $+0.000108\pm0.000068$ & $-0.000993$\\
Epinions & $+0.000806\pm0.000245$ & $+0.000924\pm0.000267$ & $+0.001730\pm0.000511$ & $-0.002525$\\
ML-25M & $+0.005934\pm0.000649$ & $+0.005934\pm0.000649$ & $+0.011867\pm0.001297$ & $0.000000$\\
Jester & $+0.005765\pm0.000623$ & $+0.005796\pm0.000624$ & $+0.011561\pm0.001247$ & $-0.000794$\\
\midrule
Macro mean & $+0.005024\pm0.000199$ & $+0.005202\pm0.000202$ & $+0.010226\pm0.000401$ & $-0.004559$\\
\bottomrule\end{tabular}
\end{table}

MovieLens has no length change, and its reward and cost terms each contribute $0.005934$. KuaiRand instead has a relevant-reward change of $-0.000096$, offset by an irrelevant-cost saving of $0.000544$. The default objective values this tradeoff positively. It does not show uniformly higher relevance or justify a claim about pure click utility.

\subsubsection{Data separation and evaluation scope}
The $F/S/C/T$ partitions are disjoint, and methods share the same evaluation users within each assignment. Selection and calibration do not use $T$ feedback. Results on domains already inspected during development are exploratory. Confirmatory evidence is reported separately for feedback obtained after the method was frozen.

\paragraph{Confirmation materials.}
The supplementary archive \texttt{confirmation\_evidence.zip} includes the pre-data protocol, endpoint, freeze record, execution amendment, source provenance and per-model results. The time record is author-maintained, not an external preregistration. A standalone script reproduces the primary interval from the packaged split results. Raw ratings and trained checkpoints are not embedded. Appendix~\ref{app:jester} distinguishes the primary cohort from later supplementary comparisons.

\subsection{Frozen user filter for the Jester confirmation}\label{app:decision-diagnosis}
S filter retains the original raw selector and selected rule, learning only which users receive that modification. Features use our own $F$ probability heads, frozen-list predicted utilities, predictive uncertainty, length, overlap and observed-item count. No external prediction component is used. Two user-level folds within $S$ generate predictions of $g,x,x^R$ from standardized linear Ridge regression with coefficient one. Predictions select users, and risk validity does not require prediction accuracy.

Options are no filtering and $\widehat g>\lambda_U\widehat x+\lambda_R\widehat{x^R}$, where each coefficient belongs to $\{0,1,4\}$, giving ten fixed options. Twenty-four pseudo-calibration new-operator gain scores on $S$ choose the option, with no filtering preferred on ties. The small gain/risk regression is then refitted on all $S$ and uses features to predict $C/T$ masks. The base recommender is not retrained.

Let the frozen mask be $m_S(u)\in\{0,1\}$. The proposal becomes
\begin{equation}
 \widetilde\nu(a\mid u)=m_S(u)\nu_j(a\mid u)+[1-m_S(u)]\delta_{\pi_0(u)}(a).
\end{equation}
\begin{proposition}\label{prop:filtered-finite-risk}
Any mask learning and option selection using $S$ only can precede the original exact dual calibration. Use all of $C$, actual action losses $\widetilde x_u=m_S(u)x_{ju}$ and $\widetilde x^R_u=m_S(u)x^R_{ju}$, and the original ranges and physical budgets. Expected fixed-population complementary risks still satisfy Theorem~\ref{thm:dual-risk}.
\end{proposition}
\begin{proof}
Condition on $S$, fitting randomness and all frozen unlabeled features. Then $\widetilde{\nu}$ is a fixed input-dependent rule, with $0\le\widetilde x_u\le H_j$ and $0\le\widetilde x^R_u\le H^R_j$. A zero mask executes the reference, so paired loss is exactly zero rather than an imputation of missing labels. Uniform calibration sampling makes the transformed losses exchangeable in the remaining fixed pool. Apply the original theorem and average over the conditioning information. Downward quantization and taking the smaller gate retain the risk bounds.
\end{proof}
One common calibration uses all of $C200$ without splitting users into calibration groups. Prediction error can hurt utility but cannot invalidate the stated fixed-population average-risk guarantee. This corollary connects the implementation to theory and is not a separate novelty claim. raw, RB and the new operator are compared on the same frozen mask to separate filtering gains from calibration gains.
For an identical rule and whole $C$, filtering decreases every loss coordinate. Each factor of the integral increases, so both gates, their minimum and downward quantization cannot decrease. The filtered common probability $p_{\mathrm F}$ is therefore no smaller than default $p_{\mathrm D}$. Nevertheless, realized utility obeys
\begin{equation}
 U_{\mathrm F}(T)-U_{\mathrm D}(T)
 =(p_{\mathrm F}-p_{\mathrm D})\overline{m_Sg}_{T}
   -p_{\mathrm D}\overline{(1-m_S)g}_{T}.\label{eq:filter-utility-decomposition}
\end{equation}
The first term measures activation changes on retained users and the second is the signed opportunity cost of rejected users. A higher valid probability cannot compensate for every filtering error. In particular, positive net rejected gain makes the second term negative. 

\section{Complete External Comparisons}\label{app:external-complete}
\subsection{Recommendation quality and uncertainty}\label{app:analysis}
Table~\ref{tab:utility} reports all eighteen utility settings with five-seed uncertainty. We retain quality changes and actual downside risks here.

\begin{table}[!htb]\centering\small
\caption{Quality changes relative to Top20, higher is better, and mean deployed downside magnitudes, lower is better. Quality differences show paired seed means and standard deviations. Length is descriptive. The target mean downsides are $0.02$ for utility and $0.01$ for Recall. These are not violation probabilities or per-split guarantees.}\label{tab:quality}
\begin{tabular}{@{}llrrrrr@{}}\toprule
Dataset & Model & $\Delta$NDCG@20 & $\Delta$Recall@20 & $R_U\downarrow$ & $R_R\downarrow$ & Length\\\midrule

\multirow{3}{*}{KuaiRec} & LG & $+0.15723\pm0.00421$ & $+0.00370\pm0.00010$ & $0.01667$ & $0.00021$ & $19.961$\\
 & X & $+0.13322\pm0.00496$ & $+0.00331\pm0.00010$ & $0.01655$ & $0.00019$ & $19.926$\\
 & SIG & $+0.03649\pm0.00504$ & $+0.00117\pm0.00023$ & $0.01501$ & $0.00017$ & $19.999$\\
\midrule
\multirow{3}{*}{KuaiRand} & LG & $-0.00521\pm0.00026$ & $-0.00967\pm0.00024$ & $0.00758$ & $0.00986$ & $19.519$\\
 & X & $-0.00534\pm0.00074$ & $-0.00953\pm0.00050$ & $0.00924$ & $0.00994$ & $19.534$\\
 & SIG & $-0.00442\pm0.00019$ & $-0.00964\pm0.00008$ & $0.01225$ & $0.00964$ & $19.594$\\
\midrule
\multirow{3}{*}{Music} & LG & $+0.00688\pm0.00105$ & $-0.00253\pm0.00084$ & $0.00882$ & $0.00303$ & $4.827$\\
 & X & $+0.00706\pm0.00178$ & $-0.00290\pm0.00140$ & $0.00555$ & $0.00323$ & $4.814$\\
 & SIG & $+0.01892\pm0.00049$ & $+0.00106\pm0.00013$ & $0.00813$ & $0.00047$ & $4.856$\\
\midrule
\multirow{3}{*}{Epinions} & LG & $+0.03615\pm0.00264$ & $-0.00118\pm0.00150$ & $0.01230$ & $0.00294$ & $6.302$\\
 & X & $+0.03556\pm0.00100$ & $-0.00045\pm0.00068$ & $0.00953$ & $0.00284$ & $6.300$\\
 & SIG & $+0.05885\pm0.00346$ & $+0.00221\pm0.00197$ & $0.01080$ & $0.00138$ & $6.327$\\
\midrule
\multirow{3}{*}{ML-25M} & LG & $+0.02340\pm0.00101$ & $+0.01448\pm0.00095$ & $0.01690$ & $0.00519$ & $20.000$\\
 & X & $+0.03066\pm0.00601$ & $+0.02640\pm0.00559$ & $0.01385$ & $0.00962$ & $19.991$\\
 & SIG & $+0.06893\pm0.01069$ & $+0.05382\pm0.00506$ & $0.01744$ & $0.00794$ & $20.000$\\

\bottomrule\end{tabular}\end{table}

\subsection{Statistical unit and paired effects}\label{app:uncertainty}
All inferential units are the five training seeds. Forty role assignments are averaged within each seed before contrasts and macro averages are formed. For seed-level paired differences $d_b$, the pointwise approximate 95\% interval is $\bar d\pm t_{4,0.975}s_d/\sqrt5$. These intervals describe training-seed variation conditional on the observed feedback, fixed protocol and working normal approximation. Repeated assignments are not independent user samples. Mechanism-family multiplicity correction and the frozen Jester confirmation are identified separately below.
\subsection{Original risk-control methods at the stated budgets}\label{app:review-completion}
Official SCoRE-MDR uses fixed risk scores to decide whether to execute original PerK or UMC lists for each user, with Top20 as fallback. Each risk calls the unchanged native algorithm and the two selection sets are intersected. Nonnegative losses ensure that intersection preserves each original risk guarantee. Scores use the author default random forest fitted only on $S400$. UMC probability calibration uses $F200$, while PerK retains its native historical validation feedback. No proposed head, candidate bank, selector, action-range computation or final calibrator enters an external method.

PerK returns an original-order prefix, so its utility downside is bounded by omitted discounted mass divided by $W_{20}$, hence by one. UMC may replace items and reorder, so its generic bound is two. The primary comparison uses the natural bound favorable to PerK. We also report sensitivity to the generic bound of two. Physical budgets are always $0.02$ and $0.01$. SCoRE MDR is equivalent to CRC on the same threshold family, so we do not count them as independent algorithms. The single-risk version is original Algorithm 1. The two-risk application explicitly intersects two native calls.

Risk scores use the default 100-tree random forest. PerK features contain twenty native probabilities, twenty padding indicators and length divided by twenty. UMC features contain its forty probabilities on reference and native Top20 lists, plus the logarithm of one plus available item count. Normalized utility risk levels are $0.02$ for PerK and $0.01$ for UMC. Recall levels are $0.01$, with default $\gamma=\alpha$.

\begin{table}[!htb]\centering\small\setlength{\tabcolsep}{2.6pt}
\caption{Original risk-control comparison at common budgets. Five seed means, each averaged over forty shared assignments. Colors rank each metric among the three risk-control methods.}\label{tab:external-risk}
\begin{tabular}{@{}llrrrrrrrrr@{}}\toprule
Dataset & Model & \multicolumn{3}{c}{SCoRE/PerK} & \multicolumn{3}{c}{SCoRE/UMC} & \multicolumn{3}{c}{\acrcours{}}\\
\cmidrule(lr){3-5}\cmidrule(lr){6-8}\cmidrule(lr){9-11}
 & & $U\uparrow$ & $R_U\downarrow$ & $R_R\downarrow$ & $U\uparrow$ & $R_U\downarrow$ & $R_R\downarrow$ & $U\uparrow$ & $R_U\downarrow$ & $R_R\downarrow$\\\midrule
\multirow{3}{*}{KuaiRec} & LG & 0.2242 & \best{0.000827} & \best{0.000093} & \second{0.2278} & \second{0.010583} & \second{0.000122} & \best{0.5391} & 0.016667 & 0.000206 \\
 & X & 0.2464 & \second{0.015507} & 0.001013 & \second{0.2499} & \best{0.011773} & \best{0.000136} & \best{0.5221} & 0.016547 & \second{0.000195} \\
 & SIG & \second{0.6812} & \best{0.000055} & \best{0.000012} & 0.6791 & \second{0.010530} & \second{0.000123} & \best{0.7543} & 0.015014 & 0.000172 \\
\midrule
\multirow{3}{*}{KuaiRand} & LG & \second{-0.4883} & \best{0.000552} & \second{0.005961} & -0.4977 & \second{0.003047} & \best{0.002622} & \best{-0.4846} & 0.007583 & 0.009855 \\
 & X & \second{-0.5008} & \best{0.000487} & \second{0.006042} & -0.5113 & \second{0.003403} & \best{0.003096} & \best{-0.4977} & 0.009244 & 0.009944 \\
 & SIG & \second{-0.4608} & \best{0.000564} & \second{0.006213} & -0.4695 & \second{0.003166} & \best{0.002419} & \best{-0.4555} & 0.012247 & 0.009640 \\
\midrule
\multirow{3}{*}{Music} & LG & \second{0.2589} & \best{0.001643} & 0.005399 & 0.2576 & 0.010203 & \best{0.000128} & \best{0.2663} & \second{0.008818} & \second{0.003031} \\
 & X & 0.2582 & \best{0.001635} & 0.005505 & \second{0.2586} & 0.009053 & \best{0.000256} & \best{0.2663} & \second{0.005549} & \second{0.003234} \\
 & SIG & 0.2511 & \best{0.001663} & 0.005099 & \second{0.2560} & 0.009319 & \best{0.000233} & \best{0.2660} & \second{0.008131} & \second{0.000475} \\
\midrule
\multirow{3}{*}{Epinions} & LG & \second{0.2304} & \best{0.001283} & 0.005387 & 0.2302 & \second{0.009795} & \best{0.000016} & \best{0.2596} & 0.012305 & \second{0.002940} \\
 & X & 0.2319 & \best{0.001287} & 0.005601 & \second{0.2336} & 0.010272 & \best{0.000080} & \best{0.2613} & \second{0.009533} & \second{0.002836} \\
 & SIG & 0.2128 & \best{0.001565} & 0.006157 & \second{0.2185} & \second{0.010268} & \best{0.000064} & \best{0.2603} & 0.010805 & \second{0.001382} \\
\midrule
\multirow{3}{*}{ML-25M} & LG & \second{0.1163} & \best{0.000366} & \second{0.002272} & 0.1110 & \second{0.003613} & \best{0.001373} & \best{0.1559} & 0.016896 & 0.005194 \\
 & X & \second{0.1117} & \best{0.001144} & \second{0.005596} & 0.1051 & \second{0.003314} & \best{0.001406} & \best{0.1616} & 0.013853 & 0.009617 \\
 & SIG & \second{-0.0762} & \best{0.000091} & \best{0.001200} & -0.0785 & \second{0.004724} & \second{0.001929} & \best{0.0407} & 0.017436 & 0.007939 \\
\bottomrule\end{tabular}
\end{table}

\subsection{External MIR comparison and calibration increments}\label{app:mir-comparison}
\paragraph{Purpose and applicable data.}
MIR\cite{mir} learns reranking through interactions between candidates and user history. It serves as a supplementary external baseline. The main algorithm and our procedure in Table~\ref{tab:utility} remain independent of MIR. The external comparison measures utility and risk of complete procedures. A separate experiment with MIR as reference tests calibration on another frozen recommendation output. We report these two conclusions separately.

Among the six original datasets, the current complete KuaiRec split contains observations whose real timestamps cannot be recovered. Jester has no usable chronology, and the processed Amazon-Music and Epinions data lack verified temporal mappings. These four datasets therefore cannot provide the reliable historical inputs needed for a faithful MIR comparison under the existing complete protocol. For this reason, we compare MIR only on MovieLens-25M and KuaiRand-random, with three backbones and five training seeds each. The other four datasets are marked not applicable. We neither remove observations lacking timestamps nor fabricate their order, and do not count missing comparisons as wins. This does not rule out MIR under other valid temporal protocols for these datasets. This supplement excludes PRM and uses previously examined data domains, so it is not a new independent confirmation.

\paragraph{Implementation and shared evaluation conditions.}
We use the authors' TensorFlow implementation, retaining its network, BCE objective, and Adam optimizer. The sole model correction replaces the positive exponential in historical distance by the negative decay specified in Equation 5 of the original paper, avoiding exponential overflow. Inputs are adapted to actual user, item, category, and history fields. Training feedback forms chronological target blocks of at most eighty items per user. History contains at most thirty positive training interactions strictly earlier than the first timestamp in each target block. These are reconstructed offline lists, not a direct replication of the original exposure sessions. Validation and evaluation labels never enter the histories.

Each MIR model trains for thirty epochs with embedding dimension sixteen, hidden dimension sixty-four, and MLP widths $500/200/80$. Batch size is sixteen, learning rate is $0.00002$, the L2 coefficient is $0.00001$, the gradient norm limit is five, and keep probability is $0.8$. Checkpoints maximize NDCG@20 on the complete original validation set, with ties resolved in favor of the earlier epoch. Of the $27038$ KuaiRand training blocks, $24235$ have no earlier positive history and use the original padding interface. All $12626$ eligible evaluation users have real training histories, averaging $14.4658$ items. This difference in history coverage between training and evaluation is relevant to interpreting performance across datasets.

Evaluation uses the same top eighty observed candidates from a frozen backbone. MIR returns its own top twenty without our heads, proposals, or risk calibration. Methods share final evaluation users, feedback, and metric denominators. MIR uses original training and validation feedback, whereas our procedure additionally uses $F$ for fitting, $S$ for selection, and $C$ for calibration. The external comparison therefore does not match information or training cost to isolate calibration. We average forty common assignments within each seed before constructing approximate $t_4$ intervals across five seeds. Intervals are pointwise and unadjusted for multiple comparisons. Repeated assignments are not independent replicates.

\paragraph{External comparison of complete procedures.}
Both downside losses in Table~\ref{tab:mir-external} use the original backbone Top20 as reference. MIR is not constrained by our risk budgets. MIR has higher utility on all three MovieLens backbones, while our independent procedure has higher utility on all three KuaiRand backbones. Both mean MIR downside losses exceed $0.02/0.01$ in all six settings. These results describe utility and risk under different objectives. They neither negate MIR's utility advantage nor establish dominance under identical constraints from risk differences alone.

\begin{table}[!htb]\centering\footnotesize\setlength{\tabcolsep}{3pt}
\caption{Original MIR versus the complete ACRC recommendation procedure on common-support users. ACRC is independent of MIR. Both downside losses use backbone Top20 as reference. Utility reports five-seed means $\pm$ approximate 95\% interval half-widths. Other entries are means. Within each dataset and backbone, red and blue mark the best and second-best displayed means. Colors imply neither equal risk constraints nor statistical significance.}\label{tab:mir-external}
\begin{tabular}{@{}lllrrrrr@{}}\toprule
Dataset & Backbone & Method & $U\uparrow$ & $D_U\downarrow$ & $D_R\downarrow$ & NDCG$\uparrow$ & Recall$\uparrow$\\\midrule
\multirow{6}{*}{ML-25M} & \multirow{2}{*}{LightGCN} & MIR\cite{mir} & \best{$0.1882\,{\scriptstyle\pm\,0.0014}$} & \second{$0.04465$} & \second{$0.01927$} & \best{$0.6540$} & \best{$0.4864$}\\
 &  & \acrcours{} & \second{$0.1559\,{\scriptstyle\pm\,0.0025}$} & \best{$0.01690$} & \best{$0.00519$} & \second{$0.6370$} & \second{$0.4672$}\\
\cmidrule(l){2-8}
 & \multirow{2}{*}{XSimGCL} & MIR\cite{mir} & \best{$0.1914\,{\scriptstyle\pm\,0.0009}$} & \second{$0.03901$} & \second{$0.01688$} & \best{$0.6565$} & \best{$0.4887$}\\
 &  & \acrcours{} & \second{$0.1616\,{\scriptstyle\pm\,0.0115}$} & \best{$0.01385$} & \best{$0.00962$} & \second{$0.6411$} & \second{$0.4774$}\\
\cmidrule(l){2-8}
 & \multirow{2}{*}{SIGformer} & MIR\cite{mir} & \best{$0.1928\,{\scriptstyle\pm\,0.0015}$} & \second{$0.02532$} & \second{$0.01290$} & \best{$0.6554$} & \best{$0.4856$}\\
 &  & \acrcours{} & \second{$0.0407\,{\scriptstyle\pm\,0.0151}$} & \best{$0.01744$} & \best{$0.00794$} & \second{$0.5674$} & \second{$0.4173$}\\
\midrule
\multirow{6}{*}{KuaiRand} & \multirow{2}{*}{LightGCN} & MIR\cite{mir} & \second{$-0.5462\,{\scriptstyle\pm\,0.0184}$} & \second{$0.09503$} & \second{$0.06821$} & \second{$0.3668$} & \second{$0.4946$}\\
 &  & \acrcours{} & \best{$-0.4846\,{\scriptstyle\pm\,0.0007}$} & \best{$0.00758$} & \best{$0.00986$} & \best{$0.4040$} & \best{$0.5041$}\\
\cmidrule(l){2-8}
 & \multirow{2}{*}{XSimGCL} & MIR\cite{mir} & \second{$-0.5515\,{\scriptstyle\pm\,0.0149}$} & \second{$0.09419$} & \second{$0.06884$} & \second{$0.3640$} & \best{$0.4933$}\\
 &  & \acrcours{} & \best{$-0.4977\,{\scriptstyle\pm\,0.0015}$} & \best{$0.00924$} & \best{$0.00994$} & \best{$0.3929$} & \second{$0.4893$}\\
\cmidrule(l){2-8}
 & \multirow{2}{*}{SIGformer} & MIR\cite{mir} & \second{$-0.5450\,{\scriptstyle\pm\,0.0031}$} & \second{$0.12449$} & \second{$0.10119$} & \second{$0.3663$} & \second{$0.4948$}\\
 &  & \acrcours{} & \best{$-0.4555\,{\scriptstyle\pm\,0.0005}$} & \best{$0.01225$} & \best{$0.00964$} & \best{$0.4311$} & \best{$0.5412$}\\
\bottomrule\end{tabular}
\end{table}

\paragraph{Supplementary calibration with MIR as reference.}
In a separate experiment, we freeze MIR models and scores and use MIR20 as both reference and fallback. Our heads use only $F200$, modification selection uses $S400$, and calibration uses $C200$. Proposals can change item membership, order, and length within the common eighty-item pool, and include MIR prefixes. MovieLens and KuaiRand share the same proposal and selection configuration. No variant is chosen per backbone using test results. This is a supplementary application of our postprocessing, not original MIR or a replacement for the main algorithm and independent procedure in Table~\ref{tab:utility}.

Table~\ref{tab:mir-calibration} fixes fitted heads, proposals, the selected rule, loss ranges, and evaluation users within every assignment, changing only raw, RB, or $w_+$. All risks here use MIR20 as reference and are not directly comparable to the backbone-Top20 risks in Table~\ref{tab:mir-external}. Original MIR has zero downside relative to itself. Equal budgets do not mean equal realized risk. With reference and proposals frozen, the existing sampling-without-replacement conditions and loss ranges allow Theorem~\ref{thm:dual-risk} to apply. It does not guarantee improvement for each user or risk within budget in every assignment.

\begin{table}[!htb]\centering\footnotesize\setlength{\tabcolsep}{3pt}
\caption{Calibration comparison with frozen MIR20 as reference. Fitted heads, proposals, and the selected modification are identical within each setting. Only the calibrator changes. Utility reports five-seed means $\pm$ approximate 95\% interval half-widths. Other entries are means. The utility and Recall downside budgets are $0.02$ and $0.01$. Within each dataset and backbone, red and blue mark the best and second-best displayed means, with ties sharing a color. Length is descriptive. This comparison isolates calibration and does not replace the main procedure.}\label{tab:mir-calibration}
\begin{tabular}{@{}lllrrrrrr@{}}\toprule
Dataset & Backbone & Method & $U\uparrow$ & $D_U\downarrow$ & $D_R\downarrow$ & NDCG$\uparrow$ & Recall$\uparrow$ & Length\\\midrule
\multirow{9}{*}{ML-25M} & \multirow{3}{*}{LightGCN} & Raw & $0.1978\,{\scriptstyle\pm\,0.0029}$ & \best{$0.01241$} & \best{$0.00720$} & $0.6573$ & $0.4878$ & $19.896$\\
 &  & RB & \second{$0.1985\,{\scriptstyle\pm\,0.0032}$} & \second{$0.01316$} & \second{$0.00764$} & \second{$0.6575$} & \second{$0.4880$} & $19.891$\\
 &  & \acrcours{} & \best{$0.1996\,{\scriptstyle\pm\,0.0038}$} & $0.01458$ & $0.00836$ & \best{$0.6581$} & \best{$0.4885$} & $19.890$\\
\cmidrule(l){2-9}
 & \multirow{3}{*}{XSimGCL} & Raw & $0.2030\,{\scriptstyle\pm\,0.0036}$ & \best{$0.01295$} & \best{$0.00715$} & $0.6617$ & $0.4920$ & $19.943$\\
 &  & RB & \second{$0.2038\,{\scriptstyle\pm\,0.0039}$} & \second{$0.01389$} & \second{$0.00766$} & \second{$0.6621$} & \second{$0.4923$} & $19.941$\\
 &  & \acrcours{} & \best{$0.2054\,{\scriptstyle\pm\,0.0044}$} & $0.01563$ & $0.00857$ & \best{$0.6630$} & \best{$0.4930$} & $19.940$\\
\cmidrule(l){2-9}
 & \multirow{3}{*}{SIGformer} & Raw & $0.1993\,{\scriptstyle\pm\,0.0011}$ & \best{$0.00980$} & \best{$0.00905$} & \best{$0.6504$} & \best{$0.4776$} & $19.613$\\
 &  & RB & \second{$0.1994\,{\scriptstyle\pm\,0.0011}$} & \second{$0.00998$} & \second{$0.00926$} & \second{$0.6503$} & \second{$0.4774$} & $19.604$\\
 &  & \acrcours{} & \best{$0.1996\,{\scriptstyle\pm\,0.0012}$} & $0.01038$ & $0.00949$ & \second{$0.6503$} & \second{$0.4774$} & $19.599$\\
\midrule
\multirow{9}{*}{KuaiRand} & \multirow{3}{*}{LightGCN} & Raw & $-0.5244\,{\scriptstyle\pm\,0.0132}$ & \best{$0.01689$} & \best{$0.00804$} & $0.3727$ & \best{$0.4865$} & $19.664$\\
 &  & RB & \second{$-0.5238\,{\scriptstyle\pm\,0.0128}$} & \second{$0.01734$} & \second{$0.00823$} & \second{$0.3729$} & \second{$0.4863$} & $19.656$\\
 &  & \acrcours{} & \best{$-0.5234\,{\scriptstyle\pm\,0.0125}$} & $0.01764$ & $0.00833$ & \best{$0.3731$} & $0.4862$ & $19.652$\\
\cmidrule(l){2-9}
 & \multirow{3}{*}{XSimGCL} & Raw & $-0.5281\,{\scriptstyle\pm\,0.0094}$ & \best{$0.01706$} & \best{$0.00802$} & $0.3712$ & \best{$0.4853$} & $19.664$\\
 &  & RB & \second{$-0.5273\,{\scriptstyle\pm\,0.0090}$} & \second{$0.01764$} & \second{$0.00824$} & \second{$0.3715$} & \second{$0.4851$} & $19.655$\\
 &  & \acrcours{} & \best{$-0.5269\,{\scriptstyle\pm\,0.0085}$} & $0.01795$ & $0.00836$ & \best{$0.3716$} & $0.4850$ & $19.650$\\
\cmidrule(l){2-9}
 & \multirow{3}{*}{SIGformer} & Raw & $-0.5223\,{\scriptstyle\pm\,0.0025}$ & \best{$0.01712$} & \best{$0.00779$} & $0.3739$ & \best{$0.4870$} & $19.675$\\
 &  & RB & \second{$-0.5215\,{\scriptstyle\pm\,0.0026}$} & \second{$0.01775$} & \second{$0.00804$} & \second{$0.3742$} & \second{$0.4867$} & $19.664$\\
 &  & \acrcours{} & \best{$-0.5210\,{\scriptstyle\pm\,0.0027}$} & $0.01811$ & $0.00818$ & \best{$0.3744$} & $0.4866$ & $19.659$\\
\bottomrule\end{tabular}
\end{table}

\paragraph{Calibration contribution and total combination gains.}
Table~\ref{tab:mir-increments} separates two effects. The combination's utility gain over MIR includes fitted heads, proposal construction, selection, and final calibration. Only the last two columns compare $w_+$ against raw and RB for the same modification, isolating the calibration increment conditional on the upstream procedure. Combination utility exceeds original MIR in all six settings. All six pointwise intervals for same-rule $w_+$ versus raw exclude zero. Against RB, the KuaiRand LG and X intervals include zero, so a statistically distinguishable improvement is not established in every setting. These internal operator comparisons across two data domains support applicability of calibration without attributing MIR's learned capability or the total combination gain to calibration.

Both mean downside losses of the joint combination are within budget in all six settings, but quality tradeoffs remain. On MovieLens with SIG, Recall decreases from $0.485602$ to $0.477370$ and NDCG from $0.655438$ to $0.650271$. KuaiRand Recall decreases by $0.008333$, $0.008361$, and $0.008179$ across the three backbones. Combination utility there remains below our original independent procedure. A supplementary control that fixes length at twenty and changes only order and item membership increases mean utility, NDCG, and Recall in all six settings. This does not justify selecting the best variant per backbone after observing test outcomes. Utility comparisons have the same directions on all eligible users. That cohort overlaps with the common-support users and is not an independent replicate.

\begingroup\footnotesize\setlength{\tabcolsep}{3.3pt}
\begin{table}[!htb]\centering\footnotesize
\caption{Paired utility differences in the MIR supplement, reported as five-seed means and approximate 95\% interval half-widths. The first contrast includes the complete modification procedure. The last two fix the selected modification and change only calibration. Columns have different comparators and are not ranked against one another.}\label{tab:mir-increments}
\begin{tabular}{@{}llrrr@{}}
\toprule
Dataset & Backbone & Combination vs. MIR & $w_+$ vs. raw & $w_+$ vs. RB\\\midrule

\multirow{3}{*}{ML-25M} & LG & $+0.01145\,{\scriptstyle\pm\,0.00469}$ & $+0.001776\,{\scriptstyle\pm\,0.001046}$ & $+0.001149\,{\scriptstyle\pm\,0.000665}$\\*
 & X & $+0.01401\,{\scriptstyle\pm\,0.00381}$ & $+0.002444\,{\scriptstyle\pm\,0.000783}$ & $+0.001586\,{\scriptstyle\pm\,0.000526}$\\*
 & SIG & $+0.00682\,{\scriptstyle\pm\,0.00191}$ & $+0.000299\,{\scriptstyle\pm\,0.000049}$ & $+0.000173\,{\scriptstyle\pm\,0.000077}$\\
\midrule
\multirow{3}{*}{KuaiRand} & LG & $+0.02281\,{\scriptstyle\pm\,0.00638}$ & $+0.000965\,{\scriptstyle\pm\,0.000806}$ & $+0.000388\,{\scriptstyle\pm\,0.000422}$\\*
 & X & $+0.02461\,{\scriptstyle\pm\,0.00645}$ & $+0.001193\,{\scriptstyle\pm\,0.000852}$ & $+0.000434\,{\scriptstyle\pm\,0.000457}$\\*
 & SIG & $+0.02395\,{\scriptstyle\pm\,0.00341}$ & $+0.001271\,{\scriptstyle\pm\,0.000432}$ & $+0.000465\,{\scriptstyle\pm\,0.000246}$\\
\bottomrule
\end{tabular}
\end{table}
\endgroup

\subsection{Risk-matched descriptive comparisons}\label{app:frontier}
\begin{figure*}[!t]
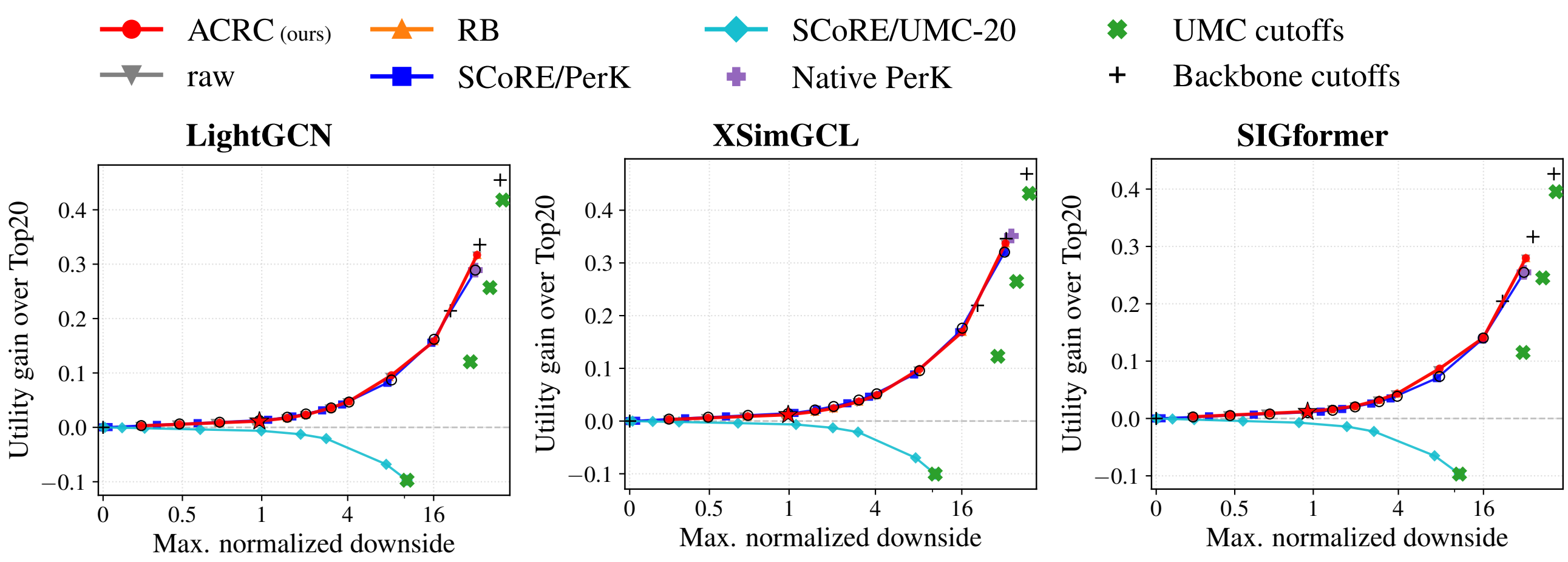

\centering
\begin{minipage}{0.98\linewidth}\begingroup\makeatletter
\resizebox{\linewidth}{!}{%
\begingroup%
\makeatletter%
%
%
\makeatother%
\endgroup%

}%
\endgroup
\end{minipage}
\caption{Risk and utility on KuaiRand. The horizontal axis is maximum normalized downside, linear to one and logarithmic above one. Gain is relative to Top20. Lines join eleven predefined budgets, stars mark our default, and open circles mark best feasible original-path or native-anchor states. The scan is descriptive and introduces no interpolated algorithm.}\label{fig:frontier-main}
\end{figure*}

All seventy-five full models on the original five datasets, three backbones and five training seeds stay frozen. Each retains forty common-support assignments. Only physical budgets change: $\epsilon_U=0.02$ and $\epsilon_R=0.01$ are multiplied by predeclared $s\in\{0.25,0.5,0.75,1,1.5,2,3,4,8,16,32\}$. This fixed-ratio path does not cover every two-dimensional budget pair. Backbones, probability heads, candidates, losses, costs and seeds are not tuned. Each point reruns the original $S$ selector and final calibration. raw, RB and our method share the point-specific selected rule and differ only in the final operator.

SCoRE uses the official version and default random forest from the preceding section, with native PerK or UMC probabilities. Two risk scores are fitted for each $S$. Their targets are budget independent, so frozen scores are reused across budgets. Every point still calls the unchanged official function with $\gamma=\alpha$ and intersects the two native selection sets. No proposed component enters an external method. Native PerK and UMC retain their objectives, order and length decisions. Their fixed anchors do not carry our two-risk requirement.

Missing feedback remains missing. The budget-sensitivity study uses only the original five datasets and excludes the MovieLens-32M confirmation feedback.

Forty assignments are averaged within each seed before averaging five seeds. The graphical coordinate $\max\{\widehat R_U/0.02,\widehat R_R/0.01\}$ is only a display of the two separately reported risks. Lines join declared budgets without adding interpolated algorithms, deleting nonmonotone points or enforcing monotonicity.

\par\medskip\noindent\subsubsection{Descriptive comparisons under both empirical risk ceilings}
For each proposed point, we scan the complete unchanged SCoRE paths on $0.25\le s\le32$, all eight native backbone and UMC outputs at $k=1,5,10,20$, and native PerK. Under both empirical risk ceilings we retain the highest utility. Default-point higher/equal/lower counts are 13/0/2, with 149/0/16 across all 165 proposed points. Every unfavorable comparison is on KuaiRand with LightGCN or XSimGCL. This test-summary scan is descriptive, supplies no deployable threshold and is not an independent significance test. Fixed budget ratios, input predictors and default $\gamma=\alpha$ delimit its scope. It does not establish dominance of the entire method class. Table~\ref{tab:frontier-matched} reports every default setting.
The eleven-point grid with all nine native anchors gives default counts 15/0/0 and all-budget counts 165/0/0. Complete paths reveal better external states between grid points. We therefore base the descriptive conclusion on the complete-path comparison. Both grid and complete-path records are retained, including unfavorable comparisons.

Complete paths enumerate every decision transition of the original function. Within each assignment and risk, a finite set of statistics computed from calibration losses and risk scores determines selection. We preserve floating-point comparisons and verify the original function immediately before, at and after every threshold, then intersect the two risk conditions. Sorting transitions by a common budget multiplier gives aggregate utility and both risks at every state. A state uses one multiplier shared across all five seeds and forty assignments, rather than choosing a separate budget for each assignment. This enumeration leaves prediction scores and decision functions unchanged.

\begin{table}[!htb]\centering\small\setlength{\tabcolsep}{2.8pt}
\caption{Default-setting comparisons. Both external empirical risks are no larger than ours, whose default risks appear in Table~\ref{tab:external-risk}. Fixed denotes a native anchor. Displayed multipliers are rounded, while exact rational budgets and states are retained. Differences subtract external utility from ours. Colors compare utility only and do not imply continuous-frontier dominance.}\label{tab:frontier-matched}
\begin{tabular}{@{}lllrrrrrr@{}}\toprule
Dataset & Model & Comparator & $s$ & \acrcours{} $U\uparrow$ & External $U\uparrow$ & $\Delta U\uparrow$ & External $R_U\downarrow$ & External $R_R\downarrow$\\\midrule
\multirow{3}{*}{KuaiRec} & LG & SCoRE/U & 1.3167 & \best{0.5391} & \second{0.2309} & +0.308189 & 0.016655 & 0.000191\\
 & X & Top20 & fixed & \best{0.5221} & \second{0.2532} & +0.268899 & 0.000000 & 0.000000\\
 & SIG & PerK & fixed & \best{0.7543} & \second{0.6812} & +0.073020 & 0.000055 & 0.000012\\
\midrule
\multirow{3}{*}{KuaiRand} & LG & SCoRE/P & 1.3551 & \second{-0.4846} & \best{-0.4836} & -0.000992 & 0.000766 & 0.009852\\
 & X & SCoRE/P & 1.4101 & \second{-0.4977} & \best{-0.4954} & -0.002220 & 0.000694 & 0.009937\\
 & SIG & SCoRE/P & 1.3282 & \best{-0.4555} & \second{-0.4573} & +0.001813 & 0.000762 & 0.009632\\
\midrule
\multirow{3}{*}{Music} & LG & SCoRE/U & 0.5055 & \best{0.2663} & \second{0.2603} & +0.005976 & 0.000216 & 0.000000\\
 & X & SCoRE/P & 0.2500 & \best{0.2663} & \second{0.2595} & +0.006732 & 0.000000 & 0.000000\\
 & SIG & SCoRE/U & 0.9177 & \best{0.2660} & \second{0.2553} & +0.010774 & 0.008129 & 0.000160\\
\midrule
\multirow{3}{*}{Epinions} & LG & SCoRE/P & 0.2500 & \best{0.2596} & \second{0.2313} & +0.028303 & 0.000000 & 0.000000\\
 & X & SCoRE/U & 0.9637 & \best{0.2613} & \second{0.2335} & +0.027743 & 0.009518 & 0.000065\\
 & SIG & SCoRE/U & 1.0291 & \best{0.2603} & \second{0.2188} & +0.041422 & 0.010804 & 0.000072\\
\midrule
\multirow{3}{*}{ML-25M} & LG & SCoRE/P & 16.7583 & \best{0.1559} & \second{0.1164} & +0.039520 & 0.000373 & 0.002353\\
 & X & SCoRE/P & 1.3954 & \best{0.1616} & \second{0.1150} & +0.046636 & 0.001684 & 0.009604\\
 & SIG & PerK & fixed & \best{0.0407} & \second{-0.0761} & +0.116829 & 0.000097 & 0.001273\\
\bottomrule\end{tabular}
\end{table}

\section{Calibration Comparisons with Fixed Proposals}\label{app:module-complete}
\subsection{Information-matched calibration comparisons}\label{app:matched-information}
\label{app:proposal-calibration-protocol}
The factorial design uses five proposals. The first reuses the stochastic rule already chosen by the main pipeline on $S$. The second reorders the original Top20 items. The third reorders and selects items from the first 80 available items ranked by the frozen backbone. The latter two retain the reference list length. Each chooses between the complete local-head and global-head rerankings using mean utility on $S$, with both heads fitted only on $F$. They share the main pipeline's fitting feedback and candidate opportunities and test how much gain ordinary upstream changes can explain. They do not represent newly trained stronger complete recommenders. The fourth and fifth proposals reuse the frozen outputs of original PerK-PDCG and UMC-Top20.

For native PerK prefixes the utility-downside bound is one, and for UMC reordering it is two. The Recall-downside bound is one for both. Candidate proposals retain their frozen list-structure bounds. Raw, RB and $w_+$ independently choose a common multiplier for both budgets from $\{0.5,0.75,1\}$ using the same four $S$ rounds. Final scalar gates use exact rational arithmetic and downward quantization. Comparisons between independently tuned methods can use different multipliers and therefore do not inherit pointwise dominance at an identical budget.

\paragraph{Comparison inputs and configurations.}
We reuse ninety complete models from six datasets, three backbones, and five training seeds, with forty role splits per model. The five proposal classes are our selected proposal, ordinary reranking within Top20, ordinary reranking over the candidate pool, native PerK proposals, and native UMC proposals. Proposals, evaluation users, risk definitions, and list-structure bounds remain fixed. Any extra calibration applied to an external proposal is a module comparison. Unmodified recommendation methods are reported separately.
The base risk scorer uses 129 unlabeled features designed for this module comparison, rather than a feature set prescribed by the SCoRE paper. These features include backbone scores and position masks for the reference and two candidate lists, pool statistics, and the mixture coefficient. The augmented version adds 120 probabilities from two existing heads at twenty positions of all three lists, giving 249 features. Both heads were fitted only on the 200 users in $F$ and remain frozen. They are not refitted on $S,C,T$. Historical training-interaction statistics may enter their inputs. Current $C/T$ feedback, realized gains, and total relevant-item counts do not.
The risk scorers use only the 400 users in $S$. Each of four selection rounds uses 100 users for fitting, 200 for pseudo-calibration, and 100 for evaluation. Forests contain 100 trees. Minimum leaf sizes are selected from $\{1,5,20\}$, with a constant-score alternative. A common multiplier for both risk budgets is selected from $\{0.5,0.75,1\}$. Ties favor the larger multiplier and then the prespecified leaf-size order. The final scorer is fitted on all of $S$, calibrated on the 200 independent users in $C$, and evaluated on the remaining users in $T$.
The three configurations use 249 features with the default configuration, 249 features tuned on $S$, and an $S$-only choice between the best configurations of the 129-feature and 249-feature families. The default uses leaf size one and multiplier one. A feature-family tie retains the original 129 features. The third configuration is the risk-prediction control with both feature families available. The first two are sensitivity analyses. The authors' SCoRE-MDR implementation\cite{score}, $\gamma=\alpha$, and the intersection of the two risk decisions remain unchanged. Unselected users receive the original Top20 rather than an empty list.
Losses are divided by the same frozen bounds before calling the official interface at normalized risk levels. Reported risks remain on their physical scales. For a randomized proposal, we take the positive loss for each action before averaging over its randomization. We do not take the positive part of the mean gain. The design crosses 90 models, 40 splits, five proposals, and three configurations.
The same $S$ participates in upstream proposal selection and downstream configuration selection, so its internal scores can be optimistic. Final $C/T$ labels are not used for selection. These already-observed domains do not provide a new independent confirmation. We first average repeated splits and the specified settings within each training seed, then form approximate 95\% $t_4$ intervals from five paired seed means. Pointwise intervals are not adjusted for multiplicity.
\begingroup
\small\setlength{\tabcolsep}{4pt}

\begin{table}[!htb]
\centering\small\setlength{\tabcolsep}{4pt}
\caption{Macro comparison across five fixed proposal classes using risk-prediction SCoRE. Both $w_+$ and original-feature SCoRE are tuned only on $S$. The table includes all three configurations. Colors rank the five calibration methods within each row. Lower risk does not by itself imply higher utility.}\label{tab:matched-information-macro}
\begin{tabular}{@{}llrrrrr@{}}
\toprule
Proposal & Metric & \acrcours{} & \multicolumn{4}{c}{SCoRE}\\
\cmidrule(lr){4-7}
 &  &  & \shortstack{Original\\features} & \shortstack{Augmented\\default} & \shortstack{Augmented\\tuned} & \shortstack{Feature\\selection}\\
\midrule
\multirow{3}{*}{Selected proposal} & $U\uparrow$ & \best{$0.18265$} & $0.16470$ & $0.16426$ & $0.16562$ & \second{$0.16572$} \\
 & $D_U\downarrow$ & $0.01249$ & \best{$0.00929$} & \second{$0.00941$} & $0.00948$ & \second{$0.00941$} \\
 & $D_R\downarrow$ & $0.00498$ & \best{$0.00319$} & $0.00324$ & $0.00324$ & \second{$0.00320$} \\
\midrule
\multirow{3}{*}{Within-Top20 reranking} & $U\uparrow$ & \best{$0.14210$} & $0.14074$ & $0.14078$ & \second{$0.14086$} & $0.14081$ \\
 & $D_U\downarrow$ & $0.01516$ & $0.01370$ & $0.01426$ & \best{$0.01364$} & \second{$0.01369$} \\
 & $D_R\downarrow$ & \best{$0.00000$} & \best{$0.00000$} & \best{$0.00000$} & \best{$0.00000$} & \best{$0.00000$} \\
\midrule
\multirow{3}{*}{Pool reranking} & $U\uparrow$ & \best{$0.17972$} & $0.15884$ & $0.15888$ & $0.15998$ & \second{$0.16025$} \\
 & $D_U\downarrow$ & $0.01356$ & \best{$0.00800$} & $0.00823$ & \second{$0.00806$} & $0.00814$ \\
 & $D_R\downarrow$ & $0.00449$ & $0.00182$ & $0.00191$ & \best{$0.00172$} & \second{$0.00180$} \\
\midrule
\multirow{3}{*}{PerK proposal} & $U\uparrow$ & $0.11394$ & $0.11431$ & $0.11427$ & \best{$0.11460$} & \second{$0.11451$} \\
 & $D_U\downarrow$ & $0.00134$ & \best{$0.00105$} & $0.00186$ & \second{$0.00113$} & $0.00118$ \\
 & $D_R\downarrow$ & $0.00495$ & \best{$0.00209$} & $0.00432$ & \second{$0.00246$} & $0.00256$ \\
\midrule
\multirow{3}{*}{UMC proposal} & $U\uparrow$ & $0.11298$ & $0.11415$ & $0.11464$ & \best{$0.11538$} & \second{$0.11501$} \\
 & $D_U\downarrow$ & $0.01154$ & \best{$0.00416$} & $0.00722$ & \second{$0.00463$} & $0.00491$ \\
 & $D_R\downarrow$ & $0.00294$ & \best{$0.00056$} & $0.00107$ & \second{$0.00058$} & $0.00067$ \\
\bottomrule
\end{tabular}
\end{table}

\endgroup
\begingroup
\small\setlength{\tabcolsep}{4pt}
\begin{table}[!htb]\centering\small
\caption{$S$-selected feature family. $\Delta U=U_{w_+}-U_{\mathrm{SCoRE}}$, with means and paired five-seed intervals. Risk columns report $D_U/D_R$. Values are means $\pm$ 95\% CI half-widths.}\label{tab:matched-information-score-info-select-S}
\begin{tabular}{llrrrcc}
\toprule
Dataset & Model & $U_{w_+}\uparrow$ & $U_{\rm SCoRE}\uparrow$ & $\Delta U\uparrow$ & $w_+$ risks & SCoRE risks \\
\midrule
\multirow{3}{*}{KuaiRec} & LightGCN & $0.5391$ & $0.4721$ & $+0.066984\,{\scriptstyle\pm\,0.016319}$ & $0.01667\,/\,0.00021$ & $0.01134\,/\,0.00015$ \\*
 & XSimGCL & $0.5221$ & $0.4571$ & $+0.064932\,{\scriptstyle\pm\,0.004587}$ & $0.01655\,/\,0.00020$ & $0.01148\,/\,0.00014$ \\*
 & SIGformer & $0.7543$ & $0.7477$ & $+0.006598\,{\scriptstyle\pm\,0.002882}$ & $0.01501\,/\,0.00017$ & $0.01279\,/\,0.00015$ \\
\midrule
\multirow{3}{*}{KuaiRand} & LightGCN & $-0.4846$ & $-0.4867$ & $+0.002078\,{\scriptstyle\pm\,0.000918}$ & $0.00758\,/\,0.00986$ & $0.00614\,/\,0.00584$ \\*
 & XSimGCL & $-0.4977$ & $-0.4986$ & $+0.000952\,{\scriptstyle\pm\,0.000741}$ & $0.00924\,/\,0.00994$ & $0.00770\,/\,0.00630$ \\*
 & SIGformer & $-0.4555$ & $-0.4569$ & $+0.001464\,{\scriptstyle\pm\,0.000202}$ & $0.01225\,/\,0.00964$ & $0.01037\,/\,0.00684$ \\
\midrule
\multirow{3}{*}{Music} & LightGCN & $0.2663$ & $0.2658$ & $+0.000510\,{\scriptstyle\pm\,0.000317}$ & $0.00882\,/\,0.00303$ & $0.00806\,/\,0.00246$ \\*
 & XSimGCL & $0.2663$ & $0.2655$ & $+0.000823\,{\scriptstyle\pm\,0.000897}$ & $0.00555\,/\,0.00323$ & $0.00487\,/\,0.00262$ \\*
 & SIGformer & $0.2660$ & $0.2653$ & $+0.000714\,{\scriptstyle\pm\,0.000585}$ & $0.00813\,/\,0.00048$ & $0.00785\,/\,0.00045$ \\
\midrule
\multirow{3}{*}{Epinions} & LightGCN & $0.2596$ & $0.2553$ & $+0.004320\,{\scriptstyle\pm\,0.002238}$ & $0.01231\,/\,0.00294$ & $0.01055\,/\,0.00239$ \\*
 & XSimGCL & $0.2613$ & $0.2582$ & $+0.003056\,{\scriptstyle\pm\,0.001125}$ & $0.00953\,/\,0.00284$ & $0.00865\,/\,0.00234$ \\*
 & SIGformer & $0.2603$ & $0.2550$ & $+0.005292\,{\scriptstyle\pm\,0.003310}$ & $0.01081\,/\,0.00138$ & $0.00962\,/\,0.00112$ \\
\midrule
\multirow{3}{*}{ML-25M} & LightGCN & $0.1559$ & $0.1507$ & $+0.005259\,{\scriptstyle\pm\,0.001261}$ & $0.01690\,/\,0.00519$ & $0.01402\,/\,0.00425$ \\*
 & XSimGCL & $0.1616$ & $0.1441$ & $+0.017500\,{\scriptstyle\pm\,0.001932}$ & $0.01385\,/\,0.00962$ & $0.00813\,/\,0.00456$ \\*
 & SIGformer & $0.0407$ & $-0.0152$ & $+0.055988\,{\scriptstyle\pm\,0.005362}$ & $0.01744\,/\,0.00794$ & $0.00845\,/\,0.00343$ \\
\midrule
\multirow{3}{*}{Jester} & LightGCN & $0.3244$ & $0.3125$ & $+0.011942\,{\scriptstyle\pm\,0.001980}$ & $0.01440\,/\,0.00778$ & $0.01047\,/\,0.00534$ \\*
 & XSimGCL & $0.3537$ & $0.3400$ & $+0.013692\,{\scriptstyle\pm\,0.002006}$ & $0.01320\,/\,0.00702$ & $0.00910\,/\,0.00476$ \\*
 & SIGformer & $0.2938$ & $0.2512$ & $+0.042601\,{\scriptstyle\pm\,0.008315}$ & $0.01668\,/\,0.00823$ & $0.00973\,/\,0.00446$ \\
\bottomrule
\end{tabular}
\end{table}
\endgroup

\paragraph{Limits of the result.}
For our selected proposal, all eighteen mean utility differences against information-matched risk-prediction SCoRE are positive. The interval for Music with XSimGCL contains zero. Macro realized risks increase from $0.009406/0.003199$ for information-matched SCoRE to $0.012495/0.004983$ for our method. Both remain below their respective limits, which does not certify each split or each user.
For PerK proposals, the utility difference of $w_+$ from the primary information-matched control is $-0.000566\,{\scriptstyle\pm\,0.000105}$. For UMC proposals it is $-0.002023\,{\scriptstyle\pm\,0.002228}$. Higher calibration activation does not universally improve signed recommendation gains. Matching permitted information does not exhaust all scorer classes or search budgets.

\subsubsection{Reward-aware SCoRE with unchanged MDR decisions}\label{app:reward-score-comparison}
SCoRE also recommends the predicted risk-to-reward ratio for MDR selection\cite{score}. We evaluate this missing score family using the same selected proposal, two frozen $F$ heads, $S400$, $C200$, and $T$ users. Original and augmented feature families contain 129 and 249 features. The recommendation backbones and original external baselines are unchanged.

Signed paired gain cannot directly serve as a nonnegative reward. Let $G_{u,a}$ be the utility change from proposal action $a$. At $\lambda=1$, define
\begin{equation}
r_u=\frac12\mathbb E_a(G_{u,a})_+\in[0,1],\qquad
g_u=2r_u-x^U_u,\qquad s_k(u)=\widehat l_k(u)/\widehat r(u).
\label{eq:score-positive-reward}
\end{equation}
For a constraint requiring calibration, $\widehat l_k$ predicts the normalized loss $x^k_u/H_k$. The positive part is taken before averaging over proposal actions. No constant is added to signed gain. A zero predicted reward receives score $+\infty$, including the $0/0$ case, with ties retained. Each risk score and reward predictor is fitted only on $S$. Four fixed folds use 100 users for fitting, 200 for pseudo-calibration, and 100 for validation. We retain the existing forest size, leaf grid and budget multipliers, and select by signed net gain on $S$. The final comparison additionally chooses between the existing risk-prediction family and the new ratio family on $S$, preferring risk prediction on ties.

Both risk decisions call the unchanged official MDR implementation with $\gamma=\alpha$ and are intersected. Conditional on the $S$-selected score functions, the existing validity argument applies. The original nonnegative-reward optimality result does not establish optimality for signed net utility or the intersection of two constraints. This is a score comparison within the fixed-proposal protocol, distinct from the original PerK and UMC pipelines in Table~\ref{tab:utility}.

\begin{table}[htb]\centering\small
\caption{Same-proposal macro results with both SCoRE score families. All choices precede final calibration. Means equally weight six domains and three backbones within five training seeds.}\label{tab:reward-score-macro}
\begin{tabular}{lrrr}\toprule Calibrator & $U\uparrow$ & $D_U\downarrow$ & $D_R\downarrow$\\\midrule
\acrcours{} & \best{$0.1826$} & $0.01249$ & $0.00498$ \\
SCoRE risk prediction & $0.1657$ & \second{$0.00941$} & \best{$0.00320$} \\
SCoRE reward ratio, default & $0.1660$ & \second{$0.00941$} & $0.00330$ \\
SCoRE reward ratio, $S$ & \second{$0.1672$} & $0.00942$ & $0.00328$ \\
SCoRE both score families, $S$ & $0.1667$ & \best{$0.00940$} & \second{$0.00323$} \\\bottomrule\end{tabular}\end{table}
The $S$-only family selection uses the reward ratio in 1627 of 3600 splits. No test-set maximum is reported as the selected method.

\subsection{Raw and RB controls with identical proposals}\label{app:revision-modules}
Heads, selected proposals, action ranges, calibration users and evaluation users are fixed within each contrast. Only the final calibration operator changes. These are internal operator controls, distinct from the information-matched SCoRE comparison above.
\begingroup\small\setlength{\tabcolsep}{3pt}

\begin{table}[!htb]
\centering\small\setlength{\tabcolsep}{4pt}
\caption{All eighteen same-proposal settings at default budgets. Entries are $w_+$ minus raw or RB, with paired five-seed approximate 95\% intervals. Forty assignments are averaged within each seed. Negative and zero-crossing intervals are retained. No multiplicity correction is applied. Values are means $\pm$ 95\% CI half-widths. All three backbone columns give utility increments, higher is better.}\label{tab:revision-setting-ci}
\begin{tabular}{@{}llrrr@{}}
\toprule
Dataset & Control & LightGCN & XSimGCL & SIGformer\\
\midrule
\multirow{2}{*}{KuaiRec} & raw & $+0.05545\,{\scriptstyle\pm\,0.00223}$ & $+0.04606\,{\scriptstyle\pm\,0.00290}$ & $+0.00541\,{\scriptstyle\pm\,0.00055}$\\
 & RB & $+0.04499\,{\scriptstyle\pm\,0.00247}$ & $+0.03996\,{\scriptstyle\pm\,0.00241}$ & $+0.00588\,{\scriptstyle\pm\,0.00173}$\\
\midrule
\multirow{2}{*}{KuaiRand} & raw & $+0.00046\,{\scriptstyle\pm\,0.00005}$ & $+0.00043\,{\scriptstyle\pm\,0.00013}$ & $+0.00045\,{\scriptstyle\pm\,0.00005}$\\
 & RB & $+0.00010\,{\scriptstyle\pm\,0.00008}$ & $+0.00013\,{\scriptstyle\pm\,0.00012}$ & $+0.00016\,{\scriptstyle\pm\,0.00004}$\\
\midrule
\multirow{2}{*}{ML-25M} & raw & $+0.00393\,{\scriptstyle\pm\,0.00049}$ & $+0.00984\,{\scriptstyle\pm\,0.00301}$ & $+0.02183\,{\scriptstyle\pm\,0.00488}$\\
 & RB & $+0.00549\,{\scriptstyle\pm\,0.00047}$ & $+0.00525\,{\scriptstyle\pm\,0.00277}$ & $+0.01002\,{\scriptstyle\pm\,0.00291}$\\
\midrule
\multirow{2}{*}{Jester} & raw & $+0.00985\,{\scriptstyle\pm\,0.00084}$ & $+0.00819\,{\scriptstyle\pm\,0.00079}$ & $+0.01665\,{\scriptstyle\pm\,0.00346}$\\
 & RB & $+0.00877\,{\scriptstyle\pm\,0.00051}$ & $+0.00749\,{\scriptstyle\pm\,0.00068}$ & $+0.00963\,{\scriptstyle\pm\,0.00053}$\\
\midrule
\multirow{2}{*}{Music} & raw & $+0.00014\,{\scriptstyle\pm\,0.00006}$ & $+0.00016\,{\scriptstyle\pm\,0.00018}$ & $+0.00003\,{\scriptstyle\pm\,0.00002}$\\
 & RB & $+0.00040\,{\scriptstyle\pm\,0.00011}$ & $+0.00028\,{\scriptstyle\pm\,0.00024}$ & $+0.00085\,{\scriptstyle\pm\,0.00012}$\\
\midrule
\multirow{2}{*}{Epinions} & raw & $+0.00182\,{\scriptstyle\pm\,0.00071}$ & $+0.00106\,{\scriptstyle\pm\,0.00044}$ & $+0.00231\,{\scriptstyle\pm\,0.00163}$\\
 & RB & $+0.00313\,{\scriptstyle\pm\,0.00025}$ & $+0.00263\,{\scriptstyle\pm\,0.00070}$ & $+0.00490\,{\scriptstyle\pm\,0.00047}$\\
\bottomrule
\end{tabular}
\end{table}
\endgroup

\section{Mechanisms and Applicability}\label{app:mechanisms-complete}
\subsection{Variance and Budget Location in the Remaining Population}\label{app:heterogeneous-empirical}
For all 3,600 splits of the ninety trained models, we inspect the full remaining population
$R=C\cup T$ under the frozen proposal.
We compute the normalized-loss mean $r_R$, population variance $\sigma_R^2$,
and $\nu_R=|R|^{-1}\sum_{u\in R}L_u(1-L_u)$.
For the utility constraint, the macro average of $\nu_R/(\sigma_R^2+\nu_R)$ is $86.81\%$.
Thus auxiliary binary randomization contributes a substantial fraction of the variance in these real fractional losses.

To describe finite-sample budget position, define
\begin{equation*}
 z_R=\frac{r_R-\alpha}
 {\sqrt{\{(M-n)/(M-1)\}\sigma_R^2/n}},
 \qquad M=|R|.
\end{equation*}
Table~\ref{tab:population-regime} labels $|z_R|\le1$ as near the budget,
$z_R<-1$ as below, and $z_R>1$ as above, with zero variance recorded separately.
This fixed descriptive convention uses one actual sample-mean standard deviation.
It is not a verification of the exact boundary assumptions in the asymptotic theorems.

For KuaiRec, the Recall mean is always below the budget and the utility constraint is near it in $70.00\%$ of splits.
For KuaiRand-random, the Recall mean is above the budget in all six hundred splits,
with an average $r_R/\alpha$ of $5.6515$.
For MovieLens-25M, both constraints are near their budgets in $22.00\%$ of splits,
corresponding to the joint-constraint setting addressed by
Theorem~\ref{thm:heterogeneous-multigate}, for which single-constraint coefficients alone are insufficient.
For Amazon-Music and Epinions, both constraints are below their budgets in $95.00\%$ and $95.50\%$ of splits.
These differences align with the budget regimes distinguished by the theory,
while showing that one boundary mechanism cannot explain every task.

Labels in $R$ are used only for this post-hoc mechanism analysis and never enter the selection-set diagnostic.
The exact identity between default activation differences and paired $T$ utility gains agrees
with the formal results in all 180 model-level comparisons across the ninety models.
These are fixed-population findings on observed support, not a new independent confirmation
or evidence that asymptotic coefficients are exact at $n=200$.

\begin{table}[!htb]
\centering\small\setlength{\tabcolsep}{4pt}
\caption{Budget regions in the full remaining population. Entries are percentages of role splits.
One near means that the other constraint is below budget. Music additionally has $0.83\%$ of splits with a zero-variance constraint.
Forty splits are averaged within each model before equal weighting within common training seeds.}\label{tab:population-regime}
\begin{tabular}{lrrrr}
\toprule
Domain & Both below & One near & Both near & Any above\\
\midrule
Jester & $19.50$ & $41.00$ & $13.67$ & $25.83$\\
KuaiRand & $0.00$ & $0.00$ & $0.00$ & $100.00$\\
KuaiRec & $22.33$ & $70.00$ & $0.00$ & $7.67$\\
MovieLens & $2.33$ & $37.67$ & $22.00$ & $38.00$\\
Music & $95.00$ & $4.17$ & $0.00$ & $0.00$\\
Epinions & $95.50$ & $4.50$ & $0.00$ & $0.00$\\
\bottomrule
\end{tabular}
\end{table}

\subsection{Assessing applicability before final calibration}\label{app:prospective-diagnostics}
Three quantities available before final calibration help characterize a proposal: its mean gain on $S$, each mean downside divided by its physical budget, and raw activation headroom. We additionally split the existing $S400$ into two fixed halves. Each half acts as pseudo-$C200$ and evaluates the other half, then we average the products of the ACRC--raw probability increment and held-half mean gain. No $C/T$ outcomes enter this calculation. Because the proposal was already selected on all of $S$, this is a descriptive screening signal, not unbiased validation or a positive-gain certificate. We do not select a new deployment threshold from the test outcomes.

\begin{table}[htb]\centering\small
\caption{Information available on $S$ and the corresponding observed calibration increment on $T$. Risk ratios compare candidate downside with the budget. Headroom is $1-p_0$ on pseudo-calibration halves. Effect columns report paired five-seed means $\pm$ 95\% interval half-widths.}\label{tab:prospective-diagnostics}
\begin{tabular}{lrrrrrr}\toprule
Dataset & $\bar g_S$ & $D_{U,S}/\beta_U$ & $D_{R,S}/\beta_R$ & $1-p_0$ & $\widehat{\Delta U}_S$ & $\Delta U_T$\\\midrule
KuaiRec & 0.249 & 0.857 & 0.021 & 0.201 & $0.035798\,{\scriptstyle\pm\,0.001353}$ & $0.035643\,{\scriptstyle\pm\,0.000621}$\\
KuaiRand & 0.068 & 0.983 & 5.475 & 0.599 & $0.000429\,{\scriptstyle\pm\,0.000056}$ & $0.000448\,{\scriptstyle\pm\,0.000051}$\\
Music & 0.009 & 0.358 & 0.189 & 0.008 & $0.000041\,{\scriptstyle\pm\,0.000038}$ & $0.000108\,{\scriptstyle\pm\,0.000068}$\\
Epinions & 0.036 & 0.524 & 0.205 & 0.028 & $0.000950\,{\scriptstyle\pm\,0.000244}$ & $0.001730\,{\scriptstyle\pm\,0.000511}$\\
ML-25M & 0.100 & 1.074 & 1.037 & 0.334 & $0.011702\,{\scriptstyle\pm\,0.001215}$ & $0.011867\,{\scriptstyle\pm\,0.001297}$\\
Jester & 0.096 & 0.866 & 0.859 & 0.234 & $0.011550\,{\scriptstyle\pm\,0.001300}$ & $0.011561\,{\scriptstyle\pm\,0.001247}$\\
\bottomrule\end{tabular}\end{table}

KuaiRec combines substantial candidate gain with incomplete raw activation. MovieLens and Jester have smaller but still useful gains and two risks near their budgets. Music has little candidate gain and only $0.0083$ raw headroom. KuaiRand has ample headroom but its candidate Recall downside is about $5.47$ times the budget, which limits useful additional activation. Thus headroom alone is not a deployment criterion. The $S$ signal also understates the observed Epinions effect, illustrating its finite-sample limitations.

The equal-domain macro increment is not uniform in magnitude. KuaiRec accounts for $58.09\%$ of the six-domain sum. Removing it leaves a five-domain mean increment of $0.005143$. These leave-one-domain-out numbers are descriptive, not a new confirmation cohort. The practical case is strongest when a useful proposal and recoverable activation coexist.

\subsection{Distribution of realized split risks}\label{app:split-risk-distribution}
Table~\ref{tab:split-risk-distribution} uses the existing $3600$ shared-support splits. The probability of exceeding a budget in one realized split is different from the expected-risk estimand. For randomized methods, each risk integrates activation and action coins analytically. The distribution therefore describes role-assignment variability, without additional deployment-coin noise or a conditional protection claim. Splits share users and models, so the frequencies and quantiles are descriptive and receive no independent-binomial confidence interval.

\begin{table}[htb]\centering\small
\caption{Realized-risk distribution over correlated role splits. Budgets are $0.02$ and $0.01$. Lower values indicate less realized downside, not higher utility. The three SCoRE variants select configurations only on $S$.}\label{tab:split-risk-distribution}
\begin{tabular}{llrrrr}\toprule
Method & Risk & Mean & Exceedance \% & 99th percentile & Maximum\\\midrule
\multirow{2}{*}{raw} & $D_U$ & 0.01113 & 2.08 & 0.02097 & \best{0.02566} \\
 & $D_R$ & 0.00448 & 6.19 & 0.01119 & \best{0.01352} \\
\midrule
\multirow{2}{*}{RB} & $D_U$ & 0.01129 & 2.11 & 0.02107 & \second{0.02571} \\
 & $D_R$ & 0.00462 & 7.58 & 0.01162 & \second{0.01545} \\
\midrule
\multirow{2}{*}{\acrcours{}} & $D_U$ & 0.01249 & 6.64 & 0.02374 & 0.02951 \\
 & $D_R$ & 0.00498 & 10.72 & 0.01236 & 0.01602 \\
\midrule
\multirow{2}{*}{SCoRE risk prediction} & $D_U$ & \second{0.00941} & \best{1.31} & \best{0.02025} & \best{0.02566} \\
 & $D_R$ & \best{0.00320} & \best{0.58} & \best{0.00950} & 0.01609 \\
\midrule
\multirow{2}{*}{SCoRE reward ratio} & $D_U$ & 0.00942 & \second{1.56} & 0.02061 & 0.02651 \\
 & $D_R$ & 0.00328 & 0.92 & 0.00986 & 0.01894 \\
\midrule
\multirow{2}{*}{SCoRE risk or reward} & $D_U$ & \best{0.00940} & 1.58 & \second{0.02059} & 0.02651 \\
 & $D_R$ & \second{0.00323} & \second{0.81} & \second{0.00971} & 0.01616 \\
\bottomrule\end{tabular}\end{table}

ACRC increases mean utility by accepting more realized downside. Its split exceedance frequencies are $6.64\%$ for utility and $10.72\%$ for Recall, versus $2.08\%$ and $6.19\%$ for raw. The corresponding mean positive excesses over the physical budgets are $0.000134$ and $0.000103$. These are averages over all splits, including zero excesses. This distribution complements, rather than strengthens, the expected-risk guarantee.

\subsection{Common activation and selection by user}\label{app:selection-mechanism}
\label{app:selection-mechanism-new}
This decomposition uses $w_+$ tuned only on $S$ and information-matched risk-prediction SCoRE with its feature family selected only on $S$. Fix a split and a proposal. Let $p$ be our common activation probability, $m_u$ the SCoRE decision for user $u$, and $\bar m$ its average activation. Let $g_u$ be the signed proposal gain and $\overline g_T$ its test-user mean. The exact identity is
\begin{equation}
U_+-U_{\rm SCoRE}=(p-\bar m)\overline g_T-\operatorname{Cov}_T(m,g).\label{eq:selection-mechanism-identity}
\end{equation}
The same identity holds for each nonnegative risk after replacing $g$ by its loss. Covariance uses the uniform distribution over the finite test set. We compute both terms for each split before applying the five-seed aggregation, rather than multiplying macro averages. The identity separates activation volume from selection covariance. It is not a unique causal decomposition.
Table~\ref{tab:matched-mechanism-decomposition} reports both terms for all five proposals. Common activation yields higher utility for the selected and reranked proposals, while the PerK proposal favors SCoRE and the UMC difference has an interval containing zero. User selection can therefore remain useful even when common activation increases.
\begingroup
\small\setlength{\tabcolsep}{4pt}
\begin{table}[!htb]\centering\small
\caption{Utility decomposition for information-matched risk-prediction SCoRE, with feature family selected using $S$ only. Entries show paired five-seed intervals. The final two columns are the activation term and the covariance to subtract. Values are means $\pm$ 95\% CI half-widths.}\label{tab:matched-mechanism-decomposition}
\begin{tabular}{lrrr}
\toprule
Proposal & $U_+-U_{\rm SCoRE}$ & $(p-\bar m)\overline g_T$ & $\operatorname{Cov}_T(m,g)$ \\
\midrule
Selected proposal & $+0.016928\,{\scriptstyle\pm\,0.001081}$ & $+0.017707\,{\scriptstyle\pm\,0.001054}$ & $+0.000778\,{\scriptstyle\pm\,0.000257}$ \\
Within-Top20 reranking & $+0.001293\,{\scriptstyle\pm\,0.000310}$ & $+0.001555\,{\scriptstyle\pm\,0.000328}$ & $+0.000263\,{\scriptstyle\pm\,0.000099}$ \\
Pool reranking & $+0.019470\,{\scriptstyle\pm\,0.002052}$ & $+0.020483\,{\scriptstyle\pm\,0.001980}$ & $+0.001013\,{\scriptstyle\pm\,0.000240}$ \\
PerK proposal & $-0.000566\,{\scriptstyle\pm\,0.000105}$ & $-0.004605\,{\scriptstyle\pm\,0.000326}$ & $-0.004039\,{\scriptstyle\pm\,0.000281}$ \\
UMC proposal & $-0.002023\,{\scriptstyle\pm\,0.002228}$ & $+0.000019\,{\scriptstyle\pm\,0.001897}$ & $+0.002042\,{\scriptstyle\pm\,0.000594}$ \\
\bottomrule
\end{tabular}
\end{table}
\endgroup
\subsection{Sample size, loss resolution and budget sensitivity}
\paragraph{Reserved-pool protocol.}
A separate mechanism experiment freezes all seventy-five models and candidate families before generating forty new assignments. Each assignment selects a rule once using $S400$ and the default raw selector. It then reserves four hundred calibration users and evaluates all variants on the same remaining $T$. Sample sizes $50,100,200,400$ use nested prefixes of the reserved pool. Unused reserved users never enter $T$. Thus these absolute utilities are not pooled with the main table.

\begin{table}[!htb]
\centering\footnotesize\setlength{\tabcolsep}{2.8pt}
\caption{Frozen-rule mechanism study. $\Delta U$ is the macro mean of $w_+$ minus raw, higher is better. Counts are positive/equal/negative group means with tolerance $10^{-12}$. Default budgets are $(0.02,0.01)$, and non-size variants use $n=200$. The resolution panel replaces each loss by $\lceil mL\rceil/m$.}\label{tab:mechanism}
\begin{tabular}{@{}lrrlrrlrr@{}}
\toprule
\multicolumn{3}{c}{Calibration sample size} & \multicolumn{3}{c}{Risk budgets} & \multicolumn{3}{c}{Loss resolution}\\
\cmidrule(lr){1-3}\cmidrule(lr){4-6}\cmidrule(lr){7-9}
$n$ & $\Delta U\uparrow$ & $+/=/-$ & $\epsilon$ & $\Delta U\uparrow$ & $+/=/-$ & $m$ & $\Delta U\uparrow$ & $+/=/-$\\
\midrule
$50$ & $+0.004147$ & $15/0/0$ & $(0.0050,0.0025)$ & $+0.003553$ & $15/0/0$ & $1$ & $0.000000$ & $0/15/0$\\
$100$ & $+0.008892$ & $15/0/0$ & $(0.0100,0.0050)$ & $+0.007100$ & $15/0/0$ & $4$ & $+0.002469$ & $15/0/0$\\
$200$ & $+0.009841$ & $15/0/0$ & $(0.0500,0.0200)$ & $+0.000450$ & $7/8/0$ & $16$ & $+0.008978$ & $15/0/0$\\
$400$ & $+0.006352$ & $14/1/0$ &  &  &  & $64$ & $+0.009952$ & $15/0/0$\\
\bottomrule
\end{tabular}
\end{table}

\paragraph{Uncertainty.}
The eleven configurations and fifteen settings form a family of 165 contrasts. Table~\ref{tab:new-intervals} reports all fifteen default $n=200$ intervals. Thirteen unadjusted intervals and six multiplicity-adjusted intervals lie above zero, and none lie entirely below zero. These assignments reuse observed feedback and provide exploratory sensitivity evidence under the reserved-pool protocol.

\begin{table}[!htb]
\centering\small\setlength{\tabcolsep}{6pt}
\caption{Utility increments and intervals under the new reserved-pool protocol with $n=200$, higher is better. A reports unadjusted 95\% intervals. B reports Bonferroni-adjusted intervals for 165 mechanism comparisons. Intervals use approximate $t_4$ inference over five training-seed means. Both interval groups use the same utility-increment means and report each mean $\pm$ the corresponding interval half-width.}\label{tab:new-intervals}
\begin{tabular}{@{}llrrr@{}}
\toprule
Interval & Dataset & LightGCN & XSimGCL & SIGformer\\
\midrule
\multirow{5}{*}{\shortstack[l]{A\\Unadjusted}} & KuaiRec & $+0.056385\,{\scriptstyle\pm\,0.005515}$ & $+0.043870\,{\scriptstyle\pm\,0.005005}$ & $+0.005452\,{\scriptstyle\pm\,0.002182}$\\
 & KuaiRand & $+0.000396\,{\scriptstyle\pm\,0.000044}$ & $+0.000403\,{\scriptstyle\pm\,0.000068}$ & $+0.000371\,{\scriptstyle\pm\,0.000021}$\\
 & Music & $+0.000135\,{\scriptstyle\pm\,0.000038}$ & $+0.000219\,{\scriptstyle\pm\,0.000263}$ & $+0.000008\,{\scriptstyle\pm\,0.000021}$\\
 & Epinions & $+0.001683\,{\scriptstyle\pm\,0.000508}$ & $+0.001018\,{\scriptstyle\pm\,0.000468}$ & $+0.002200\,{\scriptstyle\pm\,0.001318}$\\
 & ML-25M & $+0.004476\,{\scriptstyle\pm\,0.001040}$ & $+0.008837\,{\scriptstyle\pm\,0.002389}$ & $+0.022159\,{\scriptstyle\pm\,0.005400}$\\
\midrule
\multirow{5}{*}{\shortstack[l]{B\\Bonferroni}} & KuaiRec & $+0.056385\,{\scriptstyle\pm\,0.023284}$ & $+0.043870\,{\scriptstyle\pm\,0.021131}$ & $+0.005452\,{\scriptstyle\pm\,0.009211}$\\
 & KuaiRand & $+0.000396\,{\scriptstyle\pm\,0.000186}$ & $+0.000403\,{\scriptstyle\pm\,0.000286}$ & $+0.000371\,{\scriptstyle\pm\,0.000088}$\\
 & Music & $+0.000135\,{\scriptstyle\pm\,0.000160}$ & $+0.000219\,{\scriptstyle\pm\,0.001111}$ & $+0.000008\,{\scriptstyle\pm\,0.000090}$\\
 & Epinions & $+0.001683\,{\scriptstyle\pm\,0.002145}$ & $+0.001018\,{\scriptstyle\pm\,0.001975}$ & $+0.002200\,{\scriptstyle\pm\,0.005564}$\\
 & ML-25M & $+0.004476\,{\scriptstyle\pm\,0.004389}$ & $+0.008837\,{\scriptstyle\pm\,0.010084}$ & $+0.022159\,{\scriptstyle\pm\,0.022798}$\\
\bottomrule
\end{tabular}
\end{table}

\subsection{Finite-population utility decomposition}\label{app:sampling-decomposition}
For each model, we fix the first new $S$ and its selected candidate, then independently draw eighty calibration sets of size $200$ from the remaining fixed population. Every evaluation uses the full complement. Across six thousand draws, the macro incremental utility is $0.00993450$. Its product term is $0.00992159$ and the covariance correction is $+0.00001291$. Sixty-eight model averages are positive and seven are equal. The median absolute correction-to-product ratio is $0.12\%$, while the maximum is $22.06\%$. The correction is often small but cannot be assumed zero or negative for the actual two-constraint procedure. This diagnostic uses the full fixed population and is not a deployment certificate estimated only from $S$.

\par\medskip

\subsection{Removing risk calibration}
Setting activation to one leaves the same selected proposal unchanged. Table~\ref{tab:ungated} retains both utility and risk consequences, including budget violations.

\begin{table}[!htb]\centering\small\setlength{\tabcolsep}{2.8pt}
\caption{Same-rule ungated ablation. Only activation is set to one, retaining heads, rules, the internal action coin and test users. Higher utility may violate the risk budgets. Colors do not imply joint dominance.}\label{tab:ungated}
\begin{tabular}{@{}llrrrrrr@{}}\toprule
Dataset & Model & Ungated $U\uparrow$ & $R_U\downarrow$ & $R_R\downarrow$ & \acrcours{} $U\uparrow$ & $R_U\downarrow$ & $R_R\downarrow$\\\midrule
\multirow{3}{*}{KuaiRec} & LG & \best{0.5814} & \second{0.018759} & \second{0.000233} & \second{0.5391} & \best{0.016667} & \best{0.000206} \\
 & X & \best{0.5561} & \second{0.018546} & \second{0.000219} & \second{0.5221} & \best{0.016547} & \best{0.000195} \\
 & SIG & \best{0.7621} & \second{0.017043} & \second{0.000191} & \second{0.7543} & \best{0.015014} & \best{0.000172} \\
\midrule
\multirow{3}{*}{KuaiRand} & LG & \best{-0.4182} & \second{0.020338} & \second{0.066605} & \second{-0.4846} & \best{0.007583} & \best{0.009855} \\
 & X & \best{-0.4360} & \second{0.021081} & \second{0.061804} & \second{-0.4977} & \best{0.009244} & \best{0.009944} \\
 & SIG & \best{-0.4200} & \second{0.020456} & \second{0.041176} & \second{-0.4555} & \best{0.012247} & \best{0.009640} \\
\midrule
\multirow{3}{*}{Music} & LG & \best{0.2664} & \second{0.008955} & \second{0.003111} & \second{0.2663} & \best{0.008818} & \best{0.003031} \\
 & X & \best{0.2664} & \second{0.005649} & \second{0.003317} & \second{0.2663} & \best{0.005549} & \best{0.003234} \\
 & SIG & \best{0.2660} & \best{0.008131} & \best{0.000475} & \best{0.2660} & \best{0.008131} & \best{0.000475} \\
\midrule
\multirow{3}{*}{Epinions} & LG & \best{0.2603} & \second{0.012612} & \second{0.003096} & \second{0.2596} & \best{0.012305} & \best{0.002940} \\
 & X & \best{0.2616} & \second{0.009629} & \second{0.002896} & \second{0.2613} & \best{0.009533} & \best{0.002836} \\
 & SIG & \best{0.2604} & \second{0.010832} & \second{0.001385} & \second{0.2603} & \best{0.010805} & \best{0.001382} \\
\midrule
\multirow{3}{*}{ML-25M} & LG & \best{0.1574} & \second{0.017515} & \second{0.005401} & \second{0.1559} & \best{0.016896} & \best{0.005194} \\
 & X & \best{0.1862} & \second{0.020456} & \second{0.014135} & \second{0.1616} & \best{0.013853} & \best{0.009617} \\
 & SIG & \best{0.1010} & \second{0.026375} & \second{0.011949} & \second{0.0407} & \best{0.017436} & \best{0.007939} \\
\bottomrule\end{tabular}
\end{table}

\subsubsection{Cost and candidate support}
With backbones and fitted heads fixed, we repeat $S$ selection and $C$ calibration for $\lambda=0.25,0.5,1,2$, and change the observed pool cap to $40$ and $160$ at the default cost. All conditions are retained in Table~\ref{tab:cost-support}. This is decision-stage sensitivity, not predictor retraining for each cost.

Let $m(a)=\sum_{r\le |a|}d_r/W_{20}$. The exact identity

\begin{equation*}
 U_\lambda(a)=\frac{1+\lambda}{2}U_1(a)+\frac{1-\lambda}{2}m(a)
\end{equation*}
computes each cost utility and range on the same observations. For each deterministic action we use

\begin{equation*}
 H_\lambda\le\frac{1+\lambda}{2}H_1+
 \frac{|1-\lambda|}{2}\max_u|m(a_u)-m(\pi_0(u))|.
\end{equation*}
Endpoint bounds are averaged with the original mixture weights. Backbones and probability heads are not retrained, and proposed prefixes retain default-cost construction. Selection and final calibration use the declared cost, without test feedback. Native external actions stay fixed, so non-default cost results do not establish superiority to cost-retuned competitors.

\begin{table}[!htb]\centering\small\setlength{\tabcolsep}{2.8pt}
\caption{Macro means over fifteen settings. Counts compare new versus raw utility on the same selected rule, using tolerance $10^{-12}$. Selection and calibration are repeated for each condition.}\label{tab:cost-support}
\begin{tabular}{@{}rrrrrrr@{}}\toprule
$\lambda$ & Pool & $\Delta U_{+0}\uparrow$ & Higher/tie/lower & $\Delta U_{+\pi_0}\uparrow$ & $R_U\downarrow$ & $R_R\downarrow$\\\midrule
0.25 & 80 & +0.002220 & 14/1/0 & +0.04796 & 0.008987 & 0.002876\\
\midrule
0.50 & 80 & +0.004828 & 14/1/0 & +0.05685 & 0.010754 & 0.003821\\
\midrule
\multirow{3}{*}{1.00} & 160 & +0.010821 & 15/0/0 & +0.07856 & 0.011896 & 0.004462\\
 & 40 & +0.007526 & 15/0/0 & +0.05933 & 0.012104 & 0.004410\\
 & 80 & +0.009959 & 15/0/0 & +0.06914 & 0.012042 & 0.004444\\
\midrule
2.00 & 80 & +0.009301 & 15/0/0 & +0.07793 & 0.013235 & 0.005351\\
\bottomrule\end{tabular}
\end{table}

\subsection{Same-rule calibration increments}
Changing only final calibration gives 117 higher, 48 equal and 0 lower utilities relative to raw across 165 settings. The macro increment is $+0.009959$ at default budgets and $+0.000212$ at thirty-twofold budgets, consistent with the saturation boundary. Operators share the selected rule within a budget, while $S$ may select different rules across budgets.
\paragraph{Where the advantage occurs.}
Among three thousand existing splits on the shared support, activation increased strictly over raw in 2,014 cases. The remaining 986 cases all had raw activation equal to one. The strictly binding constraint changed in 52 cases. Relative to RB, the executed probability increased in 2,713 cases. It was unchanged in 210 cases because the auxiliary count of a binding constraint did not straddle the truncation point. In another 77 cases, the mathematical gain fell below execution precision and disappeared after quantization. The theoretical conditions explain these cases without selecting another rule or changing the budgets.

Retaining the exponential integral reduced the mean upper bound from $0.083414$ to $0.066783$, against an observed mean activation increment of $0.063625$. The mean interval width decreased by $42.38\%$, with no bound violation across all six thousand decisions. The tighter bound evaluates the exponential integral in the same proof exactly. These descriptive mechanism results use existing observed feedback. Repeated splits do not constitute independent experiments or provide new significance evidence.

The amount of additional activation and the value of the affected rule must be reported together. Table~\ref{tab:revision-regime} decomposes the same-rule increment by dataset. Its final column is $\sum_t d_t\overline g_{T,t}/\sum_t d_t$. KuaiRand has substantial auxiliary fractionality, yet the two constraints and smaller gains associated with extra activation limit its utility improvement. On Amazon-Music, raw is already saturated in $495/600$ cases, and the rules receiving extra activation have little remaining utility advantage. This decomposition explains the effect size without attributing gains from input-adaptive rule selection to the calibrator.

\begin{table*}[t]
\centering
\caption{Mechanism decomposition on the existing shared support. Each row contains three backbones, five seeds, and forty role splits. $d$ denotes the executed activation increment over raw. The final column weights candidate utility gains by that increment. All entries are descriptive.}\label{tab:revision-regime}
\small
\begin{tabular}{lrrrr}
\toprule
Dataset & Zero gains & $\overline d\uparrow$ & $\Delta U\uparrow$ & $\frac{\sum_t d_t\overline g_{T,t}}{\sum_t d_t}$\\
\midrule
KuaiRec & 128/600 & 0.12143 & 0.035643 & 0.29354\\
KuaiRand-random & 0/600 & 0.01811 & 0.000448 & 0.02473\\
Amazon-Music & 495/600 & 0.01620 & 0.000108 & 0.00665\\
Epinions & 319/600 & 0.04958 & 0.001730 & 0.03489\\
MovieLens-25M & 44/600 & 0.11281 & 0.011867 & 0.10520\\
\bottomrule
\end{tabular}
\end{table*}
\subsection{Component computational costs}
Timings use a single-thread process and one fixed training seed and role assignment per setting, at default and eightfold budgets. One warm-up precedes seven timed repetitions. We report each case median. Proposed probability-result and harmonic-denominator caches are cleared every time. Timings include exact rational arithmetic and both risks. The implementation jointly computes raw, RB and the new operator, so the reported calibration time includes this comparison overhead.

The $S$ stage runs our original rule selector or fits two default SCoRE random forests. The $C$ stage computes our exact gates or predicts SCoRE calibration scores. The $T$ stage samples our exact activation and action coins and reads frozen lists, or predicts SCoRE test scores, calls both official decisions and reads native or fallback lists. We report CPU and wall times separately. Backbone training, probability fitting, candidate construction and data loading are outside every timing. These are additional costs conditional on frozen outputs, not complete online-system latency.

\begin{table}[!htb]\centering\small\setlength{\tabcolsep}{2.8pt}
\caption{Additional times after frozen outputs, in milliseconds, lower is better. Each entry averages repeated case medians across fifteen real settings. $S=400$, $C=200$, with application normalized to one thousand users.}\label{tab:frontier-time}
\begin{tabular}{@{}rlrrrrrr@{}}\toprule
$s$ & Method & S CPU & C CPU & T CPU/1000 & S wall & C wall & T wall/1000\\\midrule
\multirow{3}{*}{1} & \acrcours{} & 68.005 & 11.264 & 0.182 & 68.018 & 11.264 & 0.182\\
 & SCoRE/P & 476.288 & 7.163 & 25.615 & 476.457 & 7.165 & 25.625\\
 & SCoRE/U & 1653.562 & 8.617 & 30.457 & 1653.925 & 8.615 & 30.460\\
\midrule
\multirow{3}{*}{8} & \acrcours{} & 66.932 & 11.302 & 0.181 & 66.940 & 11.302 & 0.181\\
 & SCoRE/P & 475.630 & 7.174 & 25.596 & 475.704 & 7.173 & 25.597\\
 & SCoRE/U & 1653.568 & 8.606 & 30.539 & 1653.935 & 8.605 & 30.553\\
\bottomrule\end{tabular}
\end{table}

\section{Confirmations after Method Freeze}\label{app:confirmation-complete}
\subsection{New feedback after method freeze}\label{app:fresh-confirmation}
The latest feedback in MovieLens-25M is dated 21 November 2019 at 09:15:03 UTC. Confirmation uses only feedback strictly after this cutoff. Alignment directly compares complete movie, rating and time records without assuming anonymous IDs are shared between releases. We match 152486 historical nodes. The new period supplies 2566 warm-feedback users and 103649 observations, with 905 users eligible under the pre-frozen minimum of twenty observed candidates. After $F200$, the full pool has $705$ users and the shared supported pool has $691$. Each shared assignment uses $S400$, $C200$ and $T91$, with $T105$ for the full pool. Forty assignments are not forty independent datasets.

The fifteen full backbones retain their original weights. Heads, UMC fitting, rule selection and final calibration rerun the previously frozen procedure. New data do not choose architectures, hyperparameters, seeds, costs or budgets. The method was frozen before downloading the new release. Confirmation feedback is temporally separated from and does not overlap with historical feedback. Confirmation is restricted to historically matched warm users and items, not every MovieLens-32M user or new item.

\begin{table}[!htb]
\centering\small\setlength{\tabcolsep}{2.8pt}
\caption{New-feedback utility and both actual downside risks. Native recommenders retain their objectives. The LightGCN empirical utility downside exceeds $0.02$, illustrating why the expectation guarantee is not a guarantee for a finite empirical average. Red and blue mark the best and second-best distinct mean for each metric within each backbone group. Ties share a color. Colors do not indicate statistical significance or joint superiority.}\label{tab:new-confirmation}
\begin{tabular}{@{}lrrrrrrrrr@{}}
\toprule
Method & \multicolumn{3}{c}{LightGCN} & \multicolumn{3}{c}{XSimGCL} & \multicolumn{3}{c}{SIGformer}\\
\cmidrule(lr){2-4}\cmidrule(lr){5-7}\cmidrule(lr){8-10}
 & $U\uparrow$ & $R_U\downarrow$ & $R_R\downarrow$ & $U\uparrow$ & $R_U\downarrow$ & $R_R\downarrow$ & $U\uparrow$ & $R_U\downarrow$ & $R_R\downarrow$\\
\midrule
Top20 & 0.2850 & \best{0.00000} & \best{0.00000} & 0.2876 & \best{0.00000} & \best{0.00000} & 0.1662 & \best{0.00000} & \best{0.00000} \\
PerK-PDCG & 0.2720 & 0.02626 & 0.07102 & 0.2703 & 0.04676 & 0.12140 & 0.1598 & 0.01043 & 0.02897 \\
UMC-top20 & 0.2402 & 0.12743 & 0.03786 & 0.1852 & 0.15663 & 0.04963 & 0.1774 & 0.08436 & 0.02729 \\
raw & 0.3249 & 0.01778 & 0.00607 & 0.3343 & 0.01671 & 0.00571 & 0.2447 & 0.01526 & 0.00551 \\
RB & \second{0.3275} & 0.01886 & 0.00644 & \second{0.3353} & 0.01705 & 0.00583 & \second{0.2458} & 0.01550 & 0.00560 \\
SCoRE/PerK UR & 0.2849 & \second{0.00219} & 0.00575 & 0.2870 & \second{0.00245} & 0.00521 & 0.1651 & \second{0.00191} & 0.00543 \\
SCoRE/UMC UR & 0.2821 & 0.00676 & \second{0.00209} & 0.2853 & 0.00424 & \second{0.00183} & 0.1681 & 0.00560 & \second{0.00180} \\
\acrcours{} & \best{0.3312} & 0.02060 & 0.00702 & \best{0.3418} & 0.01935 & 0.00663 & \best{0.2567} & 0.01762 & 0.00638 \\
\bottomrule
\end{tabular}
\end{table}

\begin{table}[!htb]
\centering\small\setlength{\tabcolsep}{4pt}
\caption{Paired seed differences on new feedback. Intervals use five training-seed differences and are approximate conditional descriptions, not distribution-free utility certificates for arbitrary populations. Values are means $\pm$ 95\% CI half-widths. Each backbone column reports $\Delta U\uparrow$.}\label{tab:new-confirmation-ci}
\begin{tabular}{@{}lrrr@{}}
\toprule
Comparator & LightGCN & XSimGCL & SIGformer\\
\midrule
SCoRE/PerK UR & $+0.04632\,{\scriptstyle\pm\,0.00331}$ & $+0.05489\,{\scriptstyle\pm\,0.01187}$ & $+0.09163\,{\scriptstyle\pm\,0.00821}$\\
RB & $+0.00372\,{\scriptstyle\pm\,0.00086}$ & $+0.00659\,{\scriptstyle\pm\,0.00278}$ & $+0.01090\,{\scriptstyle\pm\,0.00258}$\\
SCoRE/UMC UR & $+0.04907\,{\scriptstyle\pm\,0.00422}$ & $+0.05649\,{\scriptstyle\pm\,0.01070}$ & $+0.08867\,{\scriptstyle\pm\,0.00927}$\\
Top20 & $+0.04620\,{\scriptstyle\pm\,0.00349}$ & $+0.05424\,{\scriptstyle\pm\,0.01068}$ & $+0.09054\,{\scriptstyle\pm\,0.00817}$\\
raw & $+0.00630\,{\scriptstyle\pm\,0.00082}$ & $+0.00756\,{\scriptstyle\pm\,0.00182}$ & $+0.01204\,{\scriptstyle\pm\,0.00173}$\\
\bottomrule
\end{tabular}
\end{table}

\subsection{Frozen Jester confirmation and external comparisons}\label{app:jester}

\begin{figure}[htbp]
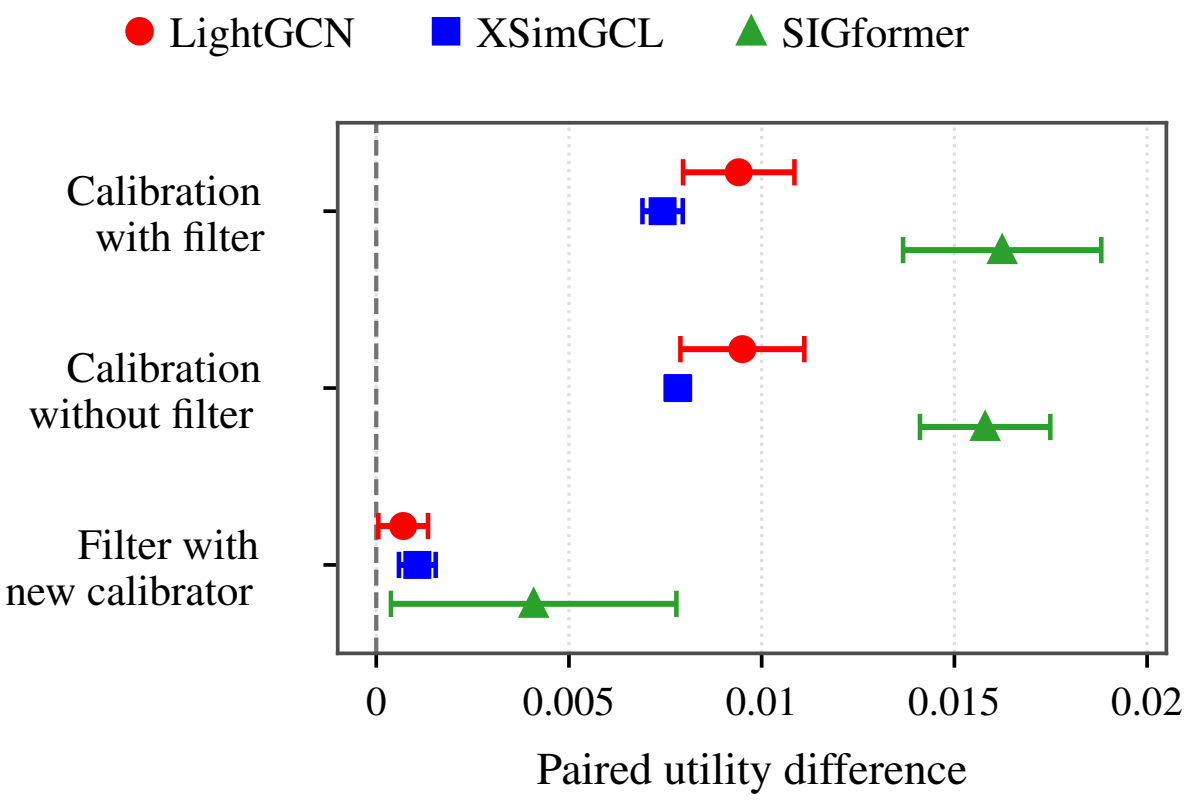

\centering
\begingroup\makeatletter
\resizebox{0.52\linewidth}{!}{%
\begingroup%
\makeatletter%
%
%
\makeatother%
\endgroup%

}%
\makeatother\endgroup
\caption{Jester confirmation on all eligible users. Paired five-seed 95\% intervals separate calibration at fixed filtering from filtering at fixed calibration. The first row is primary. Contrasts are not additive.}
\label{fig:jester-module-evidence}
\end{figure}
\subsubsection{Evaluation protocol}
The three official Jester-1 files contain 73421 users, 100 jokes, and 4136360 observed ratings\cite{jester}. We retain the complete source and split observed pairs within user into 60/10/30 training, validation, and evaluation portions without using the labels. Ratings above zero are positive. The missing code 99 never becomes a negative evaluation label. The warm training graph has 72443 users and 1437095 positive edges. Evaluation requires twenty observed candidates, leaving 31653 eligible users before fitting. After fixed $F200$, the full pool has 31453 users and the original PerK-supported pool has 30699. Each assignment uses $S400$, $C200$, and the remaining $T$. No timestamps are available, so this is neither a temporal-shift experiment nor an online causal evaluation.

Each of the three complete 64-dimensional, three-layer backbones uses five training seeds. LightGCN and XSimGCL run for 1000 epochs, while SIGformer follows author-validation early stopping. The method and primary endpoint were frozen before downloading Jester. Head fitting, raw rule selection, and exact two-risk calibration follow Appendix~\ref{app:implementation}, and the optional filter follows Appendix~\ref{app:decision-diagnosis}.

The primary contrast replaces raw by $w_+$ for the same selected rule, filter, and complete $C200$. The unfiltered contrast was prespecified as secondary. Inference uses all eligible users, while external comparisons use the separate PerK-supported cohort. Results from these populations are not interchangeable.

\subsubsection{Calibrator and filter contributions}
The prespecified macro increment is $0.011025\,{\scriptstyle\pm\,0.001082}$ using an approximate 95\% confidence interval. All fifteen backbone/seed means are positive. Holm-adjusted $p$ values for the three backbone contrasts are $0.0001104$, $0.0000074$, and $0.0001104$ in LightGCN, XSimGCL, SIGformer order. These tests use five training seeds on the same fixed feedback and a working normal approximation, rather than a distribution-free user-population test.

Without filtering, the calibration increment is $0.011040\,{\scriptstyle\pm\,0.001031}$. Holding the new calibrator fixed, filtering adds $0.001949\,{\scriptstyle\pm\,0.000950}$. Calibration gains therefore persist without filtering. Relative to unfiltered raw, the complete filtered new procedure adds $0.012990$. Component contrasts depend on the path through this factorial comparison and should not be added arbitrarily.

\begin{table}[!htb]
\centering\small\setlength{\tabcolsep}{4pt}
\caption{Prespecified confirmation and secondary paired effects on all eligible Jester users. Approximate 95\% intervals use five seed means, with forty splits averaged within seed. Values are means $\pm$ 95\% CI half-widths. Each backbone column reports $\Delta U\uparrow$.}\label{tab:jester-effects}
\begin{tabular}{@{}lrrr@{}}
\toprule
Contrast & LightGCN & XSimGCL & SIGformer\\
\midrule
Calibration with filter & $+0.009405\,{\scriptstyle\pm\,0.001445}$ & $+0.007430\,{\scriptstyle\pm\,0.000523}$ & $+0.016240\,{\scriptstyle\pm\,0.002571}$\\
Calibration without filter & $+0.009499\,{\scriptstyle\pm\,0.001608}$ & $+0.007825\,{\scriptstyle\pm\,0.000313}$ & $+0.015798\,{\scriptstyle\pm\,0.001691}$\\
Filter at new calibrator & $+0.000699\,{\scriptstyle\pm\,0.000644}$ & $+0.001065\,{\scriptstyle\pm\,0.000477}$ & $+0.004084\,{\scriptstyle\pm\,0.003703}$\\
\bottomrule
\end{tabular}
\end{table}

\subsubsection{Risk-budget use and effect magnitude}
Compared with raw on the same filtered rule, $w_+$ increases activation and uses more risk budget. Macro utility downside rises from $0.012431$ to $0.014512$, and Recall downside from $0.006466$ to $0.007563$, below budgets $0.02/0.01$. Finite-repeat means and seed intervals do not guarantee risk conditional on one calibration sample. Mean utility improvements relative to raw are approximately $3.21\%$, $2.27\%$, and $6.31\%$ for the three backbones. The prespecified $0.005$ margin is a research screening criterion for normalized utility, rather than a commercially validated revenue threshold.

\begin{table}[!htb]
\centering\small\setlength{\tabcolsep}{4pt}
\caption{Risk-budget use for the same filtered rule. Lower risks are better. Intervals describe training-seed variation and do not replace the expected-risk theorem. Values are means $\pm$ 95\% CI half-widths. Red and blue mark the best and second-best distinct mean for each metric within each backbone group. Ties share a color. Colors do not indicate statistical significance or joint superiority.}\label{tab:jester-risk}
\begin{tabular}{@{}llrrr@{}}
\toprule
Metric & Calibrator & LightGCN & XSimGCL & SIGformer\\
\midrule
\multirow{2}{*}{$R_U\downarrow$} & raw & \best{$0.012164\,{\scriptstyle\pm\,0.000581}$} & \best{$0.011597\,{\scriptstyle\pm\,0.000256}$} & \best{$0.013532\,{\scriptstyle\pm\,0.000750}$} \\
 & \acrcours{} & \second{$0.014300\,{\scriptstyle\pm\,0.000420}$} & \second{$0.013399\,{\scriptstyle\pm\,0.000275}$} & \second{$0.015838\,{\scriptstyle\pm\,0.000601}$} \\
\midrule
\multirow{2}{*}{$R_R\downarrow$} & raw & \best{$0.006513\,{\scriptstyle\pm\,0.000147}$} & \best{$0.006141\,{\scriptstyle\pm\,0.000090}$} & \best{$0.006743\,{\scriptstyle\pm\,0.000296}$} \\
 & \acrcours{} & \second{$0.007682\,{\scriptstyle\pm\,0.000221}$} & \second{$0.007089\,{\scriptstyle\pm\,0.000075}$} & \second{$0.007917\,{\scriptstyle\pm\,0.000398}$} \\
\bottomrule
\end{tabular}
\end{table}

\subsubsection{Complete original external comparisons}
\begin{figure*}[t]
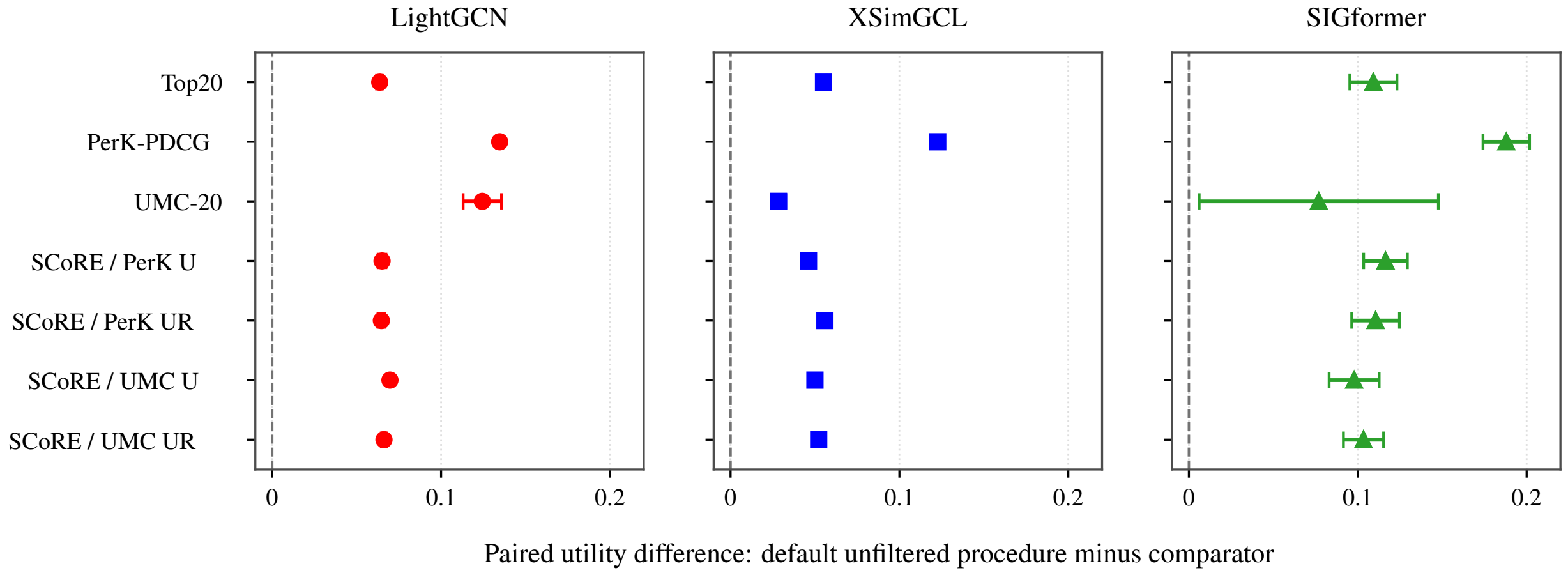

\centering
\begingroup\makeatletter
\resizebox{\linewidth}{!}{%
\begingroup%
\makeatletter%
%
%
\makeatother%
\endgroup%

}%
\makeatother\endgroup
\caption{Complete-pipeline utility comparisons on the Jester cohort supported by original PerK. Differences subtract each comparator from the default new-calibrator procedure without the optional filter. Points and bars show five-seed means and approximate pointwise 95\% $t_4$ intervals, with forty splits averaged within seed. The panels share one linear scale and retain all seven comparators. These supplementary comparisons include differences in prediction heads and list rules, so the gains cannot be attributed solely to calibration. They do not compare matched empirical risks or constitute new prespecified confirmation tests. SCoRE U uses one utility-risk call and UR intersects two unmodified calls.}
\label{fig:jester-external-evidence}
\end{figure*}

External methods follow Appendices~\ref{app:implementation} and~\ref{app:review-completion}, retaining native calibration and list rules, the declared risk ranges, and physical budgets $0.02/0.01$. Backbone and UMC cutoffs remain $k=1,5,10,20$. SCoRE U uses one native utility-risk call and UR intersects two native calls. The supplementary SCoRE comparisons were added after the primary confirmation without redefining its endpoint or tuning parameters.

For the dense 100-item Jester catalog, PerK's negative sampler replenishes its pregenerated buffer with identically distributed uniform draws only when exhausted. This implementation adaptation preserves its sampling distribution and sample count, Platt objective, and prefix decision.

All methods are evaluated within the same supported cohort, and external operating points are not selected using test risks. Figure~\ref{fig:jester-external-evidence} and Table~\ref{tab:jester-external-effects} report all 21 backbone/comparator effects, whose means are positive. These complete-pipeline effects include differences in prediction heads and list rules and cannot be attributed solely to calibration.

\begin{table}[!htb]
\centering\small\setlength{\tabcolsep}{4pt}
\caption{Paired utility differences between the default unfiltered new-calibrator procedure and external methods on the supported cohort. Approximate five-seed intervals are supplementary pointwise summaries, not new prespecified confirmation tests. Values are means $\pm$ 95\% CI half-widths. All five seeds have positive differences in each of the twenty-one comparisons. Each backbone column reports $\Delta U\uparrow$.}\label{tab:jester-external-effects}
\begin{tabular}{@{}lrrr@{}}
\toprule
Comparator & LightGCN & XSimGCL & SIGformer\\
\midrule
Top20 & $+0.06367\,{\scriptstyle\pm\,0.00192}$ & $+0.05505\,{\scriptstyle\pm\,0.00182}$ & $+0.10924\,{\scriptstyle\pm\,0.01395}$\\
PerK-PDCG & $+0.13470\,{\scriptstyle\pm\,0.00174}$ & $+0.12270\,{\scriptstyle\pm\,0.00222}$ & $+0.18794\,{\scriptstyle\pm\,0.01380}$\\
UMC-20 & $+0.12442\,{\scriptstyle\pm\,0.01136}$ & $+0.02839\,{\scriptstyle\pm\,0.00430}$ & $+0.07691\,{\scriptstyle\pm\,0.07082}$\\
SCoRE / PerK U & $+0.06504\,{\scriptstyle\pm\,0.00233}$ & $+0.04618\,{\scriptstyle\pm\,0.00166}$ & $+0.11642\,{\scriptstyle\pm\,0.01293}$\\
SCoRE / PerK UR & $+0.06462\,{\scriptstyle\pm\,0.00201}$ & $+0.05588\,{\scriptstyle\pm\,0.00177}$ & $+0.11054\,{\scriptstyle\pm\,0.01412}$\\
SCoRE / UMC U & $+0.06972\,{\scriptstyle\pm\,0.00171}$ & $+0.04997\,{\scriptstyle\pm\,0.00244}$ & $+0.09784\,{\scriptstyle\pm\,0.01482}$\\
SCoRE / UMC UR & $+0.06618\,{\scriptstyle\pm\,0.00161}$ & $+0.05219\,{\scriptstyle\pm\,0.00196}$ & $+0.10339\,{\scriptstyle\pm\,0.01189}$\\
\bottomrule
\end{tabular}
\end{table}

\begin{table}[!htbp]\centering\small\setlength{\tabcolsep}{2.6pt}
\caption{Jester, LightGCN, PerK-supported cohort. The first three metrics show seed mean and standard deviation, the rest show means. Red and blue rank reported means at the displayed precision, not significance or certification.}\label{tab:jester-shared-lightgcn}
\begin{tabular}{@{}lrrrrrrr@{}}\toprule
Method & $U\uparrow$ & NDCG$\uparrow$ & Recall$\uparrow$ & F1$\uparrow$ & Length & $R_U\downarrow$ & $R_R\downarrow$\\\midrule
Top20 & $\second{0.2608}\pm0.0010$ & $\second{0.7619}\pm0.0007$ & $\second{0.8136}\pm0.0002$ & $\second{0.6705}$ & $20.000$ & $\best{0.0000}$ & $\best{0.0000}$\\
PerK-PDCG & $0.1897\pm0.0017$ & $0.4930\pm0.0021$ & $0.4627\pm0.0021$ & $0.4501$ & $10.903$ & $0.1179$ & $0.3509$\\
UMC-20 & $0.2000\pm0.0098$ & $0.7201\pm0.0059$ & $0.7971\pm0.0050$ & $0.6568$ & $20.000$ & $0.1084$ & $0.0430$\\
\acrcours{} & $\best{0.3244}\pm0.0012$ & $\best{0.8030}\pm0.0010$ & $\best{0.8388}\pm0.0016$ & $\best{0.6911}$ & $19.989$ & $0.0144$ & $0.0078$\\
SCoRE / PerK UR & $0.2598\pm0.0010$ & $0.7587\pm0.0007$ & $0.8085\pm0.0002$ & $0.6686$ & $19.853$ & $\second{0.0022}$ & $0.0052$\\
SCoRE / UMC UR & $0.2583\pm0.0008$ & $0.7602\pm0.0006$ & $0.8131\pm0.0003$ & $0.6700$ & $20.000$ & $0.0046$ & $\second{0.0017}$\\
\bottomrule
\end{tabular}

\par\smallskip

\caption{Jester, XSimGCL. Cohort, reporting, and colors follow Table~\ref{tab:jester-shared-lightgcn}.}\label{tab:jester-shared-xsimgcl}
\begin{tabular}{@{}lrrrrrrr@{}}\toprule
Method & $U\uparrow$ & NDCG$\uparrow$ & Recall$\uparrow$ & F1$\uparrow$ & Length & $R_U\downarrow$ & $R_R\downarrow$\\\midrule
Top20 & $0.2986\pm0.0009$ & $0.7913\pm0.0005$ & $0.8159\pm0.0007$ & $0.6708$ & $20.000$ & $\best{0.0000}$ & $\best{0.0000}$\\
PerK-PDCG & $0.2310\pm0.0010$ & $0.4942\pm0.0013$ & $0.4330\pm0.0015$ & $0.4455$ & $9.629$ & $0.1274$ & $0.3829$\\
UMC-20 & $\second{0.3253}\pm0.0021$ & $\second{0.8025}\pm0.0014$ & $\second{0.8362}\pm0.0017$ & $\second{0.6894}$ & $20.000$ & $0.0590$ & $0.0262$\\
\acrcours{} & $\best{0.3537}\pm0.0016$ & $\best{0.8242}\pm0.0009$ & $\best{0.8476}\pm0.0007$ & $\best{0.6976}$ & $20.000$ & $0.0132$ & $0.0070$\\
SCoRE / PerK UR & $0.2978\pm0.0009$ & $0.7882\pm0.0005$ & $0.8112\pm0.0006$ & $0.6690$ & $19.854$ & $\second{0.0020}$ & $0.0048$\\
SCoRE / UMC UR & $0.3015\pm0.0006$ & $0.7926\pm0.0004$ & $0.8181\pm0.0006$ & $0.6728$ & $20.000$ & $0.0055$ & $\second{0.0023}$\\
\bottomrule
\end{tabular}

\par\smallskip

\caption{Jester, SIGformer. Cohort, reporting, and colors follow Table~\ref{tab:jester-shared-lightgcn}.}\label{tab:jester-shared-sigformer}
\begin{tabular}{@{}lrrrrrrr@{}}\toprule
Method & $U\uparrow$ & NDCG$\uparrow$ & Recall$\uparrow$ & F1$\uparrow$ & Length & $R_U\downarrow$ & $R_R\downarrow$\\\midrule
Top20 & $0.1845\pm0.0189$ & $0.7071\pm0.0122$ & $0.7859\pm0.0078$ & $0.6508$ & $20.000$ & $\best{0.0000}$ & $\best{0.0000}$\\
PerK-PDCG & $0.1058\pm0.0152$ & $0.3874\pm0.0178$ & $0.3714\pm0.0182$ & $0.3546$ & $9.538$ & $0.1345$ & $0.4145$\\
UMC-20 & $\second{0.2169}\pm0.0551$ & $\second{0.7314}\pm0.0351$ & $\second{0.7943}\pm0.0170$ & $\second{0.6560}$ & $20.000$ & $0.0669$ & $0.0312$\\
\acrcours{} & $\best{0.2938}\pm0.0084$ & $\best{0.7804}\pm0.0053$ & $\best{0.8226}\pm0.0065$ & $\best{0.6799}$ & $19.974$ & $0.0167$ & $0.0082$\\
SCoRE / PerK UR & $0.1832\pm0.0190$ & $0.7039\pm0.0120$ & $0.7809\pm0.0076$ & $0.6485$ & $19.870$ & $\second{0.0021}$ & $0.0051$\\
SCoRE / UMC UR & $0.1904\pm0.0162$ & $0.7111\pm0.0103$ & $0.7879\pm0.0059$ & $0.6523$ & $20.000$ & $0.0048$ & $\second{0.0022}$\\
\bottomrule
\end{tabular}
\end{table}

\end{document}